%% file: arxiv.tex
\documentclass{article}

\usepackage{microtype}
\usepackage{graphicx}
\usepackage{booktabs} 
\usepackage{subfiles}
\usepackage{physics}

\usepackage[conf={normal}]{proof_at_the_end}

\usepackage[round]{natbib}

\usepackage{enumitem}

\usepackage{xspace}
\makeatletter
\DeclareRobustCommand\onedot{\futurelet\@let@token\@onedot}
\def\@onedot{\ifx\@let@token.\else.\null\fi\xspace}

\def\ie{\emph{i.e}\onedot}

\def\wrt{w.r.t\onedot} 
\def\etal{\emph{et al}\onedot}
\def\aka{\emph{a.k.a}\onedot}

\def \slmm {SLmM\xspace}
\newcommand{\of}{\bar{f}}

\usepackage{gordon}

\newcommand{\TD}{\ensuremath{\mathtt{D}}}
\newcommand{\dxxv}{\TD_{\xv\xv}}

\def \a {\alpha}
\def \b {\beta}
\def \e {\epsilon}
\def \R {\mathbb{R}}
\def \E {\mathds{E}}

\def \s {\sigma}
\def \d {\delta}
\def \g {\gamma}
\def \bmat {\begin{matrix}}
\def \emat {\end{matrix}}
\def \l {\lambda}
\def \X {\mathcal{X}}
\def \Y {\mathcal{Y}}
\def \be {\begin{eqnarray}}
\def \en {\end{eqnarray}}

\def \tr {\nonumber \\}

\def \k {\kappa}
\def \n {\partial}

\newcommand{\Fa}{\texttt{F}\xspace}
\newcommand{\La}{\texttt{L}\xspace}

\usepackage{fullpage}

\usepackage{hyperref}
\hypersetup{
    colorlinks,
    citecolor=black,
    filecolor=black,
    linkcolor=black,
    urlcolor=black
}
\usepackage{cleveref}

\begin{document}

\title{Newton-type Methods for Minimax Optimization}

\author{Guojun Zhang$^{12}$, Kaiwen Wu$^{12}$, Pascal Poupart$^{123}$ and Yaoliang Yu$^{12}$\\
 $^1$University of Waterloo, Waterloo AI Institute \\ 
 $^2$Vector Institute\\
 $^3$Borealis AI \\
  \texttt{\{guojun.zhang, kaiwen.wu, ppoupart, yaoliang.yu\}@uwaterloo.ca} \\
  }

\maketitle

\begin{abstract}
Differential games, in particular two-player sequential zero-sum games (a.k.a.~minimax optimization), have been an important modeling tool in applied science and received renewed interest in machine learning due to many recent applications, such as adversarial training, generative models and reinforcement learning. However, existing theory mostly focuses on convex-concave functions with few exceptions. In this work, we propose two novel Newton-type algorithms for nonconvex-nonconcave minimax optimization. We prove their local convergence at strict local minimax points, which are surrogates of global solutions. We argue that our Newton-type algorithms nicely complement existing ones in that (\textbf{a}) they converge faster to strict local minimax points; (\textbf{b}) they are much more effective when the problem is ill-conditioned; (\textbf{c}) their computational complexity remains similar. We verify the effectiveness of our Newton-type algorithms through experiments on training GANs which are intrinsically nonconvex and ill-conditioned. Our code is available at \url{https://github.com/watml/min-max-2nd-order}.
\end{abstract}

\section{Introduction}
\label{sec:intro}
\begin{table*}[t]
\caption{Comparison among algorithms for minimax optimization.
$p$ and $p^\prime$ are the numbers of conjugate gradient (CG) steps to solve $\yyv^{-1}\n_\yv$ and $\dxxv^{-1} \n_\xv$ respectively.
$\rho_\La$ and $\rho_\Fa$ are the asymptotic linear rates defined in Thm.~\ref{thm:tgda_fr}. $n$ and $m$ are dimensions of the leader and the follower.
The convergence rates of TGDA/FR/GDN/CN are exact when we take enough number of CG steps. By solving ill-conditioning we mean that the convergence rates are not affected the condition numbers. } 
\begin{center}
\label{tbl:compare}
\begin{tabular}{lllll}
\toprule
\textbf{Algorithm}  &\textbf{Time per step} & \textbf{Memory} & \textbf{Convergence rate} & \textbf{Solve ill-conditioning?} \\
\midrule
GDA & $O(n + m)$ & $O(n + m)$ & linear; $\rho_\La \vee \rho_\Fa$ at best & no\\
TGDA/FR  & $O(n + mp)$ & $O(n + m)$ & linear; $\rho_\La\vee \rho_\Fa$ & no\\
GDN & $O(n + m p)$ & $O(n + m)$ & linear; $\rho_\La$  & yes for $\yyv f$\\
CN & $O((n + m)p' + mp)$ & $O(n + m)$ & quadratic & yes for $\yyv f$ and $\dxxv f$ \\
\bottomrule
\end{tabular}
\end{center}
\end{table*}

Differential games have always played an important role in applied science, from their early applications in economics \citep{morgenstern1953theory} to their Lagrangian reformulation in optimization theory and algorithms, and to their recent resurgence in machine learning (ML). Two-player sequential games, \aka minimax optimization, have been the key piece in recent models such as generative adversarial networks (GANs) \citep{goodfellow2014generative, arjovsky2017wasserstein},
adversarial training \citep{ganin2016domain, madry2017towards}, reinforcement learning \citep{du2017stochastic, dai2018sbeed}, federated learning \citep{MohriSS19} and algorithmic fairness \citep{song2019learning}. Solution concepts, most notably Nash equilibrium \citep{nash1950equilibrium}, have been invented, for which numerous algorithms have been designed, including  gradient-descent-ascent (GDA) \citep{arrow1958studies}, extra-gradient (EG) \citep{korpelevich1976extragradient, popov1980modification, mertikopoulos2018optimistic, hsieh2019convergence}, mirror-prox \citep{nemirovski2004prox}. They largely rely on the utility function to be convex-concave, which modern ML applications do not necessarily satisfy. 


%

Motivated by recent applications in ML, we study \emph{nonconvex-nonconcave} minimax optimization.
The nonconvexity brings two immediate difficulties: (\textbf{a}) global solution concepts no longer apply and we are forced to consider their localized versions; (\textbf{b}) the order of which player moves first becomes consequential, due to the lack of strong duality. How to theoretically and algorithmically cope with these new challenges has become a hot research topic, of which we mention the differential Stackelberg equilibrium \citep{fiez2019convergence}, the local minimax points \citep{jin2019minmax, wang2019solving, zhang2020optimality} and the proximal equilibrium \citep{farnia2020gans}. Many algorithms, old and new, have been thoroughly tested, especially by researchers interested in training GANs. 

In fact, many existing algorithms can be treated as \emph{inexact} implementations of Uzawa's approach \citep{arrow1958studies}, i.e., fast follower \Fa and slow leader \La. One could use Gradient Ascent (GA) with a large step size for \Fa and Gradient Descent (GD) with a small step size for \La, which is known as two-time-scale update \citep{Borkar08, heusel2017gans,jin2019minmax}; or perform $k$ steps of GA update for \Fa after every step of GD update of \La \citep{goodfellow2014generative, madry2017towards}. 
Following Uzawa's approach, we use a Newton step for \Fa and a GD update for $\La$, that we call the \emph{Gradient-Descent-Newton} (\aka GD-Newton, GDN) algorithm. Although the method sounds simple and natural, surprisingly, it has not been well studied for non-convex-concave minimax problems, especially for the solution concepts such as the differential Stackelberg equilibrium \citep{fiez2019convergence} and strict local minimax points \citep{evtushenko1974some, jin2019minmax}. We compare GDN with similar algorithms that use the Hessian inverse, such as Total Gradient Descent Ascent \citep[TGDA,][]{evtushenko1974iterative, fiez2019convergence} and Follow-the-Ridge \citep[FR,][]{evtushenko1974iterative, wang2019solving}. Although the three algorithms share the same complexity, GDN has faster local convergence when the \emph{follower problem} is ill-conditioned. A similar conclusion can be drawn by comparing with GDA methods.

Algorithms above achieve local linear convergence and still suffer from the ill-conditioning of the \emph{leader problem} (and the follower problem except our GDN). Fortunately, in \S\ref{sec:second} we show that the Hessian for the leader is also well-defined and we propose the Complete Newton (CN) algorithm that performs Newton updates for both the leader and follower. CN enjoys local \emph{quadratic} convergence and evades the ill-conditioning of both leader and follower problems in a local neighborhood.

To the best of our knowledge, this is the first \emph{genuine} second-order algorithm for nonconvex-nonconcave minimax optimization that (locally) converges super-linearly to (strict) local minimax solutions.\footnote{In \citet{evtushenko1974iterative}, a superlinear algorithm was proposed, but its convergence was not formally proved. See Appendix~\ref{app:related}.}
Rather surprisingly, we show that CN, being a second-order algorithm, can be implemented in similar complexity as the first-order alternatives such as TGDA, FR, and GDN.
In \S\ref{sec:exp}, we verify theoretical properties of our Newton-type algorithms through experiments on training GANs. 
We exclude studying other first-order algorithms such as CO \citep{mescheder2017numerics}, LOLA \citep{foerster2018learning} and SGA \citep{balduzzi2018mechanics} since they may not converge to local minimax points. 

\paragraph{Contributions.}

We propose two Newton-type algorithms (GDN and CN) for minimax optimization that share similar complexity as existing alternatives but locally converge much faster, especially for ill-conditioned problems.
To implement the Newton update, we take a \emph{Hessian-free approach} \citep{Martens10} using only Hessian-vector products and Conjugate Gradient (CG), for which the per iteration complexity and memory usage are linear (see App.~\ref{app:impltsn}).

Our results are summarized in Table~\ref{tbl:compare}.
We also perform experiments on training GANs to complement our theoretical results which offer empirical insights on the aforementioned algorithms.
Our proofs are deferred to Appendix~\ref{app:proof}.

\subfile{sections/2-background}
\subfile{sections/3-first}

\subfile{sections/4-second}

\subfile{sections/5-exp}

\subfile{sections/6-conclusion}


\bibliographystyle{apalike}
\bibliography{refs, nonconvex_concave}

\onecolumn

\appendix
\subfile{appendices/B-proofs}

\subfile{appendices/E-2ts-gda}
\subfile{appendices/C-exp}

\subfile{appendices/D-simu-alt}

\section{Related work}\label{app:related}

We introduce two papers \citep{evtushenko1974some, evtushenko1974iterative} by Evtushenko. His work provides many insights for minimax optimization but is largely unfamiliar to the machine learning community. In \citet{evtushenko1974some}, the author provided the following definition of local minimax points:
\begin{defn}[{\textbf{local minimax}, \citet[][Definition 1]{evtushenko1974some}}]
$\zv^* = (\xv^*, \yv^*)$ is a local minimax point of function $f: \R^n \times \R^m \to \R$ if there exists a neighborhood $\Xc\times \Yc$ of $\zv^*$ s.t.
\begin{itemize}[topsep=0pt]
\item For any $\yv\in \Yc$, $f(\xv^*, \yv) \leq f(\xv^*, \yv^*)$; 
\item For any $\xv\in \Xc$, $\max_{\yv\in \Yc} f(\xv, \yv) \geq \max_{\yv\in \Yc} f(\xv^*, \yv)$.
\end{itemize}
\end{defn}
Specifically, a local minimax point is called \emph{global minimax} if $\X = \R^n$ and $\Y = \R^m$ in the definition above. In \cite{evtushenko1974some}, it is shown that there is a unique global (local) minimax point which is also a stationary point, under some assumptions:
\begin{thm}[{\textbf{stationarity}, \citet[][Theorem 2]{evtushenko1974some}}]\label{thm:stationary}
Let $\yyv f(\xv, \yv) \cl \zero$ and $f(\xv, r(\xv))$ be strict convex for any $\xv\in \R^n$ and $\yv\in \R^m$. The local minimax point of $f$ is unique and at the same time a global minimax and stationary point.
\end{thm}

The sufficient and necessary second-order conditions for local minimax points are also given:
\begin{thm}[{\textbf{sufficient and necessary conditions}, \citet[][Theorem 1]{evtushenko1974some}}]
Let $f: \R^n \times \R^m \to \R$ be thrice continuous differentiable. In order for $\zv^* = (\xv^*, \yv^*)$ to be local minimax, it is sufficient that:
\begin{align}
&\n_\yv(\zv^*) = \n_\xv(\zv^*) = \zero, \, \yyv f(\zv^*) \cl \zero \cl \dxxv f(\zv^*);
\end{align}
it is necessary that $\n_\yv(\zv^*) = \zero$ and $\yyv f(\zv^*) \cle \zero$, and $\dxxv f(\zv^*) \cge \zero$ if $\yyv f(\zv^*) \cl \zero$.
\end{thm}

The assumption in Theorem~\ref{thm:stationary} is a generalization of strictly-convex-strictly-concave functions, that is called strictly minimaximal:
\begin{defn}[{\textbf{strictly minimaximal}, \citet[][p.~134]{evtushenko1974some}}]
A function $f: \R^n \times \R^m \to \R$ is called strictly minimaximal if for all $\zv\in \R^{n+m}$, $\yyv f(\zv) \cl \zero \cl \dxxv f(\zv)$.
\end{defn}

 By connecting the local minimax point to the saddle point of a transformed function, $g(\xv, \yv):= f(\xv, \yv + r(\xv))$, the author proposed the continuous dynamics of FR, TGDA (\Cref{sec:tgda_fr_main}) and TGD-Newton (\Cref{app:tgd}):
 \begin{align}
    \label{eq:alg_fr} \textrm{FR: }& \dv{\xv}{t} = -\n_\xv f(\xv, \yv), \,  \dv{\yv}{t} = (\n_\yv + \yyv \cdot \yxv \cdot \n_\xv)f(\xv, \yv); \\
    \textrm{TGDA: }& \dv{\xv}{t} = -\TD_\xv f(\xv, \yv), \, \dv{\yv}{t} = \n_\yv f(\xv, \yv); \\
    \textrm{TGD-Newton: } & \dv{\xv}{t} = -\TD_\xv f(\xv, \yv), \, \dv{\yv}{t} = -(\yyv \cdot \n_\yv)f(\xv, \yv).
 \end{align}

In a followup paper, the author analyzed these methods and pointed out their local linear convergence \citep[][Theorem 1]{evtushenko1974iterative} at {\slmm}s. \citet[][Theorem 1]{evtushenko1974iterative} also showed if the step size is small enough, the discrete versions (i.e.~by replacing $\dv{\xv}{t} = g_1(\xv, \yv), \dv{\yv}{t} = g_2(\xv, \yv)$ with $\xv_{t+1} - \xv_t = \a g_1(\xv_t, \yv_t), \yv_{t+1} - \yv_t = \a g_2(\xv_t, \yv_t)$, where $g_1: \R^n \times \R^m \to \R^n$ and $g_2: \R^n \times \R^m \to \R^m$) also have linear convergence, though only the explicit form of FR (with $\a_\La = \a_\Fa = \a$) was given. However, there are a few caveats: (a) the exact neighborhood of initialization was not given; (b) the exact linear rates were not given, although a similar spectral analysis was done; (c) all the discrete versions have $\a_\La = \a_\Fa = \a$.

Finally, \citet{evtushenko1974iterative} pointed out an algorithm that uses the inverse of the total second-order derivative $\TD_{\xv\xv}$, that has a local quadratic convergence rate \citep[][Theorem 2]{evtushenko1974iterative}:
\be\label{eq:evtushenko_cn}
\xv_{t+1} = \xv_t - [(\TD_{\xv\xv}^{-1} \cdot \TD_\xv) f](\xv_t, \yv_t), \, 
\yv_{t+1} = \yv_t - [(\yyv^{-1} \cdot \n_\yv) f](\xv_{t}, \yv_t) - [(\yyv^{-1} \cdot\yxv) f](\xv_{t}, \yv_t)\cdot (\xv_{t+1} - \xv_t).
\en
This is close to the algorithm in \Cref{app:tgd_cn}, if we modify \eqref{eq:newton_tdg} to be:
\be
\xv_{t+1} = \xv_t - [(\TD_{\xv\xv}^{-1} \cdot \TD_\xv) f](\xv_t, \yv_t), \, 
\yv_{t+1} = \yv_t - \yyv^{-1}f(\xv_{t}, \yv_t)\cdot \n_\yv f(\xv_{t+1}, \yv_t),
\en
and use the approximation:
\be
\n_\yv f(\xv_{t+1}, \yv_t) \approx \n_\yv f(\xv_{t}, \yv_t) + \yxv f(\xv_{t}, \yv_t) \cdot (\xv_{t+1} - \xv_t).
\en
However, for the algorithm described by \eqref{eq:evtushenko_cn}, no explicit proof was given and the exact local convergence rate was not shown. Our GDN/CN methods are novel in the sense that: (\textbf{a}) we use alternating updates and the algorithms \eqref{eq:gdn}, \eqref{eq:newton} are much simpler than \citet{evtushenko1974iterative}, especially for CN; (\textbf{b}) we give detailed theorems of the exact \emph{non-asymptotic} convergence rates and the local neighborhoods to initialize, and formally prove the theorems (Theorems~\ref{thm:GDN} and \ref{thm:CN}); (\textbf{c}) we are the first to \emph{efficiently} implement Newton's methods in GAN training problems, using the Newton-CG method.

\end{document}

%% file: sections/2-background.tex

\vspace{-0.3em}
\paragraph{Notations.}\label{sec:notation}
\vspace{-0.3em}
Given a function $f: \R^n \times \R^m \to \R$ that is twice differentiable, with the arguments $\xv\in \R^n$ and $\yv\in \R^m$, we use $\n_\xv f$ and $\n_\yv f$ to denote the partial derivatives of a function $f$ \wrt $\xv$ and $\yv$, respectively, and similarly we use $\xxv f, \xyv f, \yxv f, \yyv f$ for the second order partial derivatives.
When $\yyv f$ is invertible, we use the shorthand $\yyv^{-1}  f := (\yyv f(\cdot))^{-1}$. We define the total derivatives:
\begin{align}
\label{eq:TD1}
\TD_\xv f &:= \n_\xv f -\n_{\xv\yv} f \cdot \n_{\yv\yv}^{-1} f \cdot \n_\yv f,
\\
\label{eq:TD2}
\TD_{\xv\xv} f &:= \xxv f - \xyv f \cdot  \yyv^{-1} f \cdot \yxv f.
\end{align}
We will discuss these total derivatives more carefully in Section \ref{sec:bg} and Appendix \ref{app:local_bound_lipschitz}. 
For convenience, we also define $$\zv := (\xv, \yv).$$ The partial derivative operators can be distributed, e.g., $(\yyv^{-1}\cdot \n_\yv)f := (\yyv f)^{-1}\cdot \n_\yv f$, where the $\cdot$ sign means matrix multiplication. For a function $g: \R^d \to \R$, we use $g'(\xv)$ and $g''(\xv)$ to denote its gradient and Hessian. We use Euclidean norms for vectors and spectral norms for matrices. $\Nc(\wv)$ denotes a neighborhood of $\wv$. We define $a\wedge b := \min\{a, b\}$ and $a\vee b := \max\{a, b\}$. 

\vspace{-0.5em}
\section{Preliminaries} 
\label{sec:bg}
\vspace{-0.3em}

Our main interest is the following nonconvex-nonconcave minimax optimization problem:
\be
\label{eq:minimax}
\min_{\xv \in \R^n} \max_{\yv \in \R^m} f(\xv, \yv),
\en
where $f: \R^{n+m} \to \R$ is twice continuously differentiable, i.e.~$f\in \Cc^2$. We focus on studying the local convergence near the following local optimal solution:

\begin{defn}[\textbf{strict local minimax (\slmm)}, \citet{evtushenko1974some}]
\label{def:sslmm} 
$(\xv^*, \yv^*)$ is a strict local minimax point (\slmm) of a twice differentiable function $f$ if it is a stationary point and
$
\n_{\yv\yv} f(\xv^*, \yv^*) \cl \zero
$ and 
$
\TD_{\xv\xv} f(\xv^*, \yv^*)
\cg {\bf 0}.
$
\end{defn}
\vspace{-0.5em}



Such a definition has also appeared in recent literature \citep{fiez2019convergence, jin2019minmax, wang2019solving}. Its analogy in familiar minimization problems is a non-degenerate minimizer at which the Hessian is strictly positive definite. It is also more general than the strict local Nash equilibrium \cite{fiez2019convergence}, by which we mean 
\be\label{eq:strict_local_Nash}
\n_{\yv\yv} f(\xv^*, \yv^*) \cl \zero\mbox{ and } 
\n_{\xv\xv} f(\xv^*, \yv^*)
\cg {\bf 0}.
\en


\begin{eg}
$(0, 0)$ is a \slmm of the function $f(x, y) =-3 x^2 + x y^2 - y^2 + 4 x y$ on $\R^2$ but not a saddle point, since 
\be
\n_{xx} f(0, 0) = -6, \,  \n_{xy} f(0, 0) = 4,\, \n_{yy} f(0, 0) = -2,\nonumber
\en
and thus $\TD_{xx} f(0, 0) = 2$. 
This function is nonconvex-nonconcave, because $\n_{yy} f(x, y) = 2 x - 2$ is not always negative for $(x, y) \in \R^2$, and $\n_{xx} f(0, 0) < 0$. 
\end{eg}

The practical relevance of {\slmm}s becomes important when training generative adversarial networks (GANs) and distributional robustness models:
\begin{eg}[\textbf{GAN}, \citet{nagarajan2017gradient}]\label{eg:gan}

Consider the following GAN training problem, where we minimize over generator $G$ with parameter $\thetav$ and maximize over discriminator $D$ with parameter $\phiv$: 
\be\label{eq:fgan}
&\min_{\thetav} \max_{\phiv} ~\ell(\thetav, \phiv), \, \text{where}\, \ell(\thetav, \phiv) = \E_{\xv\sim p_\xv} [f(D_{\phiv}(\xv))] +
\E_{\zv\sim p_\zv} [f(-D_{\phiv}(G_{\thetav}(\zv)))]. \nonumber
\en
Under some mild assumptions \citep{nagarajan2017gradient}, at a stationary point the partial Hessians satisfy:
\begin{align}
&\n_{\thetav \thetav} \ell = \zero, \, \n_{\phiv \phiv} \ell = 2f''(0) \E_{\xv\sim p_{\xv}} [\n_{\phiv} D_{\phiv}(\xv) \cdot \n_{\phiv} D_{\phiv}^\top(\xv) ],\, \n_{\thetav \phiv} \ell = -f'(0) \cdot \n_{\thetav} \E_{\xv \sim p_{\zv}}[ \n_{\phiv} G_{\thetav}(D_{\phiv}(\xv))].\nonumber
\end{align}
Typically, $f'(0) \ne 0$ and $f''(0) < 0$. For example, for vanilla GAN \citep{goodfellow2014generative}, $$f(x) = -\log(1 + e^{-x}),$$ giving $f'(0) = \tfrac12$ and $f''(0) = -\tfrac14$. Therefore, under full rank assumptions \citep{nagarajan2017gradient}, $$ \n_{\phiv\phiv} \ell \cl \zero, \TD_{\thetav\thetav} \ell  = (\n_{\thetav\thetav}- \n_{\thetav\phiv}  \cdot \n_{\phiv\phiv}^{-1} \cdot \n_{\thetav\phiv}^\top) \ell\cg \zero,$$ \ie the stationary point is a \slmm. The loss $\ell$ is usually \emph{not} a convex function of the generator parameter $\thetav$.
\end{eg}

\begin{eg}[\textbf{Distributional robustness}, \citet{sinha2018certifiable}]\label{eg:distributional_robustness}
Given $N$ data samples $\{\xiv_i\}_{i=1}^N$, the Wasserstein distributional robustness model can be written as:
\be
\min_{\thetav} \max_{\Omegav} f(\thetav, \Omegav) = \sum_{i=1}^N \ell(\thetav, \omegav_i) - \gamma \|\omegav_i - \xiv_i \|^2,
\en
where we denote $\Omegav = \{\omegav_i\}_{i=1}^N$ as the collection of adversarial samples. Here $\thetav$ are the model parameters, and $\ell$ is the loss function. The goal of this task is to find robust model parameters $\thetav$ against adversarial perturbation of samples, $\omegav_i$.  At a stationary point $(\thetav^*, \Omegav^*)$, \citet{sinha2018certifiable} shows that for large $\gamma$, $\partial_{\Omegav \Omegav} f(\thetav^*, \Omegav^*)$ is negative definite. Moreover, the total Hessian $\TD_{\thetav \thetav} f(\thetav^*, \Omegav^*)$ is:
\be
\sum_{i=1}^N \n_{\thetav \thetav} \ell(\thetav^*, \omegav_i^*) - \Mv_i (\n_{\omegav \omegav} \ell(\thetav^*, \omegav_i^*) - 2\gamma \Iv)^{-1} \Mv_i^\top, 
\en
where $\Mv_i := \n_{\thetav \omegav} \ell(\thetav^*, \omegav_i^*)$. Under assumptions that $\thetav^*$ is a local minimum of the adversarial training loss $\sum_{i=1}^N \ell(\cdot, \omegav_i^*)$ and that $\Mv_i $ is full row rank for at least one adversarial example $\omegav^*_i$, we can show that $(\thetav^*, \Omegav^*)$ is a \slmm for large $\g$. Moreover, $(\thetav^*, \Omegav^*)$ is not necessarily a strict local Nash equilibrium. We provide a detailed proof in Appendix \ref{app:analysis_of_distributional_robustness}. 
\end{eg}


\noindent At a \slmm $(\xv^*, \yv^*)$, from $$\n_{\yv} f(\xv^*, \yv^*) = \zero, \, \n_{\yv\yv} f(\xv^*, \yv^*) \cl \zero$$ and the implicit function theorem, we know that for $f\in \Cc^2$, there are neighborhoods $\Nc(\xv^*) \subset \R^n$,
 $\Nc(\yv^*) \subset \R^m$
and a continuously differentiable function 
\be\label{eq:neighborhoods}
r: \Nc(\xv^*) \to \Nc(\yv^*)\textrm{ s.t. }\n_{\yv} f(\xv, r(\xv)) = \zero
\en
and $r(\xv)$ is a local maximizer of the function $f(\xv, \cdot)$. Also, 
\be\label{eq:deriv_r}
r'(\xv) = - (\yyv^{-1} \cdot \yxv) f(\xv, r(\xv))
\en
for any $\xv \in \Nc(\xv^*)$. We call this function the \emph{local best-response function}. The local best response function leads to our definition of total derivatives. Define the ``local maximum function'' $\psi(\xv) := f(\xv, r(\xv))$ on $\Nc(\xv^*)$, from \eqref{eq:deriv_r} we can derive that (see Lemma \ref{lem:Lipschitz_of_psi_derivs} in Appendix \ref{app:proof}):
\be
\psi'(\xv) = \TD_{\xv} f(\xv, r(\xv)), \, \psi''(\xv) = \TD_{\xv\xv} f(\xv, r(\xv)).
\en
Thus, Definition~\ref{def:sslmm} gives a sufficient condition that $\yv^*$ is a local maximum of $f(\xv^*, \cdot)$ and $\xv^*$ is a local minimum of $\psi(\xv)$.

We now assume the second-order Lipschitz condition on the neighborhood $\Nc(\xv^*) \times \Nc(\yv^*)$ (same as in \eqref{eq:neighborhoods}), which allows us to give the non-asymptotic (local) guarantee of our Newton-type algorithms in Theorems~\ref{thm:GDN} and \ref{thm:CN}.

\begin{assump}[\textbf{Lipschitz Hessian}]\label{asp:lip_hessian}
There exist constants $L_{xx}, L_{xy}, L_{yy}$ such that for any $\xv_1, \xv_2 \in \Nc(\xv^*)$ and $\yv_1, \yv_2 \in \Nc(\yv^*)$, we have
\begin{align}
&\|\xxv f(\zv_1) - \xxv f(\zv_2)\| \leq L_{xx} \|\zv_1 - \zv_2\|, \tr
&\|\xyv f(\zv_1) - \xyv f(\zv_2)\| \leq L_{xy} \|\zv_1 - \zv_2\|,\tr
&\|\yyv f(\zv_1) - \yyv f(\zv_2)\| \leq L_{yy} \|\zv_1 - \zv_2\|,\nonumber
\end{align}
with $\zv_i = (\xv_i, \yv_i)$ for $i = 1, 2$.
\end{assump}

%% file: sections/3-first.tex
\vspace{-0.5em}
\section{Gradient-Descent-Newton (GDN) and related algorithms}\label{sec:gdn}
\vspace{-0.3em}

We propose our first Newton-based algorithm (GDN) for solving the nonconvex-nonconcave minimax problem \eqref{eq:minimax}, and make connections and comparisons to existing algorithms. In the GDA algorithm, the follower takes one gradient ascent step to approximate the best response function. However, such step might be insufficient for the approximation. 
Instead, we use a Newton step to approximate the local best response function $r(\xv)$, which is also more appealing if the inner maximization is ill-conditioned.

Many existing algorithms, including GDN, are based on a classic idea that goes back to Uzawa \citep{arrow1958studies}: we employ iterative algorithms \Fa and \La for the follower $\yv$ and leader $\xv$, respectively. The key is to allow \Fa to \emph{adapt quickly} to the update in \La. Naturally, we propose to apply gradient descent as $\La$ and Newton update as $\Fa$:
\begin{align}
\label{eq:gdn}
\vspace{-1.2em}
\begin{split}
&\xv_{t+1} = \xv_t - \a_{\La}  \cdot \n_\xv f(\xv_t, \yv_t), \\
&\yv_{t+1} = \yv_t - (\n_{\yv\yv}^{-1} \cdot \n_\yv) f(\xv_{t+1}, \yv_t),
\end{split}
\end{align}
Newton's method is affine invariant \citep[Section 9.5.1]{boyd2004convex}: under any invertible affine transformation, Newton's update remains essentially the same while gradient updates change drastically. Thus, for ill-conditioned follower problems (where the largest and smallest eigenvalues of $\n_{\yv\yv} f(\xv^*, \yv^*)$ differ significantly), we expect Newton's algorithm to behave well while gradient algorithms will largely depend on the condition number.


\paragraph{Newton-CG method.} Efficient implementations of Newton's algorithm have been actively explored in deep learning since \citet{Martens10}. The product $(\n_{\yv\yv}^{-1} \cdot \n_\yv) f$ can be efficiently computed using conjugate gradient (CG) equipped with Hessian-vector products computed by autodiff. Complexity analysis of the Newton-CG method can be found in \citet{royer2020newton} and references therein.

\paragraph{Extensions} We can further stabilize GDN by employing a damping factor or suitable regularization and approximation \citep{Martens10} (see also Appendix~\ref{app:damp_reg}). To accelerate GDN, we can also add momentum or replace the partial derivative $\n_\xv f$ with the total derivative $\TD_\xv f$, as discussed in Appendix~\ref{app:algmod}. Generalization to general sum games can be found in Appendix~\ref{app:gen_sum}. 

We now present the \emph{non-asymptotic} local linear convergence rate of GDN to a \slmm. Note that the neighborhoods $\Nc(\xv^*)$ and $\Nc(\yv^*)$ are the same as in \eqref{eq:neighborhoods}. 
\begin{restatable}[\textbf{GD-Newton}]{thm}{thmgdn}
\label{thm:GDN}
Given a {\slmm} $(\xv^*, \yv^*)$, suppose on the neighborhoods $\Nc(\xv^*)\times \Nc(\yv^*)$, Assumption \ref{asp:lip_hessian} holds and $\mu_x \Iv \cle \TD_{\xv\xv} f(\xv, \yv) \cle M_x \Iv$ for any $(\xv, \yv) \in \Nc(\xv^*)\times \Nc(\yv^*)$. Define:
\begin{align}\label{ineq:init_gdn}
&\Nc_{\rm GDN} := \{(\xv, \yv) \in  \Nc(\xv^*) \times \Nc(\yv^*): \, \|\xv - \xv^*\| \leq \d := \min\left\{\d_0, \frac{\e}{\a_\La M (1 + 4 V^2/\rho_{\La}^2)}\right\}, \, \|\yv - \yv^*\| \leq 2V \d \},
\end{align}
where $\d_0$, $V$, $M$ are absolute constants, $$\rho_{\La} =  |1 - \a_\La \mu_x|\vee |1 - \a_\La M_x|\mbox{ and }0 < \e < 1 - \rho_{\La}.$$ 
Given an initialization $(\xv_1, \yv_1) \in \Nc_{\rm GDN}$, $\|\yv_1 - \yv^*\| \leq 2V \|\xv_1 - \xv^*\|$ and suppose that $(\xv_2, \yv_2) \in  \Nc_{\rm GDN}$ and $\|\yv_2 - \yv^*\| \leq 2V \|\xv_2 - \xv^*\|$, the convergence of the GD-Newton to $(\xv^*, \yv^*)$ is linear, \ie, for any $t \geq 2$, we have:
\begin{align}\label{ineq:linear_gdn}
\begin{split}
&\|\xv_{t + 1} - \xv^*\|\leq (\rho_\La + \e)^{t-1} \| \xv_2 - \xv^*\|, \, \|\yv_{t + 1} - \yv^*\| \leq 2V (\rho_\La + \e)^{t-1} \| \xv_2 - \xv^*\|.
\end{split}
\end{align}
\end{restatable}

\vspace{-0.3em}
The exact values of $\d_0$, $V$, $M$ depend on the local information of function $f$ on the neighborhood $\Nc(\xv^*) \times \Nc(\yv^*)$ which we will defer to {Appendix}~\ref{app:proof}.

Our result is on the local convergence, and we need a good initialization that is close to the \slmm (see more detail in Appendix \ref{app:Newton_type_proof}).  
To obtain a good initialization, in practice we consider the method of \emph{pre-training and fine-tuning} \citep{hinton2006reducing}, which we will discuss more at the end of \S\ref{sec:second} and implement in \S\ref{sec:exp}. 


The definition of $\d$ in
\eqref{ineq:init_gdn} tells us that when $\e$ is small, the second term dominates. This means that if we want a better local convergence rate we need to be closer to the \slmm. Also, a smaller $\a_\La$ can control the neighborhood and thus GDN becomes more stable. 

\paragraph{Proof techniques} Our proof relies on two parts: the leader takes gradient descent on $\psi(\xv)$ with approximation error controlled by $r(\xv_t) - \yv_t$; the follower takes Newton updates to approximate the local best response $r(\xv_t)$ at each step. This reflects the sequential nature of the minimax game. The difficulty lies in how to bound the approximation errors.

In fact, GDN is an approximation of Uzawa's approach:
\begin{align}
\label{eq:Uzawa}
\xv_{t+1} = \xv_t - \a_\La \cdot \n_{\xv}f(\xv_t, \yv_t),\,  \yv_{t+1} = r(\xv_{t+1}),
\end{align}
where recall that $r$ is the local best response. The update \eqref{eq:Uzawa} is essentially the original proposal by Uzawa \citep{arrow1958studies} as also in e.g.~\citet[][Sec.~3.1]{fiez2019convergence} and \citet[][Sec.~4]{jin2019minmax} for different settings. In Theorem \ref{thm:ugda} we will see another approach to approximate \eqref{eq:Uzawa}.


As expected, the condition number of the follower Hessian $\n_{\yv\yv} f$ has no effect on the local convergence rate of GDN thanks to the Newton update on $\yv$. When $\a_\La = 2/(\mu_x + M_x),$ $\rho_\La$ is minimized to be 
$$
\rho_\La = \frac{\kappa_\La - 1}{\kappa_\La + 1}\mbox{ where }\kappa_\La = M_x/\mu_x\mbox{ is the condition number.}
$$
The condition number $\kappa_{\La}$ of the leader problem does appear, since GDN still employs a gradient update for the leader $\xv$. We will see how to remove this dependence in \S\ref{sec:second}. 

To fully appreciate our method, in the following subsections, we make comparisons between GDN and existing alternative algorithms and reveal interesting connections. Note that in Theorem~\ref{thm:GDN}, if we take $\e \to 0$, we obtain an asymptotic (local) linear convergence rate $\rho_\La$: 
\be\label{eq:rho_la_gdn}
\rho_\La = |1 - \a_{\La} \lambda_1| \vee |1 - \a_{\La} \lambda_n|,
\en
with $\l_1$ and $\l_n$ being the largest and smallest eigenvalues of $\TD_{\xv\xv} f(\zv^*)$. In general, the asymptotic linear convergence rate can be obtained by performing spectral analysis on the Jacobian of the iterative update \cite{polyak87book}. For simplicity, we will use this tool to compare with other algorithms. 

\vspace{-0.3em}
\subsection{Comparing with Total Gradient Descent Ascent (TGDA) and its variants}\label{sec:tgda_fr_main}

We compare GDN with two methods in this section: Total Gradient Descent Ascent (TGDA) and Follow-the-Ridge (FR). They both involve $\yyv^{-1} f$ and have similar computation complexity as GDN. Their convergence is also \emph{local} without adding more assumptions to our paper. 

TGDA takes a GA step for the follower and a \emph{total} gradient ascent step for the leader:
\begin{align}
\label{eq:fiezt_main}
\begin{split}
&\xv_{t+1} = \xv_t - \a_{\La} \cdot \TD_\xv f (\xv_t, \yv_t), \, \yv_{t+1} = \yv_t + \a_{\Fa} \cdot \n_\yv f(\xv_t, \yv_t),
\end{split}
\end{align}
where we use the total gradient $\TD_\xv$ instead of the partial derivative $\n_{\xv}$ for the update on the leader $\xv$. Its continuous dynamics was originally studied in \citet{evtushenko1974iterative} with linear convergence proved (see \Cref{app:related}). More recently, the stochastic setting and the two-time-scale variant are studied in \citet{fiez2019convergence} for general sum games. A similar algorithm to TGDA is Follow-the-Ridge (FR) \citep{evtushenko1974iterative, wang2019solving}.
In fact, the two algorithms amount to performing some pre-conditioning on GDA, and their preconditioning operators are simply transpose of each other (see \Cref{sec:tgda_fr}).
It follows that TGDA and FR have the same Jacobian spectrum at a \slmm. 

We now present their asymptotic local convergence:

\begin{restatable}[]{thm}{tgdaFr}\label{thm:tgda_fr}
 TGDA and FR achieve the same asymptotic linear convergence rate $\rho = \rho_{\La} \vee \rho_{\Fa}$ at a {\slmm} $(\xv^*,\yv^*)$,
\begin{align}
\mbox{ where }\rho_{\La} = |1 - \a_{\La} \l_1| \vee |1 - \a_{\La} \l_n|\,\mbox{ and }\rho_{\Fa} = |1 - \a_{\Fa} \mu_1| \vee |1 - \a_{\Fa} \mu_m|,\nonumber
\end{align}
with $\l_1$ and $\l_n$ (resp.~$\mu_1$ and $\mu_m$) being the largest and smallest eigenvalue of $\TD_{\xv\xv}f(\xv^*, \yv^*)$ (resp.~of $-\yyv f(\xv^*,\yv^*)$). 
\end{restatable}

Note that by an asymptotic linear rate $\rho$ we meant $\rho = \limsup_{t\to\infty} {\|\zv_{t+1} - \zv^*\|}/{\|\zv_t - \zv^*\|}.$ Choosing $\alpha_{\La} = 2/(\lambda_1 + \lambda_n)$, $\alpha_{\Fa} = 2/(\mu_1 + \mu_m)$ gives the optimal convergence rate $$ \frac{\kappa_{\La}-1}{\kappa_{\La}+1} \vee \frac{\kappa_{\Fa}-1}{\kappa_{\Fa}+1},$$ where $\kappa_{\La} := \lambda_1/\lambda_n$ and $\kappa_{\Fa} := \mu_1/\mu_m$. A slightly weaker result for FR
has appeared in \citet{wang2019solving}. Compared to Thm.~\ref{thm:GDN}, the local convergence of GDN is always faster, especially when the follower problem is ill-conditioned (\ie when $\kappa_{\Fa}$ is large compared to $\kappa_{\La}$), a point that we will verify in our experiments. 

\vspace{-0.3em}
\subsection{Comparing with Gradient Descent Ascent (GDA) and its variants} \label{sec:gda_main}
\paragraph{Two-time-scale GDA.}

One of the first algorithms for the minimax problem \eqref{eq:minimax} is  gradient-descent-ascent (GDA) \citep{arrow1958studies}, where we adopt GD as \La for updating the leader while we use GA as \Fa for updating the follower:
\begin{align}
\label{eq:ttsgd}
\begin{split}
&\xv_{t+1} = \xv_t - \a_{\La} \cdot \n_\xv f(\xv_t, \yv_t),\, \yv_{t+1} = \yv_t + \a_{\Fa} \cdot \n_\yv f(\xv_t, \yv_t).
\end{split}
\end{align}
We consider two different scales of the step sizes \citep{heusel2017gans, jin2019minmax}, \ie $\a_{\La} = o(\a_{\Fa})$, as is typical in stochastic approximation \citep{Borkar08}, to converge linearly at a {\slmm}. However, in practice two-time-scale GDA (2TS-GDA) is hard to tune, especially when the follower problem is ill-conditioned, as we will verify in our experiments below. 
2TS-GDA (locally) converges slower than TGDA and FR, hence also slower than GDN (see App.~\ref{app:grad_alg}).

\paragraph{$k$-step gradient descent ascent.}
We also compare GDN with $k$-step gradient descent ascent (GDA-$k$) as proposed in \citet{goodfellow2014generative}. After each GD update on the leader, GDA-$k$ performs $k$ GA updates on the follower:
\be
&\xv_{t+1} = \xv_t - \a \cdot \n_{\xv}f(\xv_t, \yv_t), \, \yv_{t+1} = g^{(k)}(\yv_t)\mbox{ with }g(\yv) = \yv + \a \cdot \n_{\yv}f(\xv_{t+1}, \yv),\nonumber
\en
where $g^{(k)}$ means composition for $k$ times. Letting $k\to \infty$ amounts to solving the follower problem exactly by gradient ascent steps (see \eqref{eq:Uzawa}). Continuing with the notation in Thm.~\ref{thm:tgda_fr}, we derive the following result:

\begin{restatable}[\textbf{GDA-$\infty$}]{thm}{Ugda}\label{thm:ugda}
 GDA-$k$ achieves an asymptotic linear convergence rate $$\rho_{\La} = |1 - \a \l_1| \vee  |1 - \a \l_n|$$ at a \slmm $(\xv^*, \yv^*)$ when $k \to \infty$ and $\a < 2/\mu_1$. If $\mu_1 < \l_1 + \l_n$, choosing $\a = 2/(\l_1+\l_n)$ we obtain the optimal convergence rate $$\frac{\kappa_{\La}-1}{\kappa_{\La} + 1},$$  otherwise with $\a$ approaching $2/\mu_1$ we obtain a suboptimal rate $1-2\l_n/\mu_1$.
\end{restatable}
Comparing Thm.~\ref{thm:ugda} with \eqref{eq:rho_la_gdn}, we find that GDN and GDA-$\infty$ share the same local convergence rate, confirming that when sufficiently close to an optimum, a single Newton step is as good as solving the problem exactly.
When $\mu_1$ is large (meaning the follower problem has a sharp curvature), we have to use a small step size $\a$ for updating the leader, and the resulting rate can be slower than GDN. 
Similar to 2TS-GDA, it is hard to gauge how many GA steps we need to approximate the exact algorithm \eqref{eq:Uzawa} sufficiently well. When the follower problem is ill-conditioned, the number of GA steps may grow excessively large and we have to use a small step size $\a$ to ensure convergence. The two-time-scale modification of GDA-$k$ can be found in \eqref{eq:gdak_2ts}.

%% file: sections/4-second.tex


\vspace{-0.3em}
\section{Complete Newton (CN)}
\vspace{-0.3em}
\label{sec:second}
Although our first Newton-type algorithm, GDN, evades possible ill-conditioning of the follower problem, it may still converge slowly if the leader problem is ill-conditioned, \ie, the largest and the smallest eigenvalues of $\dxxv f$ differ significantly. We propose a new Newton-type algorithm that evades ill-conditioning of both leader and follower problems, and locally converges super-linearly to a \slmm. With total second-order derivatives, we replace the gradient update of the leader in GDN with a Newton update, which we call the \emph{Complete Newton} (CN) method:
\begin{align}\label{eq:newton}
\vspace{-0.3em}
\begin{split}
&\xv_{t+1} = \xv_t - (\TD_{\xv\xv}^{-1} \cdot \n_\xv) f(\xv_t, \yv_t), \, \yv_{t+1} = \yv_t - (\yyv^{-1} \cdot \n_\yv) f(\xv_{t+1}, \yv_t).
\end{split}
\end{align}
CN is a \emph{genuine} second-order method that (we prove below) achieves a super-linear rate, as compared to other methods in Section~\ref{sec:tgda_fr_main} that use the Hessian inverse.
The Newton update $\dxxv^{-1} f\cdot \partial_{\xv} f = \left(\partial_{\xv\xv} - \partial_{\xv\yv} \partial_{\yv\yv}^{-1} \partial_{\yv\xv}\right)^{-1} f\cdot \partial_{\xv} f$ can be efficiently implemented as solving a single linear system of size $(m + n) \times (m + n)$ (see also Lemma~\ref{lma:inverse_schur_complement} in Appendix~\ref{app:exp}):
\begin{align}
&\begin{bmatrix}
\n_{\xv\xv} f& \n_{\xv\yv}  f\\ \yxv f & \yyv f
\end{bmatrix}
\begin{bmatrix}
\Delta\xv \\ \Delta \vv
\end{bmatrix}
=
\begin{bmatrix}
\n_{\xv} f \\ \zero
\end{bmatrix}
\iff \tr
& \Delta \xv = 
\begin{bmatrix}
\Iv & \zero
\end{bmatrix}
\begin{bmatrix}
\n_{\xv\xv} f& \n_{\xv\yv} f\\ \yxv f& \yyv f
\end{bmatrix}^{-1} 
\begin{bmatrix}
\n_{\xv} f \\ \zero
\end{bmatrix}
= (\TD_{\xv\xv}^{-1} \cdot \n_{\xv})f.\nonumber
\end{align}
As a result, when $m \approx n$, CN has the same complexity as TGDA, FR and GDN, which all use second order information (\Cref{tbl:compare}).
However, only CN enjoys the following local \emph{quadratic} convergence rate:

\begin{restatable}[\textbf{Complete Newton}]{thm}{thmcn}
\label{thm:CN}
Given a \slmm $\zv^*:= (\xv^*, \yv^*)$, suppose in the neighborhood $\Nc(\xv^*) \times \Nc(\yv^*)$, Assumption \ref{asp:lip_hessian} holds.
Let $L$ be an absolute constant and $\Nc_{\rm CN} \subset \Nc(\xv^*) \times \Nc(\yv^*)$ be a neighborhood of $\zv^*$.
The local convergence of CN is at least quadratic, i.e.:
\be
 \|\zv_t - \zv^*\| \leq \frac{1}{2L} \left(2L(\|\zv_1 - \zv^*\|\vee \|\zv_2 - \zv^*\|)\right)^{2^{\lfloor (t-1)/2 \rfloor}}\nonumber
\en
with the initializations $\zv_1 \in \Nc_{\rm CN}$ and $\zv_2 \in \Nc_{\rm CN}$.
\end{restatable}

In Thm.~\ref{thm:CN}, 
the exact form of $L$ and $\Nc_{\rm CN}$ can be found in its more detailed version, Thm.~\ref{thm:CN2} in App.~\ref{app:proof}. The local super-linear convergence of CN means that this method is not heavily affected by the ill-conditioning of either the leader or the follower problem, when the initialization is close to the \slmm $\zv^*$. To obtain a good initialization, we consider the following method of \emph{pre-training and fine-tuning}:

\paragraph{Pre-training and fine-tuning.} We point out the sensitivity to initialization of our Newton-based algorithms: CN and GDN require the initialization to be close to the optimal solution, similar to the conventional Newton algorithm for minimization \citep{bertsekas1997nonlinear}.
Fortunately, we can employ a ``pre-training + fine-tuning'' approach \citep{hinton2006reducing}. For instance, we may run GDA for the initial phases, even though GDA is slowed down by the ill-conditioning, as soon as it goes in the neighborhood where Newton-type algorithms have convergence guarantees, we can switch to CN or GDN to converge quickly and to evade ill-conditioning, as we will show in \Cref{sec:exp}.

%% file: sections/5-exp.tex
\vspace{-0.3em}
\section{Experiments}\label{sec:exp}
\vspace{-0.3em}

\begin{figure*}[t]
\label{fig:linear}
\centering
\begin{subfigure}{0.48\textwidth}
    \includegraphics[width=\textwidth]{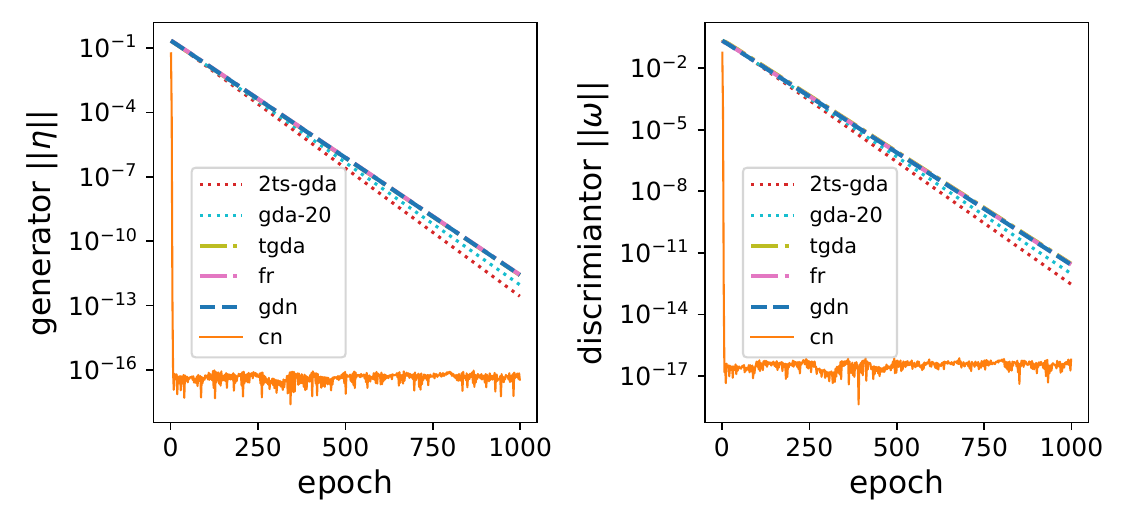}
\caption{well-conditioned Gaussian mean estimation}
\label{fig:estimating_mean_well_conditioned}
\end{subfigure}
\begin{subfigure}{0.48\textwidth}
    \includegraphics[width=\textwidth]{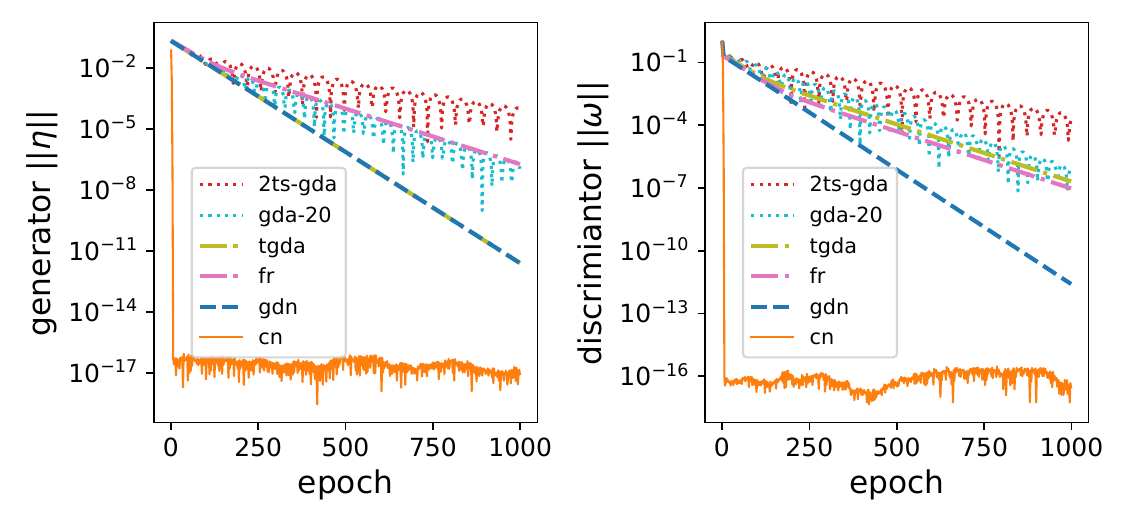}
\caption{ill-conditioned Gaussian mean estimation}
\label{fig:estimating_mean_ill_conditioned}
\end{subfigure}
\begin{subfigure}{0.48\linewidth}
    \includegraphics[width=\textwidth]{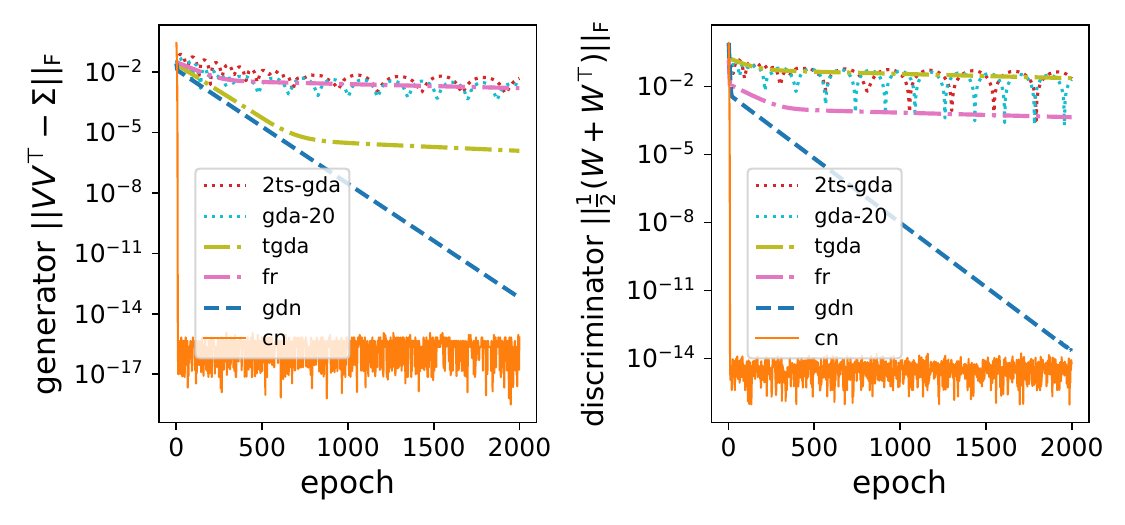}
\caption{ill-conditioned Gaussian covariance estimation}
\label{fig:estimating_covariance}
\end{subfigure}
\begin{subfigure}{0.48\textwidth}
\centering
\includegraphics[width=\linewidth]{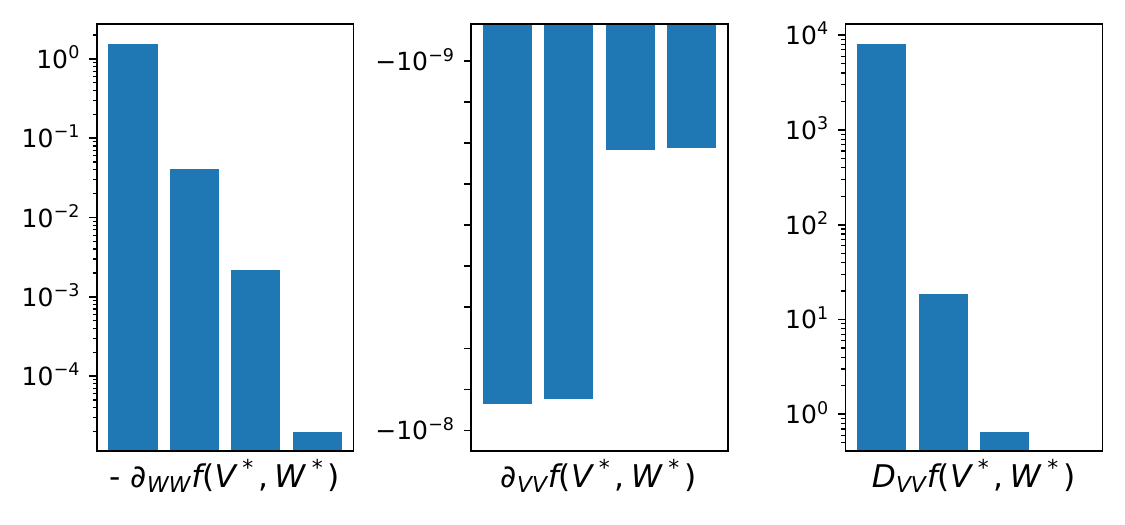}
\caption{eigenvalues of $-\partial_{\Wv\Wv}$, $\partial_{\Vv\Vv}$ and $\TD_{\Vv\Vv}$ at the \slmm.}
\label{fig:eigen_values}
\end{subfigure}
\caption{Convergence on learning Gaussian distributions using JS-GAN.
\textbf{Top:} Estimating the mean of a Gaussian.
We compare the convergence rate in a well-conditioned and an ill-conditioned setting, and plot the norm of the generator and the discriminator respectively.
\textbf{Bottom:} Estimating the covariance of a Gaussian.
We plot the convergence behavour of different algorithms and the eigenvalues at the SLmM.
In both cases, CN quickly reaches the \emph{precision limit} of double precision floating point numbers.
} 
\vspace{-0.5em}
\label{fig:learning_mean}
\end{figure*}

Our numerical experiments confirm:
\begin{itemize}
\item The concept of strict local minimax is applicable in GAN training and ill-conditioned problems may arise even when learning simple distributions using GANs;
\item Newton's algorithms can address the ill-conditioning problem and achieve much faster local convergence rate while keeping similar running time with existing algorithms such as GDA-$k$, TGDA and FR. 
\end{itemize}
All our experiments in this section are run on an {\tt Intel i9-7940X CPU} and a \texttt{NVIDIA TITAN~V GPU}.
Further experimental settings (e.g., step sizes, network architectures, initializations) and results are deferred to Appendix~\ref{app:exp}.
\paragraph{Learning a Gaussian distribution.} 
Consider learning a Gaussian distribution $\xv \sim \Nc(\muv, \Sigmav)$ using a JS-GAN \citep{goodfellow2014generative},
where the latent variable $\zv$ follows a standard Gaussian. 
First, we estimate the mean $\muv$ with two different covariance matrices: a well-conditioned covariance $\Sigmav = \Iv$ and an ill-conditioned covariance $\Sigmav = \diag(1, 0.05)$.
We use a discriminator $D(\xv)$ and a generator $G(\zv)$, s.t.~
\be
 D(\xv) = \sigma\left(\omegav^\top \xv \right), \, G(\zv) = \zv + \etav 
\en
The corresponding GAN training problems are not convex-concave, yet the optimal solutions are {\slmm}s (see \Cref{app:exp}). Comparison among algorithms are presented in \Cref{fig:estimating_mean_well_conditioned,fig:estimating_mean_ill_conditioned}.
While the convergence rates for most algorithms on the well-conditioned Gaussian are similar, all existing methods severely slow down on the ill-conditioned Gaussian.\footnote{We emphasize that adaptive gradient methods cannot handle ill-conditioning either. See \Cref{app:adaptive} for details.}
Only Newton-type methods retain their fast convergence, confirming our theory that they can cope with ill-conditioned problems.
In particular, in both cases CN converges to a high precision solution only in a few iterations, verifying its superlinear convergence rate.

\begin{figure*}[t]
\centering

\begin{subfigure}{0.19\textwidth}
\begin{subfigure}{\textwidth}
    \includegraphics[width=\textwidth]{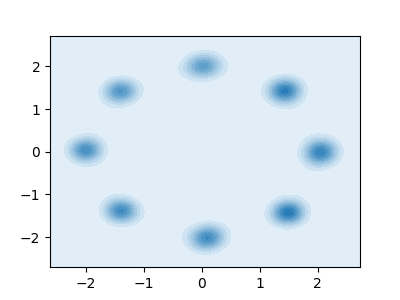}
\end{subfigure}

\begin{subfigure}{\textwidth}
    \includegraphics[width=\textwidth]{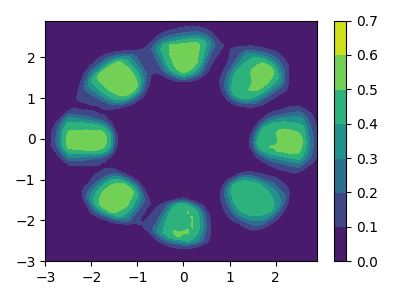}
\end{subfigure}
\caption{TGDA}
\end{subfigure}
\begin{subfigure}{0.19\textwidth}
\begin{subfigure}{\textwidth}
    \includegraphics[width=\textwidth]{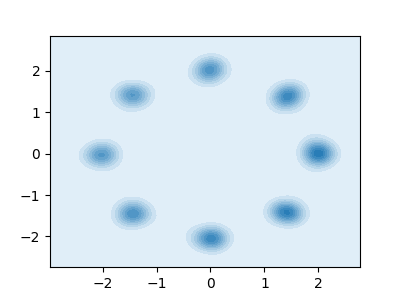}
\end{subfigure}
\begin{subfigure}{\textwidth}
    \includegraphics[width=\textwidth]{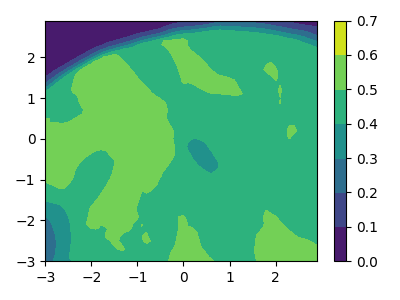}
\end{subfigure}
\caption{FR}
\end{subfigure}
\begin{subfigure}{0.19\textwidth}
\begin{subfigure}{\textwidth}
    \includegraphics[width=\textwidth]{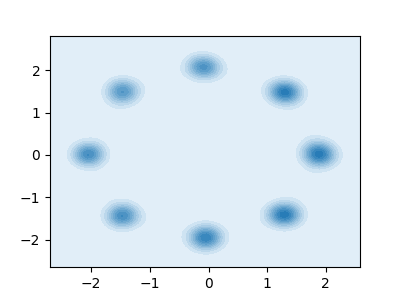}
\end{subfigure}
\begin{subfigure}{\textwidth}
    \includegraphics[width=\textwidth]{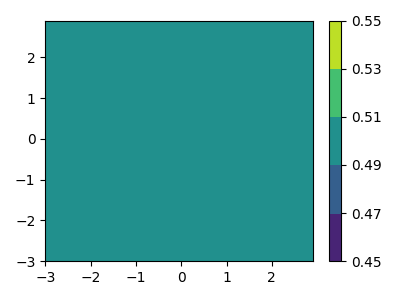}
\end{subfigure}
\caption{GD-Newton}
\end{subfigure}
\begin{subfigure}{0.19\textwidth}
\centering
\begin{subfigure}{\textwidth}
    \includegraphics[width=\textwidth]{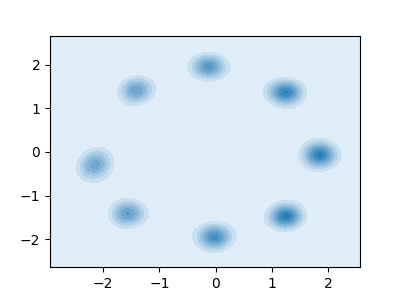}
\end{subfigure}

\begin{subfigure}{\textwidth}
    \includegraphics[width=\textwidth]{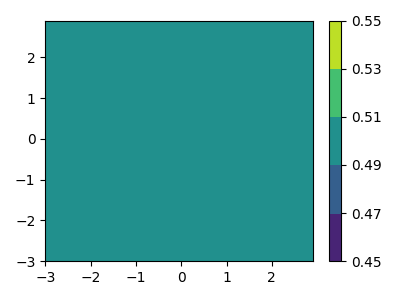}
\end{subfigure}
\caption{CN}
\end{subfigure}
\begin{subfigure}{0.19\textwidth}
\begin{subfigure}{\textwidth}
    \includegraphics[width=\textwidth]{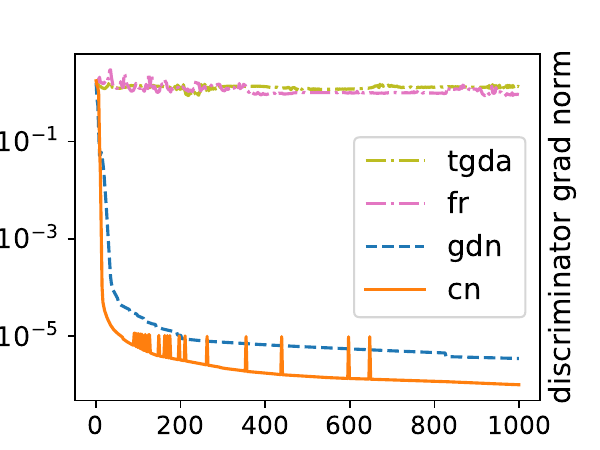}
\end{subfigure}
\begin{subfigure}{\textwidth}
    \includegraphics[width=\textwidth]{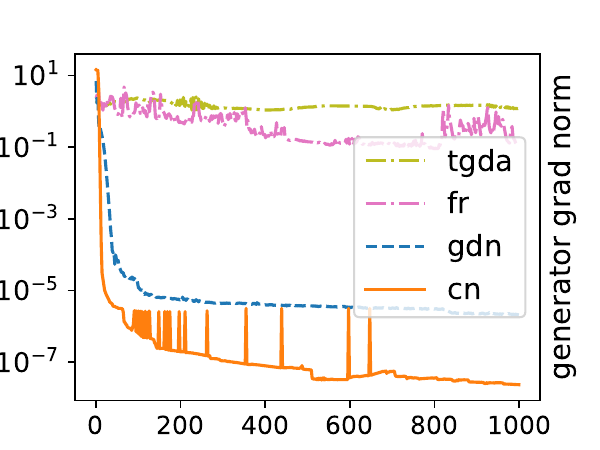}
\end{subfigure}
\caption{Grad Norm}
\end{subfigure}
\caption{Convergence on a mixture of 8 Gaussians.
\textbf{Top:} samples from generator. \textbf{Bottom:} discriminator prediction.
\textbf{Last column:} gradient norms during training. The x-axis is epoch.}
\label{fig:gmm}
\end{figure*}

 Second, we estimate an ill-conditioned covariance $\Sigmav = \diag(1, 0.04)$ with a fixed mean $\muv = \mathbf{0}$.
We use a discriminator $D(\xv)$ and a generator $G(\zv)$ s.t.~
\be
D(\xv) = \sigma\left( \xv^\top \Wv \xv \right), \, G(\zv) = \Vv \zv.
\en
We plot the eigenvalues at the optimal solution in \Cref{fig:eigen_values}:
\begin{itemize}
\item the solution here is almost a SLmM, as the total derivative $\TD_{\Vv\Vv}$ is approximately positive definite (the only negative eigenvalue is on the order of $10^{-9}$), and $\partial_{\Wv\Wv}$ is negative definite;
\item the problem is ill-conditioned, as the condition number of $\n_{\Wv\Wv}$ is greater than $10^4$.
\end{itemize}
Because of the poor conditioning, we observe again that GDA and TGDA/FR severely slow down, while only GDN and CN can retain their fast convergence rate (\Cref{fig:estimating_covariance}).
In particular, CN converges superlinearly and reach the precision limit of floating numbers in only a few iterations.
Note that the solution is not a saddle point, as $\partial_{\Vv\Vv}$ in \Cref{fig:eigen_values} is negative definite.
Thus algorithms for strongly-convex-strongly-concave functions may not work. 

From Thm.~\ref{thm:tgda_fr}, TGDA and FR have the same convergence rate since their preconditioners on GDA are transpose of each other (App.~\ref{sec:tgda_fr}). 
The convergence behaviours of the leader and the follower slightly differ: TGDA converges faster on the generator while FR converges faster on the discriminator. 


\begin{figure*}[t]
\centering
\vspace{-0.5em}
\begin{subfigure}{0.18\textwidth}
\includegraphics[width=\textwidth]{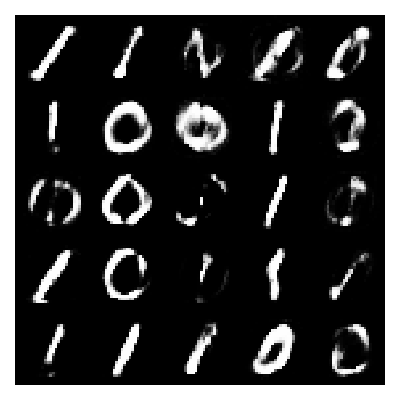}
\caption{GDA-20}
\end{subfigure}
\begin{subfigure}{0.18\textwidth}
\includegraphics[width=\textwidth]{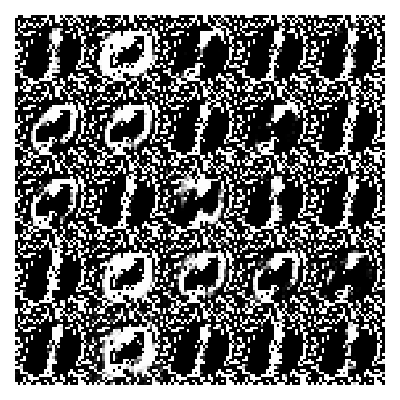}
\caption{TGDA}
\end{subfigure}
\begin{subfigure}{0.18\textwidth}
\includegraphics[width=\textwidth]{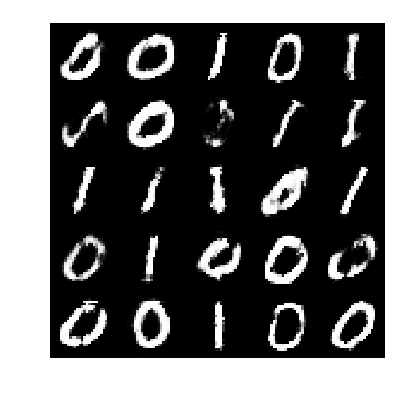}
\caption{FR}
\end{subfigure}
\begin{subfigure}{0.18\textwidth}
    \includegraphics[width=\textwidth]{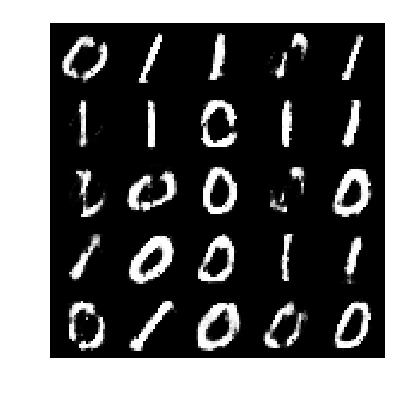}
\caption{\textbf{GDN}}
\end{subfigure}
\begin{subfigure}{0.18\textwidth}
\includegraphics[width=\textwidth]{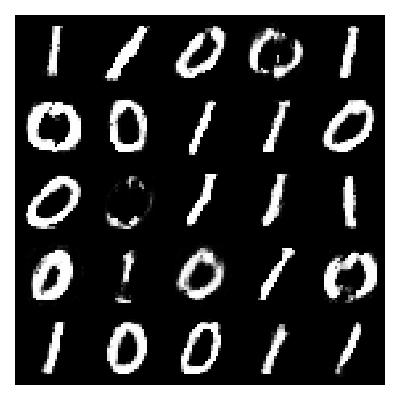}
\caption{\textbf{CN}}
\end{subfigure}
\vspace{-0.5em}

\caption{Digits generated by different algorithms on MNIST 0/1 subset.}
\label{fig:mnist}
\end{figure*}

\paragraph{Mixture of Gaussians.}
We learn a mixture of Gaussians using JS-GAN in \Cref{fig:gmm} (see details in App.~\ref{app:gmm}). 
To inspect local convergence, we first run GDA and use its output as initialization.
We plot the distribution learned by the generator, the discriminator prediction, and gradient norms during training. 
The discriminator trained by GDN/CN is totally fooled by the generator, predicting constant $\frac12$ almost everywhere, and the gradient norms shrink quickly after a few epochs.
In contrast, the gradient norms of TGDA and FR decrease, if at all, very slowly.

Although this is a two dimensional example, the minimax optimization problem has several hundred thousand variables since the generator and the discriminator are deep networks, demonstrating the moderate scalability of Newton-type algorithms to high dimensional problems.

\paragraph{MNIST.} 
We compare different algorithms for generating digits on the 0/1 MNIST subset.
We use Wasserstein GAN \citep{arjovsky2017wasserstein} to learn the distribution, with 2-hidden-layer MLPs ($512$ neurons for each hidden layer) for both the generator and the discriminator, and we impose spectral normalization \citep{miyato2018spectral} on the discriminator.
We first run GDA, which is oscillating on a neighborhood, and use its output as initialization. 
We compare the per epoch running time of different algorithms in \Cref{tab:running_time}.
TGDA, FR, GDN and CN have similar running time as they solve linear systems of similar sizes in their updates. Since we choose a small number of CG iterations (\texttt{max\_iteration = 16} for the discriminator and \texttt{max\_iteration = 8} for the generator), they have similar running time as GDA-$20$, as predicted by our \Cref{tbl:compare}. 

\begin{table}
\centering
\footnotesize
\caption{Running time per epoch on MNIST.}
\label{tab:running_time}
\begin{tabular}{c c c c c c}
\toprule
method &  GDA-$20$  & TGDA & FR & GDN & CN \\
\midrule
time (in sec) &  $2.78$ & $6.08$ & $6.22$ & $4.46$ & $7.04$ \\
\bottomrule
\end{tabular}
\vspace{-0.5em}
\end{table}
\vspace{-0.3em}

\begin{figure}[ht]
\centering
\includegraphics[width=0.8\linewidth]{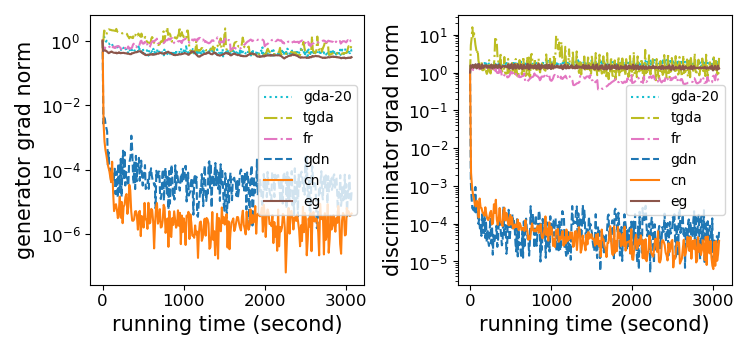}
\caption{Gradient norms on MNIST 0/1 subset.}
\label{fig:grad_mnist}
\vspace{-1em}
\end{figure}

Even though all of our algorithms have similar running time,
we find the convergence speeds are quite different.
We plot the change of their gradient norms with respect to the running time in \Cref{fig:grad_mnist}, where we also compared with the method of extra-gradient (EG, \citet{korpelevich1976extragradient}). 
TGDA/GDA-$20$/FR do not converge or converge quite slowly.
In contrast, GDN converges much faster than all these algorithms above with the same step sizes, as predicted by Theorems~\ref{thm:GDN}, \ref{thm:tgda_fr} and \ref{thm:ugda}.
The convergence speed can be further improved by CN, where the gradient norms diminish faster even if we take a small number of CG iterations.
We plot the digits learned by these algorithms in Figure~\ref{fig:mnist}.
It can be seen that GDN/CN generate high-quality digits that are as good as, if not better than, other optimizers. 


%% file: sections/6-conclusion.tex
\vspace{-0.3em}
\section{Conclusions}
\vspace{-0.3em}

In this work, we developed two Newton-type algorithms for local convergence of \emph{nonconvex-nonconcave} minimax optimization which have wide applications in, e.g., GAN training and adversarial robustness. 
Our algorithms 
\begin{itemize}
    \item share the same computational complexity as existing alternatives that explore second-order information;
    \item have much faster local convergence, especially for ill-conditioned problems.
\end{itemize}  

Experiments show that our algorithms cope with the ill-conditioning that arises from practical GAN training problems. Since we only study the local convergence of Newton-type methods, we consider them as a strategy to ``fine-tune'' the solution and accelerate the local convergence, after finding a good initialization or pre-training with other methods, such as GDA or damped Newton. How to use second-order information to obtain fast global convergence to local optimal solutions in nonconvex minimax optimization with theoretical guarantees remains an important problem.

%% file: appendices/B-proofs.tex

\section{Analysis of Example \ref{eg:distributional_robustness}}\label{app:analysis_of_distributional_robustness}

In this appendix we give a detailed analysis of Example \ref{eg:distributional_robustness}. We prove the following statement:
\begin{prop}
Suppose $(\thetav^*, \Omegav^*) = (\thetav^*, \omegav_1^*, \dots, \omegav_N^*)$ is a stationary point of
\be\label{eq:def_f_distributional_robust}
f(\thetav, \Omegav) = \sum_{i=1}^N \ell(\thetav, \omegav_i) - \gamma \|\omegav_i - \xiv_i \|^2,
\en
where $\ell$ is twice differentiable.
If at this point, $\thetav^*$ is a local minimum of 
$\sum_{i=1}^N \ell(\cdot, \omegav_i^*)$ and there exists at least an adversarial sample $\omegav_i^*$ such that
\be
\Mv_i = \n_{\thetav \omegav} \ell(\thetav^*, \omegav_i^*)
\en
is full row rank, and 
\be\label{eq:assumption_on_gamma}
\gamma > \frac{1}{2} \max_{i=1, \cdots, N}  \lambda_{\max}(\n_{\omegav \omegav} \ell(\thetav^*, \omegav_i^*)), 
\en
with $\lambda_{\max}(\cdot)$ being the largest eigenvalue of a matrix, then $(\thetav^*, \Omegav^*)$ is a \slmm of $f$ but not necessarily a strict local Nash equilibrium.  
\end{prop}

Before we move on to the proof, let us first interpret the stationary point. Solving the condition that:
\be
\n_{\thetav} f(\thetav^*, \Omegav^*) = \sum_{i=1}^N \n_{\thetav} \ell(\thetav^*, \omegav_i^*) = \zero, \, \n_{\omegav_i} \ell(\thetav^*, \omegav_i^*) - 2\gamma (\omegav_i - \xiv_i) = \zero,
\en
i.e., 
\be
\sum_{i=1}^N \n_{\thetav} \ell(\thetav^*, \omegav_i^*) = \zero, \; \omegav_i^* = \xiv_i + \frac{1}{2\gamma}\n_{\omegav_i} \ell(\thetav^*, \omegav_i^*).
\en
For large $\gamma$, this tells us that $\thetav^*$ is a stationary point of the original training loss given the adversarial examples $\omegav_i^*$, and $\omegav^*_i$ is a perturbation of the original samples $\xiv_i$. We furthermore want $\thetav^*$ to be a local minimum of the loss $\sum_{i=1}^N  \ell(\cdot, \omegav_i^*)$, and thus from the second-order necessary condition, we have:
\be
\sum_{i=1}^N \n_{\thetav \thetav} \ell(\thetav^*, \omegav_i^*)  \cge \zero.
\en
Note that it is very common in deep learning that the matrix $\sum_{i=1}^N \n_{\thetav \thetav} \ell(\thetav^*, \omegav_i^*)$ is singular \cite{sagun2016singularity}, and thus $(\thetav^*, \Omegav^*)$ is not a strict local Nash equilibrium (see \eqref{eq:strict_local_Nash}). However, we can show that $(\thetav^*, \Omegav^*)$ is a \slmm under mild assumptions. We note that \eqref{eq:assumption_on_gamma} can be guaranteed if $\gamma$ is greater than some Lipschitz smoothness constant of $\ell$, as shown in \citet{sinha2018certifiable}. 

\begin{proof}
We compute from \eqref{eq:def_f_distributional_robust} that:
\begin{align}
&\n_{\omegav_i^*, \omegav_i^*} f(\thetav^*, \Omegav^*) = \n_{\omegav \omegav} \ell(\thetav^*, \omegav_i^*) - 2\gamma \Iv, \tr
&\TD_{\thetav \thetav} f(\thetav^*, \Omegav^*) = \sum_{i=1}^N \n_{\thetav \thetav} \ell(\thetav^*, \omegav_i^*) - \Mv_i (\n_{\omegav \omegav} \ell(\thetav^*, \omegav_i^*) - 2\gamma \Iv)^{-1}  \Mv_i^\top.
\end{align}
If $\gamma > \frac{1}{2} \max_{i=1, \cdots, N}  \lambda_{\max}(\n_{\omegav \omegav} \ell(\thetav^*, \omegav_i^*))$, then for any $i = 1, \cdots, N$, 
\begin{align}\label{eq:partial_omegav_omegav}
\n_{\omegav_i, \omegav_i} f(\thetav^*, \Omegav^*) &= \n_{\omegav \omegav} \ell(\thetav^*, \omegav_i^*) - 2\gamma \Iv \tr
&\cl (\lambda_{\max}(\n_{\omegav \omegav} \ell(\thetav^*, \omegav_i^*)) - \max_{j=1, \cdots, N}  \lambda_{\max}(\n_{\omegav \omegav} \ell(\thetav^*, \omegav_j^*)))\Iv &\tr
&\cle \zero,
\end{align}
where in the second line, we used the fact that for a symmetric matrix $\Av$, we have $\Av \cle \lambda_{\max}(\Av) \Iv$. Hence we obtain that  $\n_{\omegav_i, \omegav_i} f(\thetav^*, \Omegav^*) \cl \zero$. We now compute $\TD_{\thetav \thetav} f(\thetav^*, \Omegav^*)$ as:
\begin{align}\label{eq:TD_thetav_thetav}
\TD_{\thetav \thetav} f(\thetav^*, \Omegav^*) &= \sum_{i=1}^N \n_{\thetav \thetav} \ell(\thetav^*, \omegav_i^*) - \Mv_i (\n_{\omegav \omegav} \ell(\thetav^*, \omegav_i^*) - 2\gamma \Iv)^{-1}  \Mv_i^\top \tr
&= \sum_{i=1}^N \n_{\thetav \thetav} \ell(\thetav^*, \omegav_i^*) - \sum_{i=1}^N \Mv_i (\n_{\omegav_i, \omegav_i} f(\thetav^*, \Omegav^*))^{-1}  \Mv_i^\top.
\end{align}
We assumed that $\thetav^*$ is a local minimum of the training loss and thus the first term is positive semi-definite. We note that the second term is negative semi-definite because for any model parameter $\thetav_0$ and any sample $\omegav_i^*$, we can write:
\begin{align}\label{eq:inner_product_with_theta0}
\thetav_0^{\top}\Mv_i (\n_{\omegav_i, \omegav_i} f(\thetav^*, \Omegav^*))^{-1}  \Mv_i^\top \thetav_0 =  (\Mv_i^\top \thetav_0)^{\top}(\n_{\omegav_i, \omegav_i} f(\thetav^*, \Omegav^*))^{-1}  \Mv_i^\top \thetav_0 \leq 0,
\end{align}
since $\n_{\omegav_i, \omegav_i} f(\thetav^*, \Omegav^*) \cl \zero $. Furthermore, if $\Mv_i^*$ has full row rank, \eqref{eq:inner_product_with_theta0} is always negative for all $\thetav_0 \neq \zero$, and hence the second term of \eqref{eq:TD_thetav_thetav} is negative definite, resulting in $\TD_{\thetav \thetav} f(\thetav^*, \Omegav^*) \cg \zero$. Assume otherwise. Since $(\n_{\omegav_i, \omegav_i} f(\thetav^*, \Omegav^*))^{-1}$ is also negative definite (this can be proved from the spectral decomposition), we must have:
\be\label{eq:product_mv_thetav}
\Mv_i^\top \thetav_0  = \zero.
\en
Since $\Mv_i$ is full row rank, the row vectors of $\Mv_i$ are linearly independent, and thus we must have $\thetav_0 = \zero$. This is a contradiction. So we have proved that $$\Mv_i (\n_{\omegav_i, \omegav_i} f(\thetav^*, \Omegav^*))^{-1}  \Mv_i^\top \cl \zero$$ and thus $\TD_{\thetav \thetav} f(\thetav^*, \Omegav^*) \cg \zero$. Therefore, \eqref{eq:partial_omegav_omegav} and \eqref{eq:TD_thetav_thetav} tell us that under our assumptions, $(\thetav^*, \Omegav^*)$ is a \slmm but not necessarily a strict local Nash equilibrium. 
\end{proof}

\section{Proofs of the non-asymptotic local convergence of Newton-type algorithms} \label{app:proof}


In this appendix, we discuss local convergence of Newton-type algorithms. The purpose of Appendix \ref{app:local_bound_lipschitz} is to derive the local Lipschitzness and boundedness of various first-order and second-order derivatives based on the assumption that $f$ is twice continuous differentiable and that the Hessian of $f$ is Lipschitz continuous (Assumption \ref{asp:lip_hessian}). Based on the derivations in Appendix \ref{app:local_bound_lipschitz}, in Appendix \ref{app:Newton_type_proof}, we derive the non-asymptotic local convergence of Newton-type algorithms, including GD-Newton and Complete Newton. 




\subsection{Local boundedness and Lipschitzness}\label{app:local_bound_lipschitz}

In order to quantify the absolute constants we mentioned in Theorems \ref{thm:GDN} and \ref{thm:CN}, we first quantify w.l.o.g.~that the neighborhoods in \eqref{eq:neighborhoods} to be:
\be
\Nc(\xv^*) = \Bc(\xv^*, \d_x) := \{\xv \in \R^n:  \|\xv - \xv^*\|\leq \d_x \}, \, \Nc(\yv^*) = \Bc(\yv^*, \d_y) := \{\yv \in \R^m:  \|\yv - \yv^*\|\leq \d_y \},
\en
where $\d_x > 0$ and $\d_y > 0$. Since $f$ is twice continuous differentiable, at its \slmm $(\xv^*, \yv^*)$, the second-order derivatives are bounded. There exist positive constants $B_{xx}, B_{xy}, B_{yy}$ such that for any $(\xv, \yv) \in \Bc(\xv^*, \d_x) \times \Bc(\yv^*, \d_y)$, 
\be\label{eq:bound_second_order}
\| \xxv f(\xv, \yv) \| \leq B_{xx}, \| \xyv f(\xv, \yv) \| \leq B_{xy}, \| \yyv f(\xv, \yv) \| \leq B_{yy}. 
\en
Since $\yxv f(\zv) = (\xyv f(\zv))^\top$ for $f\in \Cc^2$ (Schwarz's theorem), we have $\| \yxv f(\xv, \yv) \| \leq B_{xy}$ (the fact that the matrix $\Av$ and its transpose $\Av^\top$ have the same spectral norm can be derived from the SVD decomposition). 
For later convenience, we denote 
\be\label{eq:def_Bczstar}
\Bc(\zv^*) := \Bc(\xv^*, \d_x) \times \Bc(\yv^*, \d_y).
\en
Since $\yyv f(\xv^*, \yv^*) \cl \zero$ and $f\in \Cc^2$, we can assume w.l.o.g.~that for any $\zv \in \Bc(\zv^*)$, $\yyv f(\zv) \cle -\mu_y \Iv$. Therefore, $(\yyv f(\cdot))^{-1}$ is bounded on $\Bc(\zv^*)$, i.e., 
\be
\| (\yyv f(\zv))^{-1} \| \leq \mu_y^{-1}, \, \forall \zv \in \Bc(\zv^*).
\en
This is because of the following lemma:
\begin{lem}[\textbf{Local Lipschitzness and boundedness of the inverse}]\label{lem:bound_lip_inverse}
Suppose on a neighborhood $\Nc \subset \R^d$, there exists $\mu > 0$ s.t.~there exists a matrix-valued function $\Av: \Nc \to \R^{k\times k}$ that satisfies:
\be
\textrm{for any }\zv\in \Nc, \, \Av(\zv) \cle -\mu \Iv\textrm{ or }\textrm{for any }\zv\in \Nc, \Av(\zv) \cge \mu \Iv.
\en
then for any $\zv\in \Nc$, $\Av(\zv)$ is invertible and $\|\Av^{-1}(\zv)\| \leq \mu^{-1}$, with $\Av^{-1} : \zv \mapsto (\Av(\cdot))^{-1}$. Moreover, if $\Av$ is $L$-Lipschitz continuous, i.e., for any $\zv_1, \zv_2 \in \Nc$, we have 
\be
\| \Av(\zv_1) - \Av(\zv_2)\| \leq L\|\zv_1 - \zv_2\|, 
\en
then $\Av^{-1} := (\Av(\cdot))^{-1}$ is $\mu^{-2}L$-Lipschitz continuous, i.e., for any $\zv_1, \zv_2 \in \Nc$, we have 
\be
\| \Av^{-1}(\zv_1) - \Av^{-1}(\zv_2)\| \leq \mu^{-2} L\|\zv_1 - \zv_2\|. 
\en

\end{lem}

\begin{proof}
WLOG we only need to prove the case when $\Av(\zv) \cge \mu \Iv$ for any $\zv\in \Nc$, because we can take $\Bv = -\Av$ for the other case and apply the result on $\Bv$. The invertibility of $\Av(\zv)$ follows from the positive definiteness. From the definition of spectral norm we have that for any $\zv\in \Nc$:
\be\label{eq:inverse_1}
\|\Av^{-1}(\zv)\| = \sup_{\|\wv'\| = 1} \|\Av^{-1}(\zv) \wv'\| = \sup_{\|\Av(\zv) \wv\| = 1} \|\wv\| 
\en
On the other hand, $\Av(\zv) \cge \mu \Iv$ tells us that for any $\wv \in \R^d$ and $\|\Av(\zv) \wv \| = 1$, we can write:
\be\label{eq:inverse_2}
\mu \|\wv\|^2 \leq \wv^\top \Av(\zv) \wv \leq \|\wv\|\cdot \|\Av(\zv) \wv \| = \|\wv\|,
\en
where we used Cauchy--Schwarz inequality.
Combining \eqref{eq:inverse_1} and \eqref{eq:inverse_2} above we obtain that for any $\|\Av(\zv) \wv \| = 1$, we have $\|\wv\|\leq \mu^{-1}$ and thus for any $\zv\in \Nc$:
\be
\|\Av^{-1}(\zv)\| \leq \mu^{-1}.
\en

Therefore, for $\zv_1, \zv_2 \in \Nc$, we have from the Lipschitzness of $\Av$ that
\begin{align}
\|\Av^{-1}(\zv_1) - \Av^{-1}(\zv_2) \| &= \|\Av^{-1}(\zv_1) \Av(\zv_1) \Av^{-1}(\zv_2) - \Av^{-1}(\zv_1) \Av(\zv_2) \Av^{-1}(\zv_2) \|  \tr
&\leq \|\Av^{-1}(\zv_1) (\Av(\zv_1) - \Av(\zv_2)) \Av^{-1}(\zv_2) \| \tr
&\leq \|\Av^{-1}(\zv_1)\| \cdot \|\Av(\zv_1) - \Av(\zv_2)\|\cdot \|\Av^{-1}(\zv_2) \|\tr
&\leq \mu^{-2} L\|\zv_1 - \zv_2\|,
\end{align}
where in the third line we used that for two matrices $\Uv \in \R^d \to \R^{k\times k}$, $\Wv \in \R^d \to \R^{k\times k}$, 
\be\label{eq:product_norm}
\|\Uv \Vv\| = \sup_{\|\zv\| = 1} \|\Uv \Vv \zv\| \leq \sup_{\|\zv\| = 1} \|\Uv\|\cdot \|\Vv \zv\| =  \|\Uv\| \sup_{\|\zv\| = 1} \| \Vv \zv\| = \|\Uv\|\cdot \|\Vv\|.
\en
\end{proof}

Lemma \ref{lem:bound_lip_inverse} tells that $\yyv^{-1} f := (\yyv f(\cdot))^{-1}$ is $\mu_y^{-2} L_{yy}$-Lipschitz continuous under Assumption \ref{asp:lip_hessian}.  

Now let us derive the local Lipschitzness of the partial derivatives $\n_\xv f$ and $\n_\yv f$ from the local boundedness of the partial Hessians. For any $\zv_1, \zv_2 \in \Bc(\zv^*)$, we have:
\begin{align}
\| \n_\xv f(\zv_1) - \n_\xv f(\zv_2) \| &=  \| \n_\xv f(\xv_1, \yv_1) - \n_\xv f(\xv_2, \yv_2) \| \tr
&=  \| \n_\xv f(\xv_1, \yv_1) - \n_\xv f(\xv_1, \yv_2) + \n_\xv f(\xv_1, \yv_2) - \n_\xv f(\xv_2, \yv_2) \| \tr
&\leq  \| \n_\xv f(\xv_1, \yv_1) - \n_\xv f(\xv_1, \yv_2)\|  +  \| \n_\xv f(\xv_1, \yv_2) - \n_\xv f(\xv_2, \yv_2) \| \tr
& \leq \| \xyv f(\xv_1, \yv_{\xi})\|\cdot \|\yv_1 - \yv_2\| + \| \xxv f(\xv_{\gamma}, \yv_2)
\| \cdot \| \xv_1 - \xv_2\| \tr
&\leq B_{xy} \|\yv_1 - \yv_2\| + B_{xx} \| \xv_1 - \xv_2\| \tr
& \leq (B_{xy} + B_{xx})\|\zv_1 - \zv_2\|,
\end{align}
where in the fourth line we used the mean-value theorem and that $\yv_{\xi} \in [\yv_1, \yv_2]$ and $\xv_{\gamma} \in [\xv_1, \xv_2]$ ($[\av, \bv]$ denotes a line segment with end points $\av$ and $\bv$); in the second last line we used \eqref{eq:bound_second_order}; in the last line we used $\|\xv_1 - \xv_2\| \leq \|\zv_1 - \zv_2\|$ and $\|\yv_1 - \yv_2\| \leq \|\zv_1 - \zv_2\|$. Similarly, we can derive that:
\be
\| \n_\yv f(\zv_1) - \n_\yv f(\zv_2) \| \leq (B_{xy} + B_{yy}) \| \zv_1 - \zv_2\|.
\en

The local Lipschitzness of $\n_\xv f$ and $\n_\yv f$ also leads to their local boundedness. On the neighborhood $\Bc(\zv^*)$, we can derive:
\begin{align}
\| \n_\xv f(\zv) \| &= \| \n_\xv f(\zv) - \n_\xv f(\zv^*) \|  \leq L_x \|\zv - \zv^*\| \leq L_x (\| \xv - \xv^*\| + \|\yv - \yv^*\|)\leq L_x(\d_x + \d_y),
\end{align}
where we defined $L_x := B_{xy} + B_{xx}$ to be the Lipschitz constant of $\n_\xv f$ on the neighborhood. Similarly, we can derive that $\|\n_\yv f(\zv)\| \leq L_y (\d_x + \d_y)$ for any $\zv \in \Bc(\zv^*)$ with $L_y := B_{xy} + B_{yy}$. To summarize we have the following lemma. 

\begin{lem}[\textbf{Local Lipschitzness and boundedness}]\label{lem:local_lip_boundedness}
At a \slmm $\zv^*$ of a function $f\in \Cc^2$, there exist positive constants $B_{xx}$, $B_{xy}$, $B_{yy}$ and $\mu_y$ such that for any $\zv \in \Bc(\zv^*)$:
\be
\| \xxv f(\zv) \| \leq B_{xx}, \| \xyv f(\zv) \| \leq B_{xy}, \| \yyv f(\zv) \| \leq B_{yy}, \,\yyv f(\zv) \cle -\mu_y \Iv, \, 
\| (\yyv f(\zv))^{-1} \| \leq \mu_y^{-1},
\en
and $\n_\xv f$ and $\n_\yv f$ are locally Lipschitz, i.e.~for any $\zv_1, \zv_2 \in \Bc(\zv^*)$, we have
\be
&&\| \n_\xv f(\zv_1) - \n_\xv f(\zv_2) \| \leq L_x \|\zv_1 - \zv_2\| := (B_{xx} + B_{xy}) \|\zv_1 - \zv_2\|, \,\tr
&&\| \n_\yv f(\zv_1) - \n_\yv f(\zv_2) \| \leq L_y \|\zv_1 - \zv_2\| := (B_{xy} + B_{yy}) \| \zv_1 - \zv_2\|.
\en
Moreover, $\n_{\xv} f(\zv)$ and $\n_\yv f(\zv)$ are bounded, i.e., for any 
\be
\| \n_{\xv} f(\zv) \| \leq B_x := L_x (\d_x + \d_y), \, \|\n_\yv f(\zv)\| \leq B_y := L_y (\d_x + \d_y).
\en
Suppose Assumption \ref{asp:lip_hessian} holds on the neighborhood $\Bc(\zv^*)$, then $\yyv^{-1} f := (\yyv f(\cdot))^{-1}$ is $\mu_y^{-2} L_{yy}$-Lipschitz continuous, i.e.~for any $\zv_1, \zv_2 \in \Bc(\zv^*)$, we have
\be
\|(\yyv f(\zv_1))^{-1} - (\yyv f(\zv_2))^{-1} \|
\leq \mu_y^{-2} L_{yy}\|\zv_1 - \zv_2\|.
\en

\end{lem}

Let us now derive the local Lipschitzness of $\TD_\xv f$ and $\TD_{\xv\xv} f$. We need the composition rules of the Lipschitzness and boundedness of addition and product. Recall from Assumption~\ref{asp:lip_hessian} that for any $\zv_1, \zv_2 \in \Bc(\zv^*)$, we have:
\begin{align}\label{eq:lip_hessian}
&\|\xxv f(\zv_1) - \xxv f(\zv_2)\| \leq L_{xx} \|\zv_1 - \zv_2\|, \, \|\xyv f(\zv_1) - \xyv f(\zv_2)\| \leq L_{xy} \|\zv_1 - \zv_2\|,\, \tr
&\|\yyv f(\zv_1) - \yyv f(\zv_2)\| \leq L_{yy} \|\zv_1 - \zv_2\|.
\end{align}

\begin{lem}[\textbf{Local Lipschitzness and boundedness of addition}]
\label{lem:addition_of_Lip}
Suppose that on a neighborhood $\Nc \subset \R^d$, we have matrix-valued functions $\Av : \Nc \to \R^{k\times k}$, $\Bv : \Nc \to \R^{k\times k}$ and vector-valued functions $\vv: \Nc \to \R^k$, $\uv: \Nc \to \R^k$. Suppose that on the neighborhood $\Nc$, $\Av$ is $L_A$-Lipschitz continuous, $\Bv$ is $L_B$-Lipschitz continuous, $\vv$ is $L_v$-Lipschitz continuous and $\uv$ is $L_u$-Lipschitz continuous. Namely, for any $\zv_1, \zv_2 \in \Nc$, we have:
\begin{align}
&\|\Av(\zv_1) - \Av(\zv_2) \| \leq L_A \|\zv_1 - \zv_2\|,\, \|\Bv(\zv_1) - \Bv(\zv_2) \| \leq L_B \|\zv_1 - \zv_2\|, \tr
&\|\vv(\zv_1) - \vv(\zv_2)\| \leq L_v \|\zv_1 - \zv_2\|, \|\uv(\zv_1) - \uv(\zv_2)\| \leq L_u \|\zv_1 - \zv_2\|.
\end{align}
Then, the matrix-matrix addition function $\Av + \Bv: \zv \mapsto \Av(\zv) + \Bv(\zv)$ is $(L_A + L_B)$-Lipschitz continuous and the vector-vector addition function  $\uv + \vv: \zv \mapsto \uv(\zv) + \vv(\zv)$ is $(L_v + L_u)$-Lipschitz continuous, i.e.~for any $\zv_1, \zv_2 \in \Nc$, we have: 
\begin{align}
&\| (\Av + \Bv)(\zv_1) - (\Av + \Bv)(\zv_2) \| \leq (L_A + L_B) \cdot \| \zv_1 - \zv_2\|, \\
&\| (\uv+ \vv)(\zv_1) - (\uv + \vv)(\zv_2) \| \leq (L_u + L_v) \cdot \| \zv_1 - \zv_2\|.
\end{align}
Suppose $\Av$, $\Bv$, $\vv$, $\uv$ are $B_A$, $B_B, B_v, B_u$ bounded on the neighborhood $\Nc$, respectively, \ie for any $\zv \in \Nc$, we have:
\be
\|\Av(\zv)\|\leq B_A, \, \|\Bv(\zv)\| \leq B_B, \, \|\vv(\zv)\| \leq B_v, \, \|\uv(\zv)\| \leq B_u.
\en
Then, $\Av + \Bv$ is $(B_A + B_B)$-bounded and $\uv  + \vv$ is $(B_u + B_v)$-bounded on the neighborhood $\Nc$. 
\end{lem}

\begin{proof}
For any $\zv_1, \zv_2 \in \Nc$, we write:
\begin{align}
\| (\Av + \Bv)(\zv_1) - (\Av + \Bv)(\zv_2) \|
&= \| \Av(\zv_1) - \Av(\zv_2) + \Bv(\zv_1) - \Bv(\zv_2)\| \tr
&\leq \| \Av(\zv_1) - \Av(\zv_2)\| + \|\Bv(\zv_1) - \Bv(\zv_2)\|  \tr
&\leq L_A\|\zv_1 - \zv_2\| + L_B \|\zv_1 - \zv_2\| \tr
&= (L_A + L_B) \|\zv_1 - \zv_2\|.
\end{align}
Similarly, we can prove $\| (\uv+ \vv)(\zv_1) - (\uv + \vv)(\zv_2) \| \leq (L_u + L_v) \cdot \| \zv_1 - \zv_2\|$. The last sentence of Lemma \ref{lem:addition_of_Lip} follows from the triangle inequalities of norms.  
\end{proof}

\begin{lem}[\textbf{Local Lipschitzness and boundedness of product}]\label{lem:lip_of_product}
Suppose that on a neighborhood $\Nc\subset \R^d$, we have matrix-valued functions $\Av : \Nc \to \R^{k\times k}$, $\Bv : \Nc \to \R^{k\times k}$ and a vector-valued function $\vv: \Nc \to \R^k$. Suppose that on the neighborhood $\Nc$, $\Av$ is $L_A$-Lipschitz continuous and $B_A$ bounded, $\Bv$ is $L_B$-Lipschitz continuous and $B_B$ bounded, $\vv$ is $L_v$-Lipschitz continuous and $B_v$ bounded. Namely, for any $\zv_1, \zv_2 \in \Nc$, we have:
\begin{align}
\|\Av(\zv_1) - \Av(\zv_2) \| \leq L_A \|\zv_1 - \zv_2\|,\, \|\Bv(\zv_1) - \Bv(\zv_2) \| \leq L_B \|\zv_1 - \zv_2\|, \, \|\vv(\zv_1) - \vv(\zv_2)\| \leq L_v \|\zv_1 - \zv_2\|, 
\end{align}
and for any $\zv\in \Nc$, 
\begin{align}
\|\Av(\zv)\| \leq B_A, \|\Bv(\zv)\| \leq B_B, \|\vv(\zv)\| \leq B_v.
\end{align}
Then, the matrix-matrix product function $\Av \Bv: \zv \mapsto \Av(\zv) \Bv(\zv)$ and the matrix-vector product function  $\Av \vv: \zv \mapsto \Av(\zv) \vv(\zv)$ on the neighborhood $\Nc$ are also Lipschitz, i.e.~for any $\zv_1, \zv_2 \in \Nc$, we have: 
\begin{align}
&\| \Av(\zv_1) \Bv(\zv_1) - \Av(\zv_2) \Bv(\zv_2) \| \leq (B_A L_B + B_B L_A) \cdot \| \zv_1 - \zv_2\|, \\
&\| \Av(\zv_1) \vv(\zv_1) - \Av(\zv_2) \vv(\zv_2) \| \leq (B_A L_v + B_v L_A) \cdot \| \zv_1 - \zv_2\|.
\end{align}
Moreover, $\Av \Bv$ is $B_A B_B$-bounded and $\Av \vv$ is $B_A B_v$-bounded on the neighborhood $\Nc$. 
\end{lem}

\begin{proof}
For any $\zv_1, \zv_2 \in \Nc$, we have: 
\begin{align}
\| \Av(\zv_1) \Bv(\zv_1) - \Av(\zv_2) \Bv(\zv_2)\| &= \| \Av(\zv_1) \Bv(\zv_1) - \Av(\zv_1) \Bv(\zv_2) +  \Av(\zv_1) \Bv(\zv_2)  - \Av(\zv_2) \Bv(\zv_2)\|  \tr
&\leq \| \Av(\zv_1) \Bv(\zv_1) - \Av(\zv_1) \Bv(\zv_2) \| +  \| \Av(\zv_1) \Bv(\zv_2)  - \Av(\zv_2) \Bv(\zv_2)\| \tr
&=  \| \Av(\zv_1) (\Bv(\zv_1) - \Bv(\zv_2)) \| +  \| (\Av(\zv_1)  - \Av(\zv_2)) \Bv(\zv_2)\|  \tr
&\leq  \| \Av(\zv_1) \| \cdot \|\Bv(\zv_1) - \Bv(\zv_2)\| +  \|\Av(\zv_1)  - \Av(\zv_2)\|\cdot \| \Bv(\zv_2)\|\tr
&\leq (B_A L_B + B_B L_A) \|\zv_1 - \zv_2\|,
\end{align}
where in the fourth line we used \eqref{eq:product_norm}.
Similarly, we can derive that for $\zv_1, \zv_2 \in \Nc$, we have:
\begin{align}
    \| \Av(\zv_1) \vv(\zv_1) - \Av(\zv_2) \vv(\zv_2)\| \leq (B_A L_v + B_v L_A)\|\zv_1 - \zv_2\|.
\end{align}
The final claim follows from \eqref{eq:product_norm} and that for any $\zv\in \Nc$, $\|\Av(\zv) \vv(\zv)\| \leq \|\Av(\zv)\| \cdot \|\vv(\zv)\|$. 
\end{proof}


We can now derive the local Lipschitzness of $\TD_\xv f$ and $\TD_{\xv\xv} f$ under Assumption~\ref{asp:lip_hessian}. On the neighborhood $\Bc(\zv^*)$, since $\yyv^{-1} f$ is $\mu_y^{-2} L_{yy}$-Lipschitz continuous from Lemma \ref{lem:bound_lip_inverse} and $\mu_y^{-1}$-bounded from Lemma \ref{lem:local_lip_boundedness}, and $\n_{\yv} f$ is $L_y$-Lipschitz continuous and $B_y$-bounded, from Lemma \ref{lem:lip_of_product},
\be\label{eq:yyv_n_yv}
\yyv^{-1} f \cdot \n_{\yv} f\mbox{ is }(\mu_y^{-1} L_y  + B_y \mu_y^{-2} L_{yy})\mbox{-Lipschitz continuous, and } \mu_y^{-1}B_y\mbox{ bounded.}
\en
Since $\xyv f$ is $B_{xy}$-bounded and $L_{xy}$-Lipschitz continuous from Assumption~\ref{asp:lip_hessian}, $\xyv f \cdot \yyv^{-1} f \cdot \n_{\yv} f$ is
\be
L_{xy}\mu_y^{-1}B_y + B_{xy}(\mu_y^{-1} L_y  + B_y \mu_y^{-2} L_{yy})
\en
Lipschitz continuous and 
\be
B_{xy}\mu_y^{-1}B_y
\en
bounded on $\Bc(\zv^*)$. Finally, from Lemma \ref{lem:addition_of_Lip}, $\TD_{\xv} f = \n_\xv f - \xyv f \cdot \yyv^{-1} f \cdot \n_{\yv} f$ is:
\be
L_x + L_{xy}\mu_y^{-1}B_y + B_{xy}(\mu_y^{-1} L_y  + B_y \mu_y^{-2} L_{yy}) 
\en
Lipschitz continuous and
$
B_x + B_{xy}\mu_y^{-1}B_y 
$
bounded.
In a similar way, $\TD_{\xv \xv} f = \n_{\xv\xv} f - \xyv f \cdot \yyv^{-1} f \cdot \n_{\yv \xv} f$ is 
\be
L_{xx} + 2 L_{xy} B_{xy} \mu_y^{-1} + B_{xy}^2 \mu_y^{-2} L_{yy}
\en
Lipschitz continuous and
$
B_{xx} + B_{xy}^2 \mu_y^{-1} 
$
bounded. We summarize our result as follows:
\begin{lem}[\textbf{Local Lipschitzness and boundedness of $\TD_{\xv}f$ and $\TD_{\xv\xv}f$}]\label{lem:tdx_and_tdxx}
Suppose on the neighborhood $\Bc(\zv^*)$ of a \slmm $\zv^*$ of a function $f\in \Cc^2$, Assumption \ref{asp:lip_hessian} holds. $\TD_{\xv} f = \n_\xv f - \xyv f \cdot \yyv^{-1} f \cdot \n_{\yv} f$ is:
\be
L_x^\TD := L_x + L_{xy}\mu_y^{-1}B_y + B_{xy}(\mu_y^{-1} L_y  + B_y \mu_y^{-2} L_{yy}) 
\en
Lipschitz continuous and
$$
B_x^\TD := B_x + B_{xy}\mu_y^{-1}B_y 
$$
bounded, i.e., for any $\zv, \zv_1, \zv_2 \in \Bc(\zv^*)$, we have that:
\be
\| \TD_\xv f(\zv_1) - \TD_\xv f(\zv_2)\| \leq L_x^\TD \|\zv_1 - \zv_2\|, \, \| \TD_\xv f(\zv)\| \leq B_x^\TD.
\en
In a similar way, $\TD_{\xv \xv} f = \n_{\xv\xv} f - \xyv f \cdot \yyv^{-1} f \cdot \n_{\yv \xv} f$ is 
\be
L_{xx}^\TD := L_{xx} + 2 L_{xy} B_{xy} \mu_y^{-1} + B_{xy}^2 \mu_y^{-2} L_{yy}
\en
Lipschitz continuous and
$$
B_{xx}^\TD  := B_{xx} + B_{xy}^2 \mu_y^{-1} 
$$
bounded, where the constants are the same as in Lemma \ref{lem:local_lip_boundedness}. i.e., for any $\zv, \zv_1, \zv_2 \in \Bc(\zv^*)$, we have:
\be
\| \TD_{\xv\xv} f(\zv_1) - \TD_{\xv\xv} f(\zv_2)\| \leq L_{xx}^\TD \|\zv_1 - \zv_2\|, \, \| \TD_{\xv\xv} f(\zv)\| \leq B_{xx}^\TD.
\en
\end{lem}

Finally, we derive the local Lipschitzness of the derivatives of the local maximum function $\psi(\xv) = f(\xv, r(\xv))$ where $\xv\in \Nc(\xv^*)$, i.e., 
\be
\psi'(\xv) = \TD_{\xv} f(\xv, r(\xv)), \, \psi''(\xv) = \TD_{\xv\xv} f(\xv, r(\xv)).
\en
This is because from the chain rule and \eqref{eq:deriv_r}, we have: 
\begin{align}\label{eq:total_derivative}
\psi'(\xv) &= \n_\xv f(\xv, r(\xv)) + r'(\xv)^\top \n_\yv f(\xv, r(\xv))  \tr
&= \n_\xv f(\xv, r(\xv)) - (\xyv \cdot \yyv^{-1}) f(\xv, r(\xv)) \cdot \n_\yv f(\xv, r(\xv)) \tr
&= (\n_\xv - \xyv \cdot \yyv^{-1} \cdot \n_\yv) f(\xv, r(\xv)) \tr
&= \TD_\xv f(\xv, r(\xv)).
\end{align}
Taking the total derivative of $\xv$ again and using $\n_\yv f(\xv, r(\xv)) = \zero$, we have:
\begin{align}\label{eq:total_second_order_derivative}
\psi''(\xv) &= \dv{}{\xv} \TD_\xv f(\xv, r(\xv)) \tr
&= \dv{}{\xv} \partial_\xv f(\xv, r(\xv))  \tr
& = \partial_{\xv\xv} f(\xv, r(\xv)) + r'(\xv)^\top  \partial_{\yv \xv} f(\xv, r(\xv)) \tr 
&=\partial_{\xv\xv} f(\xv, r(\xv)) - (\xyv \cdot \yyv^{-1}) f(\xv, r(\xv))\cdot \yxv f(\xv, r(\xv)) \tr
&= (\partial_{\xv\xv} - \xyv \cdot \yyv^{-1} \cdot \yxv) f(\xv, r(\xv)) \tr
&= \TD_{\xv\xv} f(\xv, r(\xv)).
\end{align}

From Lemma \ref{lem:tdx_and_tdxx}, for any $\xv_1, \xv_2 \in \Nc(\xv^*) = \Bc(\xv^*, \d_x)$, we have that
\begin{align}
\|\psi'(\xv_1) - \psi'(\xv_2)\| &= \|\TD_{\xv} f(\xv_1, r(\xv_1)) - \TD_{\xv} f(\xv_2, r(\xv_2))\| \tr
&\leq L_x^\TD \|(\xv_1, r(\xv_1)) - (\xv_2, r(\xv_2))\| \tr
&\leq L_x^\TD (\|\xv_1 - \xv_2\| + \|r(\xv_1) - r(\xv_2) \|) \tr
&= L_x^\TD (\|\xv_1 - \xv_2\| + \||r'(\xv_\g)(\xv_1 - \xv_2)\|) \tr
&\leq L_x^\TD (1 + \|r'(\xv_\g)\|)\|\xv_1 - \xv_2\| \tr
&=  L_x^\TD (1 + \|-(\yyv^{-1} \cdot \yxv) f(\xv_\g, r(\xv_\g))\|)\|\xv_1 - \xv_2\| \tr
&\leq L_x^\TD (1 + \mu_y^{-1}B_{xy})\|\xv_1 - \xv_2\|,
\end{align}
where in the fourth line we used the mean-value theorem and that $\xv_\g$ is on the line segment with end points $\xv_1$ and $\xv_2$. In the sixth line we used \eqref{eq:deriv_r} and in the last line we used the local boundedness of $\yyv^{-1} f$ and $\yxv f$ in Lemma \ref{lem:local_lip_boundedness} and \eqref{eq:product_norm}. Similarly, we can derive that:
\begin{align}
\|\psi''(\xv_1) - \psi''(\xv_2)\| &= \|\TD_{\xv\xv} f(\xv_1, r(\xv_1)) - \TD_{\xv\xv} f(\xv_2, r(\xv_2))\| \tr
&\leq L_{xx}^\TD \|(\xv_1, r(\xv_1)) - (\xv_2, r(\xv_2))\| \tr
&\leq  L_{xx}^\TD (1 + \mu_y^{-1}B_{xy})\|\xv_1 - \xv_2\|.
\end{align}
We summarize these conclusions:
\begin{lem}[\textbf{Local Lipschitzness of $\psi'(\xv)$ and $\psi''(\xv)$}]\label{lem:Lipschitz_of_psi_derivs}
Under the same assumption as in Lemma \ref{lem:tdx_and_tdxx} we define $$\psi(\xv) := f(\xv, r(\xv))\mbox{ where }\xv \in \Nc(\xv^*).$$ We have $\psi'(\xv) = \TD_{\xv} f(\xv, r(\xv))$ and $\psi''(\xv) = \TD_{\xv\xv} f(\xv, r(\xv))$. Moreover, $\psi'(\xv)$ and $\psi''(\xv)$ are Lipschitz continuous on $\Nc(\xv^*)$, namely, for any $\xv\in \Nc(\xv^*)$, we have that:
\begin{align}
&\|\psi'(\xv_1) - \psi'(\xv_2)\| = \|\TD_{\xv} f(\xv_1, r(\xv_1)) - \TD_{\xv} f(\xv_2, r(\xv_2))\| \leq L_{x}^\psi \|\xv_1 - \xv_2\|, \\
&\|\psi''(\xv_1) - \psi''(\xv_2)\| = \|\TD_{\xv\xv} f(\xv_1, r(\xv_1)) - \TD_{\xv\xv} f(\xv_2, r(\xv_2))\| \leq L_{xx}^\psi \|\xv_1 - \xv_2\|,
\end{align}
where we define 
\be\label{eq:definition_psi_derivs} 
L_{x}^\psi := L_{x}^\TD (1 + \mu_y^{-1}B_{xy})\mbox{ and }
L_{xx}^\psi := L_{xx}^\TD (1 + \mu_y^{-1}B_{xy}),
\en
and the constants $L_{x}^\TD, L_{xx}^\TD, \mu_y, B_{xy}$ are defined in Lemmas \ref{lem:local_lip_boundedness} and \ref{lem:tdx_and_tdxx}.
\end{lem}

Since $\TD_{\xv\xv} f \cg \zero$ for any $\zv\in \Bc(\zv^*)$ and we have proved in Lemma \ref{lem:tdx_and_tdxx} that $\TD_{\xv\xv} f$ is (Lipschitz) continuous, there exist a positive constant $\mu_x > 0$ s.t.~
\be
\TD_{\xv\xv} f(\zv) \cge \mu_x \Iv, \, \textrm{ for any }\zv \in \Bc(\zv^*). 
\en
From Lemma \ref{lem:bound_lip_inverse} we obtain that:
\begin{lem}\label{lem:inverse_of_tdxx}
At a \slmm $\zv^*$ of a function $f\in \Cc^2$, suppose that Assumption \ref{asp:lip_hessian} holds on the neighborhood $\Bc(\zv^*)$. There exists $\mu_x > 0$ s.t.~$\TD_{\xv\xv}^{-1} f := (\TD_{\xv\xv} f(\cdot))^{-1}$ is locally bounded and Lipschitz continuous, i.e., for any $\zv \in \Bc(\zv^*)$, we have:
\be
\|\TD_{\xv\xv}^{-1} f(\zv) \| \leq \mu_x^{-1}, 
\en
and for any $\zv_1, \zv_2 \in \Bc(\zv^*)$, we have:
\be
\|\TD_{\xv\xv}^{-1} f(\zv_1) - \TD_{\xv\xv}^{-1} f(\zv_2)  \| \leq \mu_x^{-2} L_{xx}^\TD \| \zv_1 - \zv_2\|,
\en
where $L_{xx}^\TD$ is defined in Lemma \ref{lem:tdx_and_tdxx}. 
\end{lem}

With all the preparations we have made, we are now ready to prove the non-asymptotic local convergence of our Newton-type methods, GD-Newton and Complete Newton.  

\subsection{Proof of the local non-asymptotic convergence of Newton-type methods}\label{app:Newton_type_proof}

We provide a more detailed version of Theorem \ref{thm:GDN}: 

\begin{restatable}[\textbf{GD-Newton}]{thm}{}
\label{thm:GDN_app}
Given a {\slmm} $(\xv^*, \yv^*)$ and $\d_x > 0$, $\d_y > 0$, suppose on the neighborhoods $\Nc(\xv^*) = \Bc(\xv^*, \d_x) $ and $\Nc(\yv^*) = \Bc(\yv^*, \d_y)$, Assumption \ref{asp:lip_hessian} holds and the local best-response function $r: \Nc(\xv^*) \to \Nc(\yv^*)$ exists. Suppose $\mu_x \Iv \cle \TD_{\xv\xv} f(\xv, \yv) \cle M_x \Iv$ for any $(\xv, \yv) \in \Nc(\xv^*)\times \Nc(\yv^*)$. Define:
\be\label{ineq:init_gdn_app}
\Nc_{\rm GDN} := \left\{\zv \in \R^{n+m} :\|\xv - \xv^*\| \leq \d, \|\yv - \yv^*\| \leq 2V \d \right\}
\en
where 
\be
\d = \min\left\{\d_x, \frac{\d_y}{2V}, \frac{\rho_\La}{4V^2 U}, \frac{\e}{\a_\La M (1 + 4 V^2/\rho_\La^2)}\right\}
\en
and $\Bc(\zv^*)$ is defined in \eqref{eq:def_Bczstar}, $\rho_\La = |1 - \a_\La \mu_x|\vee |1 - \a_\La M_x|, 0< \e \leq 1 - \rho_\La$, and $U$, $V$, $M$ satisfy: 
\begin{align}\label{eq:UVM}
& U := L_{yy}(2 \mu_y)^{-1}, V := ( B_y L_{yy} + \mu_y L_y) \mu_y^{-2},\, M := (B_{xy} + L_x^\TD) L_{yy} (2\mu_y^3)^{-1} L_y^2,
\end{align}
where $\mu_y$, $B_y$, $L_x$, $L_y$, $L^\TD_{x}$ are defined in Lemmas \ref{lem:local_lip_boundedness} and \ref{lem:tdx_and_tdxx}.
Given an initialization $(\xv_1, \yv_1) \in \Nc_{\rm GDN}$, $\|\yv_1 - \yv^*\| \leq 2V \|\xv_1 - \xv^*\|$ and suppose that $(\xv_2, \yv_2) \in  \Nc_{\rm GDN}$ and $\|\yv_2 - \yv^*\| \leq 2V \|\xv_2 - \xv^*\|$, the convergence of the GD-Newton to $(\xv^*, \yv^*)$ is linear, \ie, for any $t \geq 2$, we have:
\be\label{ineq:linear_gdn_app}
&&\|\xv_{t + 1} - \xv^*\|\leq (\rho_\La + \e)^{t-1} \| \xv_2 - \xv^*\|, \, \|\yv_{t + 1} - \yv^*\| \leq 2V (\rho_\La + \e)^{t-1} \| \xv_2 - \xv^*\|.
\en
\end{restatable}

Before we move on to the proof, we observe the dependence of the neighborhood on condition numbers. In fact, for the GD-Newton method, there are two condition numbers. We denote
\be\label{eq:kappa_ys}
\kappa_{1, y} := \frac{L_y}{\mu_y}, \, \kappa_{2, y} := \frac{L_{yy}}{\mu_y}.
\en
$\kappa_{1, y}$ is the usual condition number we when study first order algorithms. For Newton-type algorithms, $\kappa_{2, y}$ arises (Prop.~1.4.1, \citet{bertsekas1997nonlinear}). From \eqref{eq:UVM}, and Lemma \ref{lem:local_lip_boundedness}, the absolute constants can be written as:
\be\label{eq:interpret_UVM_kappa}
U = \kappa_{2, y}/2,\, V = \kappa_{1, y}\kappa_{2, y} (\d_x + \d_y) + \kappa_{1, y}, \, M = (B_{xy} + L_x^\TD) \kappa_{2, y}\kappa_{1, y}^2/2.
\en
Namely, the neighborhood size depends on the condition numbers. This is not uncommon in conventional minimization, for both first- and second-order algorithms (e.g.~\citet{nesterov2003introductory}, Theorems 1.2.4 and 1.2.5). 

We also note that the constant $M_x$ in $\mu_x \Iv \cle \TD_{\xv\xv} f(\zv) \cle M_x \Iv$ can be taken to be $B_{xx}^\TD$ as in Lemma \ref{lem:tdx_and_tdxx}, because for any $\zv \in \Bc(\zv^*)$ and $\xv\in \R^n$ s.t.~$\|\xv\| = 1$, we have $\|\TD_{\xv\xv} f(\zv) \xv\| \leq B_{xx}^\TD$, and thus from Cauchy--Schwarz inequality we have $\xv^\top \TD_{\xv\xv} f(\zv) \xv 
\leq \|\xv\|\cdot \|\TD_{\xv\xv} f(\zv) \xv\| \leq B_{xx}^\TD$. Therefore, for any $\zv\in \Bc(\zv^*)$, we obtain $D_{\xv \xv} f(\zv) \cle B_{xx}^\TD \Iv$.  

\begin{proof}

Now let us study the exact convergence rate. We can prove that $\Nc_{\rm GDN} \subset \Bc(\xv^*, \d_x) \times \Bc(\xv^*, \d_y)$ because for any $\zv = (\xv, \yv) \in \Nc_{\rm GDN}$, we have $\|\xv - \xv^*\| \leq \d \leq \d_x$ and $\|\yv - \yv^*\| \leq 2V \d \leq 2V\cdot \tfrac{\d_y}{2V} = \d_y$. Hence, all our results in Appendix \ref{app:local_bound_lipschitz} are valid on $\Nc_{\rm GDN}$. 

Suppose $(\xv_k, \yv_k) \in \Nc_{\rm GDN}$ for $k\leq t$ and $t\geq 2$. We first prove that:
\be\label{eq:first_y_gdn}
\|\yv_{t+1} - \yv^*\|\leq U \|\yv_t - \yv^*\|^2 + V \|\xv_{t+1} - \xv^*\|,
\en
and then
\be\label{eq:second_x_gdn}
\|\xv_{t+1} - \xv^*\| \leq \rho_\La \|\xv_t - \xv^*\| + \a_\La  M ( \| \xv_t - \xv^*\|^2 + \|\yv_{t-1} - \yv^*\|^2),
\en
where
\be
\rho_\La = |1 - \a_\La \mu_x|\vee |1 - \a_\La M_x|, \, M = (B_{xy} + L_x^\TD) L_{yy} (2\mu_y^3)^{-1} L_y^2.
\en
With these two inequalities, we will prove in Part III that for $t\geq 2$:
\be
&&\|\xv_{t + 1} - \xv^*\|\leq (\rho_\La + \e)^{t-1} \| \xv_2 - \xv^*\|, \, \|\yv_{t + 1} - \yv^*\| \leq 2V (\rho_\La + \e)^{t-1} \| \xv_2 - \xv^*\|.
\en

\paragraph{Part I} To prove \eqref{eq:first_y_gdn}, note that
\begin{align}
\|\yv_{t+1} - \yv^*\| &= \|\yv_t - \yv^* - (\yyv^{-1}\cdot\n_\yv) f(\xv^*, \yv_t) + (\yyv^{-1}\cdot\n_\yv) f(\xv^*, \yv_t) - (\yyv^{-1}\cdot\n_\yv) f(\xv_{t+1}, \yv_t)\|
\nonumber\\
&\leq \|\yyv^{-1}f(\xv^*, \yv_t)(\yyv f(\xv^*, \yv_t)(\yv_t - \yv^*) - \n_\yv f(\xv^*, \yv_t))\| +
\nonumber\\
&\qquad + \|(\yyv^{-1}\cdot\n_\yv) f(\xv^*, \yv_t) - (\yyv^{-1}\cdot\n_\yv) f(\xv_{t+1}, \yv_t)\|.
\end{align}
From the local Lipschitzness of $(\yyv^{-1}\cdot \n_\yv) f$, \eqref{eq:yyv_n_yv}, we know that the second term is at most $$ (\mu_y^{-1} L_y  + B_y \mu_y^{-2} L_{yy}) \|\xv_{t+1} - \xv^*\| = V \|\xv_{t+1} - \xv^*\|.$$
Since we assumed that $(\xv_t, \yv_t) \in \Nc_{\rm GDN} \subset \Bc(\zv^*)$, we can derive that $(\xv^*, \yv_t) \in \Bc(\zv^*)$ and $\yv_t \in \Nc(\yv^*)$. The first term can be upper bounded as:
\begin{align}\label{eq:first_term_gdn}
&\;\|\yyv^{-1}f(\xv^*, \yv_t)\|\cdot \|(\yyv f(\xv^*, \yv_t)(\yv_t - \yv^*) - \n_\yv f(\xv^*, \yv_t) + \n_\yv f(\xv^*, \yv^*)\|  \tr
&\leq \mu_y^{-1} \cdot \|(\yyv f(\xv^*, \yv_t)(\yv_t - \yv^*) - \n_\yv f(\xv^*, \yv_t) + \n_\yv f(\xv^*, \yv^*)\| \tr
&\leq \mu_y^{-1} \int_0^1 \|\yyv f(\xv^*, \yv_t) -  \n_{\yv\yv} f(\xv^*, \yv^* + s (\yv_t - \yv^*))\|\cdot \|\yv_t - \yv^*\| ds \tr
&\leq \mu_y^{-1} \int_0^1 L_{yy} (1 -s) \|\yv_t - \yv^*\|^2 ds \tr
& = L_{yy} (2 \mu_y)^{-1}\|\yv_t - \yv^*\|^2 \tr
&= U \|\yv_t - \yv^*\|^2,
\end{align}
where in the second line we used Lemma \ref{lem:bound_lip_inverse}; in the third line we used the following identity:
\be\label{eq:int_ny_gdn}
\n_\yv f(\xv, \yv_1) - \n_\yv f(\xv, \yv_2)  =  \int_{0}^1 \n_{\yv\yv} f(\xv, \yv_2 + s (\yv_1 - \yv_2)) (\yv_1 - \yv_2) ds,
\en
and in the fourth line we used Assumption \ref{asp:lip_hessian}. Therefore we have proved \eqref{eq:first_y_gdn}.


\paragraph{Part II} To prove \eqref{eq:second_x_gdn}, we observe that:
\begin{align}\label{eq:update_x_gdn}
\|\xv_{t+1} - \xv^*\| &= \|\xv_t - \xv^* - \a_\La \n_\xv f(\xv_t, \yv_t)\| \tr
&= \|\xv_t - \xv^* - \a_\La \TD_\xv f(\xv_t, r(\xv_t)) + \a_\La (\TD_\xv f(\xv_t, r(\xv_t)) - \TD_\xv f(\xv_t, \yv_t)) + \a_\La (\TD_\xv f(\xv_t, \yv_t) - \n_\xv f(\xv_t, \yv_t))\| \tr
&= \|\xv_t - \xv^* - \a_\La \TD_\xv f(\xv_t, r(\xv_t))\| + \a_\La \| \TD_\xv f(\xv_t, r(\xv_t)) - \TD_\xv f(\xv_t, \yv_t) \| + \a_\La \| \TD_\xv f(\xv_t, \yv_t) - \n_\xv f(\xv_t, \yv_t) \|.
\end{align}
Note that $r(\xv_t) \in \Nc(\yv^*)$ because of \eqref{eq:neighborhoods}. So $(\xv_t, r(\xv_t)) \in \Bc(\zv^*)$ and our analysis is valid.
Now let us bound the three terms separately. The first term can be computed as
\begin{align}\label{eq:spec_first}
\|\xv_t - \xv^* - \a_\La \TD_\xv f(\xv_t, r(\xv_t))\| &= \|\xv_t - \xv^* - \a_\La (\psi'(\xv_t) - \psi'(\xv^*))\| \tr
&= 
\| \xv_t - \xv^* - \a_\La \int_0^1 \psi''(\xv^* + s(\xv_t - \xv^*))(\xv_t - \xv^*)ds \| \tr
&= 
\| \int_0^1 (\Iv - \a_{\La}\psi''(\xv^* + s(\xv_t - \xv^*)) )(\xv_t - \xv^*)ds \| \tr
&\leq \int_0^1\| \Iv - \a_{\La}\psi''(\xv^* + s(\xv_t - \xv^*))\|\cdot \| \xv_t - \xv^*\| ds \tr
&\leq \rho_\La \| \xv_t - \xv^*\|,
\end{align}
where in the first line we used from Lemma \ref{lem:Lipschitz_of_psi_derivs}, $\psi'(\xv) = \TD_\xv f(\xv, r(\xv))$ and $\psi'(\xv^*) = \TD_\xv f(\xv^*, r(\xv^*)) = \zero$. In the last line, we used from Lemma \ref{lem:Lipschitz_of_psi_derivs} that $\psi''(\xv) = \TD_{\xv\xv} f(\xv, r(\xv))$ and our assumption $\mu_x \Iv \cle \TD_{\xv\xv} f(\zv) \cle M_x \Iv$ for any $\zv\in \Bc(\zv^*)$. More specifically, since \be
\xv_s = \xv^* + s(\xv_t - \xv^*) \in \Nc(\xv^*), 
\en
we have
\be
\Iv - \a_\La \psi''(\xv_s) = \Iv - \a_\La \TD_{\xv\xv}(\xv_s, r(\xv_s)),
\en
and 
\be
(1- \a \La M_x)\Iv \cle \Iv - \a_\La \TD_{\xv\xv}(\xv_s, r(\xv_s)) \cle (1- \a \La \mu_x)\Iv.
\en
Therefore, 
\be
\|\Iv - \a_\La \TD_{\xv\xv}(\xv_s, r(\xv_s))\| \leq |1 - \a_{\La} \mu_x| \vee 
|1 - \a_{\La} M_x| = \rho_\La.
\en

From Lemma \ref{lem:tdx_and_tdxx} the second term can be bounded as:
\be\label{eq:eqa18}
\a_\La \| \TD_\xv f(\xv_t, r(\xv_t)) - \TD_\xv f(\xv_t, \yv_t) \| \leq \a_\La L^\TD_{x} \| r(\xv_t) - \yv_t \|,
\en
From the definition of $\TD_\xv f := \n_\xv f - \xyv f \cdot (\yyv f)^{-1} \cdot \n_\yv f$, the third term can be bounded as:
\begin{align}\label{eq:eqa20}
\a_\La \| \TD_\xv f(\zv_t) - \n_\xv f(\zv_t) \| &= \a_\La \| (\n_{\xv \yv} \cdot \n_{\yv \yv}^{-1} \cdot \n_\yv) f(\zv_t)\| \tr
&\leq \a_\La\| \n_{\xv \yv} f(\zv_t)\|\cdot \|\n_{\yv \yv}^{-1} f(\zv_t)\| \cdot \|\n_\yv f(\zv_t)\| \tr
&\leq \a_\La B_{xy} \mu_y^{-1} \| \n_\yv f(\zv_t)\|,
\end{align}
where we used Lemma \ref{lem:local_lip_boundedness} and the assumption $\zv_t \in \Bc(\zv^*)$ from induction. To upper bound $\| \n_\yv f(\zv_t)\|$, note that:
\be\label{eq:rxtyt_gdn}
\| \n_\yv f(\zv_t) \| = \|\n_\yv f(\xv_t, \yv_t)\| = \|\n_\yv f(\xv_t, \yv_{t-1} -\Delta \yv)\|,
\en
with $\Delta \yv = (\yyv^{-1}\cdot \n_\yv) f(\xv_t, \yv_{t-1})$. Therefore, 
\begin{align}\label{eq:bound_yt_xt_gdn}
\|\n_\yv(\zv_t)\| = \|\n_\yv f(\xv_t, \yv_{t-1} -\Delta \yv)\| &= \|\n_\yv f(\xv_t, \yv_{t-1}-\Delta \yv) - \n_\yv f(\xv_t, \yv_{t-1}) - \yyv f(\xv_t, \yv_{t-1})(-\Delta \yv)\|\tr
&= \|\int_0^1 \left(\yyv f(\xv_t, \yv_{t-1} - s \Delta \yv) - \yyv f(\xv_t, \yv_{t-1})\right)(-\Delta \yv) ds \| \tr
&\leq \int_0^1 \| \left(\yyv f(\xv_t, \yv_{t-1} - s \Delta \yv) - \yyv f(\xv_t, \yv_{t-1})\right)(-\Delta \yv) \| ds \tr
&\leq  \int_0^1 \|\yyv f(\xv_t, \yv_{t-1} - s \Delta \yv) - \yyv f(\xv_t, \yv_{t-1})\|\cdot \|(-\Delta \yv) \| ds \tr
&\leq \int_0^1 L_{yy} s \|\Delta \yv\|^2 ds \tr
&= \tfrac{1}{2}L_{yy}\|\Delta \yv\|^2 \tr
&= \tfrac{1}{2} L_{yy}\|(\yyv^{-1} \cdot \n_\yv) f(\xv_t, \yv_{t-1})\|^2 \tr
&\leq \tfrac{1}{2} L_{yy}\|\yyv^{-1} f(\xv_t, \yv_{t-1})\|^2 \cdot \| \n_\yv f(\xv_t, \yv_{t-1}) \|^2 \tr
&\leq L_{yy} (2\mu_y^2)^{-1} \|\n_\yv f(\xv_t, \yv_{t-1}) - \n_\yv f(\xv^*, \yv^*)\|^2 \tr
&\leq L_{yy} (2\mu_y^2)^{-1} L_y^2 \left( \|\xv_t - \xv^*\|^2 +  \|\yv_{t-1} - \yv^*\|^2\right), 
\end{align}
where in the second line we used \eqref{eq:int_ny_gdn}; in the fifth line we used Assumption~\ref{asp:lip_hessian}; in the seventh line we used the definition of $\Delta \yv$; in the second last line we used $\|\yyv^{-1} f(\zv)\| \leq \mu_y^{-1}$ any $\zv\in \Bc(\zv^*)$ from Lemma \ref{lem:local_lip_boundedness} and $\n_\yv f(\zv^*) = \zero$; in the last line we used the Lipschitz condition in Lemma \ref{lem:local_lip_boundedness}. 
Note that $\zv_t, \zv_{t-1} \in \Bc(\zv^*)$, and thus $\yv_{t-1} \in \Nc(\yv^*)$, $\xv_t \in \Nc(\xv^*)$ and $(\xv_t, \yv_{t-1})\in \Bc(\zv^*)$. So all our discussion is within the neighborhood $\Bc(\zv^*)$ and thus valid. 
On the other hand, from $\yyv f(\zv) \cle -\mu_y \Iv$ for all $\zv\in \Bc(\zv^*)$, as in Lemma \ref{lem:local_lip_boundedness}, and the Cauchy-Schwarz inequality, we obtain:
\begin{align}\label{eq:upper_bound_ry_ny}
&\|r(\xv_t) - \yv_t\|\cdot \|\n_\yv f(\xv_t, r(\xv_t)) - \n_\yv f(\xv_t, \yv_t)\| \tr
&\geq-(r(\xv_t) - \yv_t)^\top (\n_\yv f(\xv_t, r(\xv_t)) - \n_\yv f(\xv_t, \yv_t)) \tr
&= -(r(\xv_t) - \yv_t)^\top \yyv f(\xv_t, \yv_{\xi}) (r(\xv_t) - \yv_t) \tr
&\geq  \mu_y \|r(\xv_t) - \yv_t\|^2,
\end{align}
where in the third line we used the mean-value theorem and that $\yv_{\xi}$ is on the line segment with $\yv_t$ and $r(\xv_t)$ as two endpoints; in the fourth line we used the definition of $\mu_y$ in Lemma \ref{lem:local_lip_boundedness}. 
Therefore, from  \eqref{eq:upper_bound_ry_ny}, $
\n_\yv f(\xv_t, r(\xv_t)) = \zero$ and \eqref{eq:bound_yt_xt_gdn} we obtain:
\begin{align}\label{eq:distance_rx_y}
\|r(\xv_t) - \yv_t\| &\leq \mu_y^{-1}\|\n_\yv f(\xv_t, r(\xv_t)) - \n_\yv f(\xv_t, \yv_t)\| \tr
&= \mu_y^{-1}\| \n_\yv f(\xv_t, \yv_t)\| \tr
&\leq L_{yy} (2\mu_y^3)^{-1} L_y^2 \left( \|\xv_t - \xv^*\|^2 +  \|\yv_{t-1} - \yv^*\|^2\right).
\end{align}
Combining \eqref{eq:update_x_gdn}, \eqref{eq:spec_first}, \eqref{eq:eqa18}, \eqref{eq:eqa20} we obtain that:
\begin{align}
\| \xv_{t+1} - \xv^*\| &\leq \rho_\La \| \xv_t - \xv^*\| + \a_\La L_x^\TD \| r(\xv_t) - \yv_t\| + \a_\La B_{xy} \mu_y^{-1} \| \n_\yv f(\zv_t)\| \tr
&\leq \rho_\La \| \xv_t - \xv^*\| +  \a_\La ({B_{xy} + L_x^\TD}) \mu_y^{-1} \| \n_\yv f(\zv_t)\| \tr
&\leq \rho_\La \| \xv_t - \xv^*\|  + \a_\La M(\|\xv_t - \xv^*\|^2 + \|\yv_{t - 1} - \yv^*\|^2),
\end{align}
where in the second line we used \eqref{eq:distance_rx_y} and in the third line we used \eqref{eq:bound_yt_xt_gdn}, and
\be
M = (B_{xy} + L_x^\TD) L_{yy} (2\mu_y^3)^{-1} L_y^2.
\en

\paragraph{Part III} Denote $a_t = \|\xv_t - \xv^*\|$ and $b_t = \|\yv_{t} - \yv^*\|$, we have proved the following claim in Part I and Part II:
\begin{claim}\label{claim:part1_and_part2}
Suppose for $t\geq 2$, if $\{\zv_k\}_{k=1}^t \subset \Nc_{\rm GDN}$, then we have:
\be
\label{eq:boundseq}
& a_{t+1} \leq \rho_\La a_t + M (a_t^2 + b_{t-1}^2), \, b_{t+1} \leq U b_t^2 + V a_{t+1}.
\en
\end{claim}


Suppose now that $\zv_t \in \Nc_{\rm GDN}$ for any $1\leq t\leq T$, let us prove $\zv_{T+1} \in \Nc_{\rm GDN}$. From Claim \ref{claim:part1_and_part2} we know that \eqref{eq:boundseq} holds for all $t = 2, \cdots, T$.
Define the upper bounding sequence $\{\bar{a}_{k}\}_{k=1}^{T+1}$ and $\{\bar{b}_{k}\}_{k=1}^{T+1}$ such that $\bar{a}_i = a_i$ and for $i = 1, 2$, and
\be\label{eq:recursion_ap}
\bar{a}_{t+1} = \rho_\La \bar{a}_t + M (\bar{a}_t^2 + \bar{b}_{t-1}^2), \bar{b}_{t+1} = U \bar{b}_t^2 + V \bar{a}_{t+1}, \mbox{ for }t = 2, \dots, T.
\en
One can show that for any $1 \leq t \leq {T+1}$, we have:
\be\label{eq:upper_bound_seqs}
a_t \leq \bar{a}_t,\, b_t \leq \bar{b}_t,
\en
which follows from induction. To prove $\zv_{T+1} \in \Nc_{\rm GDN}$, it suffices to show that for any $t = 2, \dots, T + 1$, we have:
\be\label{eq:assumption_of_bt_barat}
\bar{b}_t \leq 2V \bar{a_t}, \bar{a}_t \leq \d,
\en
which is true for $t = 1, 2$ from our assumption that $\bar{a}_i = a_i$ and $b_i = b_i$ for $i = 1, 2$ and the definition of $\Nc_{\rm GDN}$. This is because we can simply apply \eqref{eq:assumption_of_bt_barat} for $t = T + 1$ and use \eqref{eq:upper_bound_seqs}. Suppose \eqref{eq:assumption_of_bt_barat} holds for $k\leq t$ and $t\geq 2$:
\be\label{eq:assumption_of_bk_barak}
\bar{b}_k \leq 2V \bar{a}_k, \bar{a}_k \leq \d \mbox{ for all }k \leq t.
\en
Taking $\bar{b}_t \leq 2V \bar{a_t}$ from \eqref{eq:assumption_of_bk_barak} we obtain:
\begin{align}\label{eq:boundab}
\bar{b}_{t+1} &= V \bar{a}_{t+1} + U \bar{b}_t^2 \tr
&\leq V \bar{a}_{t+1} + 4 V^2 U \bar{a}_t^2 \tr
&\leq V\left(1 + 4\frac{V^2 U}{\rho_\La}\bar{a}_{t}\right) \bar{a}_{t+1} \tr
&\leq 2V \bar{a}_{t+1},
\end{align}
where in the third line we used $\rho_\La \bar{a}_t \leq \bar{a}_{t+1}$ that can be derived from \eqref{eq:recursion_ap}; in the last line we used the assumption in \eqref{eq:assumption_of_bk_barak} and $\bar{a}_t \leq \d \leq \tfrac{\rho_\La}{4V^2 U}$. 
Also, from \eqref{eq:recursion_ap}, we have
\begin{align}
\bar{a}_{t+1} &= \rho_\La \bar{a}_t + M (\bar{a}_t^2 + \bar{b}_{t-1}^2) \tr
&\leq \rho_\La \bar{a}_t + M (\bar{a}_t^2 + 4V^2 \bar{a}_{t-1}^2) 
\tr
&\leq \rho_\La \bar{a}_t + M \left(1 + \frac{4 V^2}{\rho_\La^2}\right)\bar{a}_t^2 \tr
&= \left(\rho_\La + M \left(1 + \frac{4 V^2}{\rho_\La^2}\right) \bar{a}_t \right) \bar{a}_t \tr
\label{eq:linear_recursion}&\leq (\rho_\La + \e)\bar{a}_t \\
&\leq \bar{a}_t \leq \d,
\end{align}
where in the second line, we used $b_{t-1} \leq 2V \bar{a}_{t-1}$ as in \eqref{eq:assumption_of_bk_barak}; in the third line, we used $\bar{a}_{t} \geq \rho_{\La}\bar{a}_{t-1}$ which can be derived from \eqref{eq:recursion_ap}; in the second last line we used the assumption in \eqref{eq:assumption_of_bk_barak} that $\bar{a}_t \leq \d \leq \tfrac{\e}{M(1 + (4V^2/\rho_\La))}$; in the last line we used $0 < \e < 1 - \rho_\La$. By induction, we have proved that for any $t = 2, \dots, T + 1$, we have \eqref{eq:assumption_of_bt_barat} and thus $\zv_{T+1} \in \Nc_{\rm GDN}$. 

So far, we have proved that for any $t \geq 1$, $\zv_{t} \in \Nc_{\rm GDN}$. This implies that for any $t \geq 2$, \eqref{eq:boundseq} is true. Taking the upper bounding sequence again as in \eqref{eq:upper_bound_seqs}. We have in fact proved from \eqref{eq:boundab} and \eqref{eq:linear_recursion} that for any $t \geq 2$, 
\be
\bar{a}_{t+1} \leq (\rho_\La + \e) \bar{a}_t, \, \bar{b}_{t + 1} \leq 2V \bar{a}_{t + 1}.
\en
Therefore, we obtain from the above that for $t \geq 2$:
\be
\bar{a}_{t+1} \leq (\rho_\La + \e)^{t - 1} \bar{a}_2 = (\rho_\La + \e)^{t - 1} a_2, 
\en
and thus for any $t \geq 2$:
\begin{align}
& \|\xv_{t+1} - \xv^*\| = a_{t + 1} \leq \bar{a}_{t+1}  \leq (\rho_\La + \e)^{t - 1} a_2 = (\rho_\La + \e)^{t - 1} \|\xv_2 - \xv^*\|, \tr
& \|\yv_{t+1} - \yv^*\| =  b_{t+1} \leq \bar{b}_{t + 1} \leq 2V \bar{a}_{t + 1} \leq 2V (\rho_\La + \e)^{t - 1} a_2 = 2V (\rho_\La + \e)^{t - 1} \|\xv_2 - \xv^*\|.
\end{align}

\end{proof}


\begin{thm}[\textbf{Complete Newton}]
\label{thm:CN2}
Given a {\slmm} $(\xv^*, \yv^*)$ and $\d_x > 0$, $\d_y > 0$, suppose on the neighborhoods $\Nc(\xv^*) = \Bc(\xv^*, \d_x) $ and $\Nc(\yv^*) = \Bc(\yv^*, \d_y)$, Assumption \ref{asp:lip_hessian} holds and the local best-response function $r: \Nc(\xv^*) \to \Nc(\yv^*)$ exists. Define the neighborhood $\Nc_{\rm CN}$ as $$\Nc_{\rm CN} := \{\zv \in \R^{n+m} : \| \zv - \zv^*\| \leq \min\{\d_x, \d_y, \tfrac{1}{3L}\}\},$$ where
\be\label{eq:lip_cn}
L = U + (V +1) (\tfrac{1}{2} \mu_x^{-1} L_{xx}^\psi+ W),
\en
Here $U$, $V$ are the same as in Theorem \ref{thm:GDN_app} and
\be\label{eq:definition_L2_theorem}
W := (L_x \mu_x^{-1} + B_x \mu_x^{-2}L_{xx}^\TD)
L_{yy} (2\mu_y^3)^{-1} L_y^2.
\en
$\mu_x, \mu_y$, $B_x$, $B_y$, $L_x$, $L_y$, $L^\TD_{x}$, $L^\TD_{xx}$, $L^\psi_{xx}$, $L_{yy}$ are defined in Lemmas \ref{lem:local_lip_boundedness}, \ref{lem:tdx_and_tdxx}, \ref{lem:Lipschitz_of_psi_derivs} and \ref{lem:inverse_of_tdxx}. The local convergence of CN to $\zv^* = (\xv^*, \yv^*)$ is at least quadratic, i.e.:
\be
\|\zv_t - \zv^*\| \leq \frac{1}{2L} \max\{2L \|\zv_1 - \zv^*\|, 2L \|\zv_2 - \zv^*\|\}^{2^{\lfloor(t - 1)/2\rfloor}}.
\en
with the initializations $\zv_1 \in \Nc_{\rm CN}, \zv_2 \in \Nc_{\rm CN}$.
\end{thm}

Before we move on to the proof. We first interpret the constant $L$. In \eqref{eq:kappa_ys} we defined the condition numbers of $\yv$ as:  
\be
\kappa_{1, y} := \frac{L_y}{\mu_y}, \, \kappa_{2, y} := \frac{L_{yy}}{\mu_y}.
\en
from which the absolute constants $U$, $V$ can be written as:
\be
U = \kappa_{2, y}/2,\, V = \kappa_{1, y}\kappa_{2, y} (\d_x + \d_y) + \kappa_{1, y}.
\en
Similarly, we define the condition numbers on $\xv$ as:
\be
\kappa_{1, x} := \frac{L_x}{\mu_x}, \, \kappa_{2, x}^\TD := \frac{L_{xx}^\TD }{\mu_x}, \, \kappa_{2, x}^\psi := \frac{L_{xx}^\psi }{\mu_x},
\en
and using Lemma \ref{lem:local_lip_boundedness}, \eqref{eq:definition_L2_theorem} can be written as:
\be
W = (\k_{1, x} + (\d_x + \d_y)\k_{1, x}\k_{2,x}^\TD) \kappa_{2, y}\kappa_{1, y}^2/2.
\en
Putting everything together, \eqref{eq:lip_cn} becomes:
\be
L = \frac{\k_{2, y}}{2} + \frac{1}{2}(\k_{1, y} \kappa_{2, y}(\d_x + \d_y) + \kappa_{1, y} + 1)(\k_{2, x}^\psi +  (\k_{1, x} + (\d_x + \d_y)\k_{1, x}\k_{2,x}^\TD) \kappa_{2, y}\kappa_{1, y}^2).
\en
This interpretation shows us that the size of the neighborhood $\Nc_{\rm CN}$ that guarantees the local quadratic convergence
can be very small, since $W$ is a product of condition numbers on both $\xv$ and $\yv$. The dependence of the neighborhood on condition numbers is not uncommon in conventional minimization, for both first- and second-order algorithms (e.g.~\citet{nesterov2003introductory}, Theorems 1.2.4 and 1.2.5). 
\begin{proof}

We assume first that $\zv_k \in \Nc_{\rm CN}$ for $k\leq t$ and $t\geq 2$. Note that $\Nc_{\rm CN} \subset \Bc(\zv^*)$ because for any $(\xv, \yv) \in \Nc_{\rm CN}$, 
\be
\|\xv - \xv^*\| \leq \|\zv - \zv^*\| \leq \d_x, \, \|\yv - \yv^*\| \leq \|\zv - \zv^*\| \leq \d_y.
\en
This satisfies our definition of $\Bc(\zv^*) = \Bc(\xv^*, \d_x) \times \Bc(\yv^*, \d_y)$ in \eqref{eq:def_Bczstar}. $\Nc_{\rm CN} \subset \Bc(\zv^*)$ tells us that we can use all the local Lipschitzness and boundedness results in Appendix \ref{app:local_bound_lipschitz}. 

Since the update of $\yv$ is the same as GDN, we can borrow \eqref{eq:first_y_gdn} to have:
\be\label{eq:first_y}
\|\yv_{t+1} - \yv^*\|\leq U \|\yv_t - \yv^*\|^2 + V \|\xv_{t+1} - \xv^*\|.
\en
We prove the next that:
\be\label{eq:second_x}
\| \xv_{t+1} - \xv^*\| &\leq (\tfrac{1}{2} \mu_x^{-1} L_{xx}^\psi+ W) \|\xv_t - \xv^*\|^2 +  W \|\yv_{t-1} - \yv^*\|^2. 
\en
where
\be
W := (L_x \mu_x^{-1} + B_x \mu_x^{-2}L_{xx}^\TD)
L_{yy} (2\mu_y^3)^{-1} L_y^2. 
\en


\paragraph{Part I} To prove \eqref{eq:second_x}, we note that:
\be\label{eq:bound_xt_xstar}
\|\xv_{t+1} - \xv^*\| &=& \|\xv_t -\xv^* - (\TD_{\xv\xv}^{-1} \cdot\n_\xv) f(\xv_t, r(\xv_t)) +  (\TD_{\xv\xv}^{-1} \cdot\n_\xv) f(\xv_t, r(\xv_t)) - (\TD_{\xv\xv}^{-1} \cdot\n_\xv) f(\xv_t, \yv_t)\|\tr
&\leq& \|\TD_{\xv\xv}^{-1}f(\xv_t, r(\xv_t))(\TD_{\xv\xv} f(\xv, r(\xv_t))(\xv_t - \xv^*) - \n_\xv f(\xv_t, r(\xv_t)))\| +\tr
&+& \|(\TD_{\xv\xv}^{-1}\cdot \n_\xv) f(\xv_t, r(\xv_t)) - (\TD_{\xv\xv}^{-1}\cdot \n_\xv) f(\xv_t, \yv_t)\|.
\en



We observe that $r(\xv_t) \in \Nc(\yv^*)$ because of \eqref{eq:neighborhoods}. So $(\xv_t, r(\xv_t)) \in \Bc(\zv^*)$ and our analysis is valid. The first term can be computed as:
\begin{align}\label{eq:bound_xt_xstar_first}
&\; \|\TD_{\xv\xv}^{-1}f(\xv_t, r(\xv_t))(\TD_{\xv\xv} f(\xv, r(\xv_t))(\xv_t - \xv^*) - \n_\xv f(\xv_t, r(\xv_t)))\|  \tr
&\leq \|\TD_{\xv\xv}^{-1}f(\xv_t, r(\xv_t))\|\cdot \|(\TD_{\xv\xv} f(\xv, r(\xv_t))(\xv_t - \xv^*) - \n_\xv f(\xv_t, r(\xv_t)))\|  \tr
& \leq \mu_x^{-1}\| \psi''(\xv_t)(\xv_t - \xv^*) - \TD_\xv f(\xv_t, r(\xv_t)) + \TD_\xv f(\xv^*, \yv^*)\| \tr
&= \mu_x^{-1}\| \psi''(\xv_t)(\xv_t - \xv^*) - \psi'(\xv_t) + \psi'(\xv^*)\| \tr
&= \mu_x^{-1}\| \psi''(\xv_t)(\xv_t - \xv^*) - \int_0^1 \psi''(\xv^* + s (\xv_t - \xv^*)) (\xv_t - \xv^*)ds \|
\tr
&= \mu_x^{-1}\| \int_0^1 (\psi''(\xv_t)-  \psi''(\xv^* + s (\xv_t - \xv^*))) (\xv_t - \xv^*)ds \| \tr
&\leq \mu_x^{-1} \int_0^1 \| \psi''(\xv_t)-  \psi''(\xv^* + s (\xv_t - \xv^*))\|\cdot \|\xv_t - \xv^* \| ds  \tr
&\leq \mu_x^{-1} L_{xx}^\psi \int_0^1 (1- s) \|\xv_t - \xv^*\|^2 ds \tr
&=  \tfrac{1}{2} \mu_x^{-1} L_{xx}^\psi  \|\xv_t - \xv^*\|^2,
\end{align}
where in the third line we used that for $\xv\in \Nc(\xv^*)$, we have from $\n_\yv f(\xv, r(\xv)) = \zero$ in \eqref{eq:neighborhoods}: 
\be
\TD_{\xv} f(\xv, r(\xv)) = \n_\xv f(\xv, r(\xv)) - (\xyv f \cdot \yyv^{-1} f \cdot \n_\yv f)(\xv, r(\xv)) = \n_\xv f(\xv, r(\xv)),
\en
and thus $\TD_{\xv} f(\xv^*, \yv^*) = \n_\xv f(\xv^*, \yv^*) = \zero$; the fifth line we used that for $\xv_1, \xv_2 \in \Nc(\xv^*)$:
\be
\psi'(\xv_1) - \psi'(\xv_2) = \int_0^1 \psi''(\xv_2 + s(\xv_1 - \xv_2)) (\xv_1 - \xv_2) ds,
\en
and in the second last line we used Lemma \ref{lem:Lipschitz_of_psi_derivs} and the definition of $L_{xx}^\psi$ in \eqref{eq:definition_psi_derivs}.

From Lemma \ref{lem:inverse_of_tdxx} we know that on $\Bc(\zv^*)$, $\TD_{\xv\xv}^{-1} f := (\TD_{\xv\xv} f(\cdot))^{-1}$ is $\mu_x^{-2}L_{xx}^\TD$-Lipschitz continuous and $\mu_x^{-1}$-bounded. From Lemma \ref{lem:local_lip_boundedness}, we know that on $\Bc(\zv^*)$, $\n_\xv f$ is $L_x$-Lipschitz continuous and $B_x$-bounded. Therefore, from Lemma \ref{lem:lip_of_product}, $\TD_{\xv\xv}^{-1} f \cdot \n_\xv f$ is $(L_x \mu_x^{-1} + B_x \mu_x^{-2}L_{xx}^\TD)$ Lipschitz continuous. The second term of \eqref{eq:bound_xt_xstar} can thus be bounded as:
\be\label{eq:bound_xt_xstar_second}
(L_x \mu_x^{-1} + B_x \mu_x^{-2}L_{xx}^\TD) \|r(\xv_t) - \yv_t\| \leq (L_x \mu_x^{-1} + B_x \mu_x^{-2}L_{xx}^\TD)
L_{yy} (2\mu_y^3)^{-1} L_y^2 \left( \|\xv_t - \xv^*\|^2 +  \|\yv_{t-1} - \yv^*\|^2\right), \en
where we used \eqref{eq:distance_rx_y}. Note that the update of $\yv_t$ is the same for both GDN and CN. To avoid heavy notation, we define
\be
W := (L_x \mu_x^{-1} + B_x \mu_x^{-2}L_{xx}^\TD)
L_{yy} (2\mu_y^3)^{-1} L_y^2.
\en
From \eqref{eq:bound_xt_xstar}, \eqref{eq:bound_xt_xstar_first} and \eqref{eq:bound_xt_xstar_second}, we obtain that:
\begin{align}
\| \xv_{t+1} - \xv^*\| &\leq \tfrac{1}{2} \mu_x^{-1} L_{xx}^\psi \|\xv_t - \xv^*\|^2 + W \left( \|\xv_t - \xv^*\|^2 +  \|\yv_{t-1} - \yv^*\|^2\right)\tr
&= (\tfrac{1}{2} \mu_x^{-1} L_{xx}^\psi+ W) \|\xv_t - \xv^*\|^2 +  W \|\yv_{t-1} - \yv^*\|^2
\end{align}

\paragraph{Part II} 
So far, we have:
\be\label{eq:first_y_again}
\|\yv_{t+1} - \yv^*\|\leq U \|\yv_t - \yv^*\|^2 + V \|\xv_{t+1} - \xv^*\|,
\en
and
\be\label{eq:second_x_again}
\| \xv_{t+1} - \xv^*\| &\leq (\tfrac{1}{2} \mu_x^{-1} L_{xx}^\psi+ W) \|\xv_t - \xv^*\|^2 +  W \|\yv_{t-1} - \yv^*\|^2.
\en
where
\be
W = (L_x \mu_x^{-1} + B_x \mu_x^{-2}L_{xx}^\TD)
L_{yy} (2\mu_y^3)^{-1} L_y^2. 
\en

Bringing \eqref{eq:second_x_again} to \eqref{eq:first_y_again}, we obtain that:
\be\label{eq:second_y_modified} 
\|\yv_{t+1} - \yv^*\|\leq U \|\yv_t - \yv^*\|^2 + V(\tfrac{1}{2} \mu_x^{-1} L_{xx}^\psi+ W) \|\xv_t - \xv^*\|^2 + V  W \|\yv_{t-1} - \yv^*\|^2,
\en
With \eqref{eq:second_x_again} and \eqref{eq:second_y_modified}, we can prove:
\begin{align}\label{eq:second_z}
\|\zv_{t+1} - \zv^*\| &\leq \|\xv_{t+1} - \xv^*\| + \|\yv_{t+1} - \yv^*\| \tr
&\leq U \|\yv_t - \yv^*\|^2 + (V + 1) (\tfrac{1}{2} \mu_x^{-1} L_{xx}^\psi+ W) \|\xv_t - \xv^*\|^2 + (V + 1) W \|\yv_{t-1} - \yv^*\|^2 \tr
&\leq U \|\zv_t - \zv^*\|^2 + (V + 1) (\tfrac{1}{2} \mu_x^{-1} L_{xx}^\psi+ W) \|\zv_t - \zv^*\|^2 + (V + 1) W \|\zv_{t-1} - \zv^*\|^2 \tr
&\leq L(\|\zv_t - \zv^*\|^2 + \|\zv_{t-1} - \zv^*\|^2),
\end{align}
where in the third line we used $\|\yv_t - \yv^*\| \leq \|\zv_t - \zv^*\|$, $\|\xv_t - \xv^*\| \leq \|\zv_t - \zv^*\|$ and $\|\yv_{t-1} - \yv^*\| \leq \|\zv_{t-1} - \zv^*\|$. Note also that we defined:
\be\label{eq:def_L}
L = U + (V +1) (\tfrac{1}{2} \mu_x^{-1} L_{xx}^\psi+ W).
\en

Now let us prove that $\zv_{t+1} = (\xv_{t+1}, \yv_{t+1})$ is still in $\Nc_{\rm CN}$. This is because from \eqref{eq:second_z}, 
\begin{align}
\|\zv_{t+1} - \zv^*\| &\leq L\|\zv_t - \zv^*\| \cdot \|\zv_t - \zv^*\| +  L\|\zv_{t-1} - \zv^*\| \cdot \|\zv_{t-1} - \zv^*\|  \tr
 &\leq L\cdot \tfrac{1}{3L}\cdot \|\zv_t - \zv^*\| +  L\cdot \tfrac{1}{3L}\cdot \|\zv_{t-1} - \zv^*\| \tr
 &= \tfrac{1}{3}\|\zv_t - \zv^*\|  + \tfrac{1}{3}\|\zv_{t-1} - \zv^*\|  \tr
 &\leq  \tfrac{1}{3} \min\{\d_x, \d_y, \tfrac{1}{3L}\}  + \tfrac{1}{3}\min\{\d_x, \d_y, \tfrac{1}{3L}\}\tr
 &\leq \min\{\d_x, \d_y, \tfrac{1}{3L}\},
\end{align}
where in the second line we used that $\|\zv_t - \zv^*\|\leq \tfrac{1}{3L}$ and $\|\zv_{t-1} - \zv^*\|\leq \tfrac{1}{3L}$ and in the fourth line we used the assumption $\|\zv_{t} - \zv^*\| \leq \min\{\d_x, \d_y, \tfrac{1}{3L}\}$ and $\|\zv_{t-1} - \zv^*\| \leq  \min\{\d_x, \d_y, \tfrac{1}{3L}\}$. These results follow from our assumption $\zv_t, \zv_{t-1} \in \Nc_{\rm CN}$ from induction. Therefore, we have proved that  $\{\zv_t\}_{t=1}^\infty \subset \Nc_{\rm CN}$ given $\zv_1, \zv_2 \in \Nc_{\rm CN}$.

Denote $u_t = \|\zv_{t} - \zv^*\|$, we have:
\be
u_{t+1} \leq L(u_t^2 + u_{t-1}^2),
\en
as in \eqref{eq:second_z} for $t\geq 2$. Multiplying both sides by $2L$, we have:
\be\label{eq:recursion_ineq_of_ut}
2L u_{t+1} \leq \frac{(2Lu_t)^2 + (2Lu_{t-1})^2}{2}.
\en
Define $v_t = 2L u_t$ for $t \geq 1$ and let us prove by induction that for any $k \geq 1$, we have:
\be\label{eq:recursion_ineq_of_vk}
v_k \leq q^{2^{\lfloor(k - 1)/2\rfloor}}, \, q = \max\{2Lu_1, 2L u_2\},
\en
which is true for $k = 1, 2$. Since $\zv_1, \zv_2 \in \Nc_{\rm CN}$, we have $u_1 = \|\zv_1 - \zv^*\| \leq \tfrac{1}{3L}$ and $u_2 = \|\zv_2 - \zv^*\| \leq \tfrac{1}{3L}$, and thus $q < 1$. Suppose \eqref{eq:recursion_ineq_of_vk} is true for $k \leq t$ and $t \geq 2$, then from \eqref{eq:recursion_ineq_of_ut} we can obtain:
\begin{align}
v_{t+1} &\leq \frac{1}{2}\left( (q^{2^{\lfloor(t - 1)/2\rfloor}})^2 + (q^{2^{\lfloor(t - 2)/2\rfloor}})^2\right) \tr
&= \frac{1}{2}\left(q^{2^{\lfloor(t + 1)/2\rfloor}} + q^{2^{\lfloor t/2\rfloor}}\right) \tr
&\leq \frac{1}{2}\left(q^{2^{\lfloor t/2\rfloor}} + q^{2^{\lfloor t/2\rfloor}}\right) \tr
&= q^{2^{\lfloor (t+1 - 1)/2\rfloor}},
\end{align}
where in the third line we used $q < 1$ and $\lfloor\tfrac{t+1}{2}\rfloor \geq \lfloor\tfrac{t}{2}\rfloor$. So, we have proved by induction that for any $t\geq 1$, the following holds:
\be
2L u_t = v_t \leq q^{2^{\lfloor(t - 1)/2\rfloor}}, \, q = \max\{2Lu_1, 2L u_2\},
\en
namely, for any $t \geq 1$, we have
\be
\|\zv_t - \zv^*\| \leq \frac{1}{2L} \max\{2L \|\zv_1 - \zv^*\|, 2L \|\zv_2 - \zv^*\|\}^{2^{\lfloor(t - 1)/2\rfloor}}.
\en

\end{proof}

%% file: appendices/E-2ts-gda.tex
\section{Asymptotic analysis of related algorithms}\label{app:asymptotic_related_proof}

In this appendix we analyze two-time-scale GDA (2TS-GDA) (\Cref{app:grad_alg}), TGDA, FR and GDA-$k$, as mentioned in Sections \ref{sec:tgda_fr_main} and \ref{sec:gda_main}. We show that TGDA/FR are both approximations of GDN (\Cref{sec:tgda_fr}) and that TGDA and FR are ``transpose'' of each other. We give asymptotic proofs of related algorithms. The following lemma will be needed.

\begin{lem}\label{lem:product}
Given $f: \R^d\to \R^{n\times m}$ and $g:\R^d\to \R^m$, assume  $g$ is Fr\'{e}chet differentiable at $\zv$ and $g(\zv) =  \zero$, and $f$ is continuous at $\zv$. Then, the product function $h = fg$ is Fr\'{e}chet differentiable at $\zv$ with $h'(\zv) = f(\zv)g'(\zv)$.
\end{lem}

\begin{proof}
It suffices to prove that $\|h(\zv + \deltav) - h(\zv) - f(\zv)g'(\zv)^\top \deltav \| = o(\|\deltav\|)$. This is because:
\begin{align}
&\quad\| h(\zv + \deltav) - h(\zv) - f(\zv)g'(\zv)^\top \deltav \| \tr
&= \| f(\zv + \deltav)g(\zv + \deltav) - f(\zv)g(\zv) - f(\zv)g'(\zv)^\top \deltav \| \tr
&= \| f(\zv + \deltav)g(\zv + \deltav) - f(\zv)g(\zv + \deltav) + f(\zv)g(\zv + \deltav) - f(\zv)g(\zv) - f(\zv)g'(\zv)^\top \deltav \| \tr
&\leq \|(f(\zv + \deltav) -f(\zv))g(\zv  + \deltav)\| + \|f(\zv)(g(\zv + \deltav) - g(\zv) - g'(\zv)^\top \deltav)\|
\tr
&\leq \|f(\zv + \deltav) - f(\zv)\| \cdot \|g(\zv  + \deltav)\| + \|f(\zv)\|\cdot \|g(\zv + \deltav) - g(\zv) - g'(\zv)^\top \deltav\|
\tr
&\leq o(1) \cdot  \|g(\zv+\deltav) - g(\zv) \| + o(\|\deltav\|)
\tr
&= o(\|\deltav\|),
\end{align}
where in the second last line, we used $g(\zv) = \zero$, the continuity of $f$ and the Fr\'{e}chet differentiability of $g$. 
\end{proof}

\subsection{Total gradient descent ascent and Follow-the-ridge}\label{sec:tgda_fr}
\paragraph{Total gradient descent ascent (TGDA)} 
Fiez \etal~\citep{fiez2019convergence} proposed TGDA with \Fa being gradient descent for the follower and \La being \emph{total} gradient ascent for the leader:
\be
\label{eq:fiezt}
\xv_{t+1} = \xv_t - \a_{\La} \cdot \TD_\xv f (\xv_t, \yv_t),\qquad \yv_{t+1} = \yv_t + \a_{\Fa} \cdot \n_\yv f(\xv_t, \yv_t),
\en
where we use the total gradient $\TD$ instead of the partial derivative $\n_{\xv}$ for the update on the leader $\xv$. Its continuous dynamics was studied in \citet{evtushenko1974iterative} with linear convergence proved. 
Interestingly, we now show that TGDA can be derived as a first-order approximation of GDN. Indeed, suppose  in GDN we perform the Newton update on the follower first: $\yv_{t+1} = \yv_t - (\n^{-1}_{\yv\yv} \cdot \n_{\yv}) f(\xv_t, \yv_t)$, and then we perform the (usual) gradient update on the leader $\xv$:
\begin{align}
\label{eq:tgda_newton}\xv_{t+1} &= \xv_t - \a_{\La} \cdot \n_{\xv} f(\xv_t, \yv_{t+1}) = \xv_t - \a_{\La} \cdot \n_{\xv} f\big(\xv_t, \yv_t - (\n^{-1}_{\yv\yv} \cdot \n_{\yv}) f(\xv_t, \yv_t) \big) \\
&\approx  \xv_t - \a_{\La} \cdot \big(\n_{\xv} - \n_{\xv\yv} \cdot \n^{-1}_{\yv\yv} \cdot \n_{\yv} \big) f(\xv_t, \yv_t) = \xv_t - \a_{\La} \cdot \TD_\xv f(\xv_t, \yv_t),
\end{align}
where we performed first-order expansion of $\n_{\xv} f$ \wrt $\yv$. Thus, TGDA approximates GDN in two aspects: (1) it performs a first-order approximation of the update on the leader $\xv$; (2) it replaces the Newton update on the follower $\yv$ with a gradient update in \eqref{eq:fiezt}.

A similar idea with TGDA is \emph{unrolled GDA} \citep{metz2016unrolled}, where the Newton step in \eqref{eq:tgda_newton} is replaced with $k$ gradient ascent steps.


\paragraph{Follow the ridge (FR)}  Follow-the-ridge was proposed in \citet{evtushenko1974iterative} and its variant is recently studied by \citet{wang2019solving}. In this algorithm, \Fa is a pre-conditioned gradient update for the follower and \La is the usual gradient update for the leader:
\be
\label{eq:wang}
\xv_{t+1} = \xv_t - \a_{\La} \cdot \n_\xv f(\xv_t, \yv_t), \qquad \yv_{t+1} = \yv_t + (\a_{\Fa} \cdot \n_\yv + \a_{\La} \cdot \n_{\yv\yv}^{-1} \cdot \n_{\yv\xv} \cdot \n_\xv)  f(\xv_t, \yv_t). 
\en
Similarly as TGDA, FR can also be derived as an approximation of GDN. Indeed, suppose in GDN we perform the usual gradient update on the follower first, and then we perform the Newton update on the follower $\yv$ using the newly updated $\xv_{t+1}$ (in $\n_{\yv} f$): 
\begin{align} 
\yv_{t+1} &= \yv_t - \n_{\yv\yv}^{-1} \cdot \n_{\yv} f(\xv_{t+1}, \yv_{t}) = \yv_t - \n_{\yv\yv}^{-1} \cdot \n_{\yv} f\big(\xv_{t} - \a_{\La} \cdot \n_{\xv} f(\xv_t, \yv_t), \yv_{t} \big) \\
&\approx  \yv_t - (\n_{\yv\yv}^{-1} \cdot \n_{\yv}) f(\xv_t, \yv_t) + \a_{\La} \cdot (\n_{\yv\yv}^{-1} \cdot \n_{\yv\xv} \cdot \n_{\xv}) f(\xv_t, \yv_t),
\end{align}
where in the last line we performed first-order expansion of $\n_{\yv}$ \wrt $\xv$. Thus, FR also approximates GDN in two aspects: (1) it performs a first-order approximation of the update on the follower $\yv$; (2) it replaces the Newton part on the resulting approximation with a gradient update in \eqref{eq:wang}.

In fact, it is not a coincidence that both TGDA and FR can be derived as first-order approximations of GDN---the two are in some sense ``transpose'' of each other. Indeed, denote $\zv = (\xv, \yv)$ and 
\be\label{eq:procondition}
\Pv =\begin{bmatrix}
-\a_{\La} \Iv & \a_{\La} (\xyv \cdot \yyv^{-1}) f \\
\zero & \a_{\Fa} \Iv
\end{bmatrix}, \, \n_\zv f = \begin{bmatrix} \n_\xv f \\ \n_\yv f \end{bmatrix}.
\en
Then, we can equivalently rewrite TGDA and FR respectively as: 
\begin{align}
\label{eq:equiv}
&\text{TGDA}: \zv_{t+1} = \zv_t + \Pv \cdot \n_\zv f(\zv_t), \\
&\text{FR}: \zv_{t+1} = \zv_t + \Pv^{\top} \cdot \n_\zv f(\zv_t).
\end{align} 
In other words, the two algorithms amount to performing some pre-conditioning on GDA, and their preconditioning operators are simply transpose of each other. Since the preconditioning operator $\Pv$ is (block) triangular, it follows that TGDA and FR have the same Jacobian spectrum around a \slmm.

\tgdaFr*
\begin{proof}
With Lemma~\ref{lem:product} we compute the Jacobian at $(\xv^*, \yv^*)$:
\be
\Jv_{\rm TGDA} = \begin{bmatrix}
\Iv - \a_\La (\xxv - \xyv \cdot \yyv^{-1} \cdot \yxv)f & \zero \\
\a_\Fa \yxv f & \Iv + \a_\Fa \yyv f
\end{bmatrix}
\en
The spectral radius can be easily computed as:
\be
\rho(\Jv_{\rm TGDA}) = \max_i |1 - \a_\La \l_i| \vee \max_j |1 - \a_\Fa \mu_j|.
\en

From \citet[Theorem 2.3]{zhang2019convergence}, alternating TGDA has the same convergence rate as simultaneous TGDA. Now let us show that the Jacobian of FR has the same spectrum as TGDA. From \eqref{eq:procondition} and the comment below we know that
\be
\Jv_{\rm TGDA} = \Iv + \Pv \Hv f(\xv^*, \yv^*), \, \Jv_{\rm FR} = \Iv + \Pv^\top \Hv f(\xv^*, \yv^*),
\en
where
$$
\Pv =\begin{bmatrix}
-\a_{\La} \Iv & \a_{\La} (\xyv \cdot \yyv^{-1}) f \\
\zero & \a_{\Fa} \Iv
\end{bmatrix}, \mbox{ and }
\Hv = \begin{bmatrix}
\xxv f & \xyv f \\
\yxv f & \yyv f
\end{bmatrix}.
$$
For simplicity we ignore the argument $(\xv^*, \yv^*)$. With the similarity transformation $\Pv^{-1}\Pv \Hv \Pv = \Hv\Pv$, we know that $\Pv \Hv$ has the same spectrum as $\Hv \Pv$, and also its transpose $(\Hv\Pv)^\top = \Pv^\top \Hv$. 

The optimal convergence rate is achieved by optimizing $\max_i |1 - \a_\La \l_i|$ and $\max_i |1 - \a_\Fa \mu_i|$ respectively, which is achieved at $\a_\La = 2/(\l_1 + \l_n)$ and $\a_\Fa = 2/(\mu_1 + \mu_m)$. 
\end{proof}



\subsection{GDA and its variants}\label{app:grad_alg}


Using results from \citet{jin2019minmax} and similar notations as in Theorem~\ref{thm:tgda_fr}, we derive the following result for 2TS-GDA:

\begin{theoremEnd}[]{thm}\label{thm:2ts-gda}
Around a {\slmm} $(\xv^*, \yv^*)$, for any $\d > 0$, $\exists \, \gamma_0 > 0$ such that for any $\g > \g_0$, $\a_{\Fa} > 0$ and $\a_{\La} = \a_{\Fa}/\gamma$, 2TS-GDA has asymptotic linear convergence rate $\rho = \rho_{\La} \vee \rho_{\Fa}$, where $\rho_{\La} := (|1 - \a_{\La} \l_1| + \a_{\La} \d) \vee (|1 - \a_{\La} \l_n| + \a_{\La} \d)$ and $\rho_{\Fa} := (|1 - \a_{\Fa}  \mu_1| + \a_{\Fa} \d) \vee  (|1 - \a_{\Fa}  \mu_m| + \a_{\Fa} \d)$. 
\end{theoremEnd}
\begin{proofEnd}
The Jacobian of 2TS-GDA \eqref{eq:ttsgd} at $(\xv^*, \yv^*)$ is:
\be
\Iv + \a_\Fa\begin{bmatrix}
-\g^{-1} \xxv f& -\g^{-1} \xyv f \\
\yxv f & \yyv f
\end{bmatrix} =: \Iv + \a_\Fa \Hv.
\en
Using \citet[Lemma 36]{jin2019minmax}, for any $\d > 0$, there exist $\g > 0$ large enough, s.t.~the eigenvalues of $\Hv$, $\nu_1, \dots, \nu_n, \nu_{n+1}, \dots, \nu_{m+n}$ satisfy:
\be\label{eq:close}
|\nu_i + \l_i/\g| < \d/\g,\, \forall i = 1, \dots, n,|\nu_{j + n} + \mu_j| < \d, \, \forall j = 1, \dots, m,
\en
where $\l_i \in \Sp((\n_{\xv \xv} - \n_{\xv \yv} \cdot \n_{\yv \yv}^{-1}\cdot \n_{\yv \xv}) f)$ and $\mu_j \in \Sp(-\n_{\yv \yv} f)$. The spectral radius is then:
\be
\max_{k\in [n+m]} |1 + \a_\Fa \nu_k| = \max_{i\in [n]} |1 + \a_\Fa \nu_i| \vee \max_{j\in [m]} |1 + \a_\Fa \nu_{j+n}|.
\en
We can use triangle inequality and \eqref{eq:close} to obtain that for any $\g\geq \g_0$:
\be
|1 + \a_\Fa \nu_i| \leq |1 - \a_\Fa \mu_i/\g| + \a_\Fa \d/\g  = |1 - \a_\La \mu_i| + \a_\La \d, \, \forall i\in [n].
\en
Similarly, $|1 + \a_\Fa \nu_{j+n}|\leq |1 - \a_\Fa \mu_j| + \a_\Fa \d$.
\end{proofEnd}

\Ugda*
\begin{proof}
The Jacobian matrix of the simultaneous version (replacing $\xv_{t+1}$ with $\xv_t$ in the update of $\yv$) update at $(\xv^*, \yv^*)$ is:
\be
\Jv_k = \begin{bmatrix}
\Iv - \a \xxv & -\a \xyv f\\
\a \sum_{i=0}^{k-1} (\Iv + \a \yyv f)^i \yxv  f& (\Iv + \a \yyv f)^k
\end{bmatrix}.
\en
This is because $g^{(k)}(\yv)$, the update in GDA-$k$, can be written iteratively:
\be
g^{(1)} = g(\xv_t, \yv), \, \dots, g^{(k)} = g(\xv_t, g^{(k-1)}),
\en
where $g(\xv_t, \yv) := \yv + \a \n_\yv f(\xv_t, \yv)$.
We verify that the total derivative follows $\dv{g^{(k)}}{\xv} = \n_\xv g + \n_\yv g \cdot \dv{g^{(k - 1)}}{\xv}$, and prove the derivative over $\xv_t$ by induction. 
\be
\sum_{i=0}^{\infty} (\Iv + \a \yyv f)^i =  (-\a \yyv f)^{-1}, \mbox{ and }(\Iv + \a \yyv f)^k \to \zero.
\en
Note that the series converges iff $|1 - \a \mu_j| < 1$ for all $\mu_j\in \Sp(-\yyv f)$ \citep[e.g.][Chapter 7]{meyer2000matrix}, i.e. $\alpha < 2/\max_j \mu_j = 2/\mu_1$. Under this condition, 
\be
\Jv_\infty = \begin{bmatrix}
\Iv - \a \xxv f& -\a \xyv f \\
-(\yyv^{-1}\cdot \yxv) f & \zero
\end{bmatrix}.
\en
Using \citet[Theorem 2.3]{zhang2019convergence}, the characteristic polynomial of GDA-$\infty$ is:
\be
\det\begin{bmatrix}
(\l - 1)\Iv + \a \xxv  f & \a \xyv f \\
\l (\yyv^{-1}\cdot \yxv) f  & \l \Iv
\end{bmatrix} = 0.
\en
Solving the eigenvalues yields $1 - \a \l_i$ with $\l_i \in \Sp(\TD_{\xv\xv} f)$. 

The optimal convergence rate is achieved by optimizing $\max_i |1 - \a \l_i|$, which is achieved at $\a = 2/(\l_1 + \l_n)$. However, we also impose $\a < 2/\mu_1$, which yields the assumption that $\mu_1 < \l_1 + \l_n$. Otherwise, a suboptimal rate is obtained via taking $\a\to 2/\mu_1$.

We note that it is possible to modify GDA-$k$ to be two-time-scale as well, i.e.,
\be\label{eq:gdak_2ts}
\xv_{t+1} = \xv_t - \a_\La \cdot \n_{\xv}f(\xv_t, \yv_t), \qquad \yv_{t+1} = g^{(k)}(\yv_t)\mbox{ with }g(\yv) = \yv + \a_\Fa \cdot \n_{\yv}f(\xv_{t+1}, \yv).
\en
With this modification, it suffices that $\a_\Fa < 2/\mu_1$ and the optimal rate is $1 - 2/(\kappa_\La + 1)$ with $\a_\La = 2/(\l_1 + \l_n)$. We do not need the constraint that $\mu_1 < \l_1 + \l_n$ and there is no suboptimal rate. However, when $\yyv$ is ill-conditioned the number of follower steps might be very large to approximate GDA-$\infty$.
\end{proof}

%% file: appendices/C-exp.tex
\section{Experimental details}\label{app:exp}

In this section, we report experimental details.


\subsection{Compute second-order derivatives and their inverses}\label{app:impltsn}

In our implementation, all hessian-vector products are computed via auto-differentiation.
For example, the product between $\yyv f(\xv, \yv)$ and any vector $\uv$ can be computed by the following trick:
\begin{align*}
    \partial_{\yv\yv} f(\xv, \yv) \cdot \uv = \pdv{}{\yv} \left( \partial_{\yv} f(\xv, \yv)^\top \uv \right),
\end{align*}
which allows us to compute any hessian-vector product in linear time and space (\wrt the number of parameters). By changing the order of one differentiation with the dot-product we can avoid the quadratic blow up~\citep{werbos1988backpropagation,pearlmutter1994fast}. This trick is a dual equivalent of another trick  used in Hessian-free optimization~\citep{pearlmutter1994fast,Martens10}.
Hessian-vector-product for $\partial_{\xv\xv}$, $\partial_{\xv\yv}$ and $\partial_{\yv\xv}$ can be computed efficiently in a similar way.







To efficiently implement Newton's methods, all matrix inverses are computed by least squares via the conjugate gradient (CG) method, and each CG step requires linear time.
However, addition effort is necessary for complete Newton.
Recall the complete Newton's update rule on $\xv$:
\begin{align}
    \xv_{t + 1} = \xv_{t} - \left(\left(\xxv - \xyv\cdot \yyv^{-1}\cdot\yxv\right)^{-1}f \cdot \partial_{\xv} f \right)(\xv_t, \yv_t)
\end{align}
Inverting $\partial_{\yv\yv} f $ and then inverting $(\xxv - \xyv \cdot\yyv^{-1}\cdot\yxv) f$ is not only time consuming (which involves two loops of conjugate gradient), but also numerically unstable.
In practice, there is a better way to compute the inverse of the Schur complement.
\begin{lem}
\label{lma:inverse_schur_complement}
If $\Dv$ and $\Sv := \Av - \Bv \Dv^{-1} \Cv $ are invertible, then the matrix $[\Av \, \Bv; \Cv \, \Dv]$ is invertible, with:
\be\label{eq:smart_formula}
\begin{bmatrix}
\Av & \Bv \\
\Cv & \Dv
\end{bmatrix}^{-1} = 
\begin{bmatrix}
\Sv^{-1} & -\Sv^{-1}\Bv \Dv^{-1} \\
-\Dv^{-1}\Cv \Sv^{-1} & \Dv^{-1} + \Dv^{-1}\Cv \Sv^{-1}\Bv \Dv^{-1}
\end{bmatrix}.
\en
\end{lem}
\begin{proof}
Multiply $[\Av, \Bv;\Cv, \Dv]$ with the right hand side of \eqref{eq:smart_formula} and use simple algebra.  
\end{proof}
With lemma~\ref{lma:inverse_schur_complement}, we directly invert a larger matrix
$\begin{bmatrix}
\partial_{\xv\xv}f & \partial_{\xv\yv}f \\
\partial_{\yv\xv}f & \partial_{\yv\yv}f
\end{bmatrix}
$.
The upper left block of its inverse is exactly the inverse of the Schur complement.

\subsection{Synthetic example} \label{app:synthetic}

\begin{figure}
    \centering
    \begin{subfigure}{0.36\textwidth}
    \includegraphics[width=\textwidth]{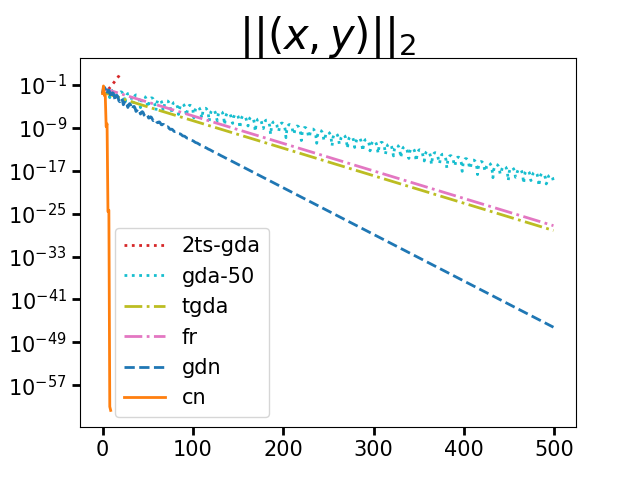}
    \end{subfigure}
    \caption{
    Synthetic example. The initialization is $\xv_0 = (0.02, 0.04)$ and $\yv_0 = (0.03, 0.05)$. For 2TS-GDA, TGDA and FR we choose $\a_\La = 0.08$ and $\a_\Fa = 0.5$; for GDA-$50$ we choose $\a = 0.08$; for GDN we choose $\a_\La = 0.08$. Convergence on learning Gaussian distributions.
    }
    \label{fig:qv}
\end{figure}

We use a synthetic example to further demonstrate the fast convergence of Newton-type methods. Consider the following minimax optimization problem:
\be
f(\xv, \yv) &=& \xv^\top \begin{bmatrix}
-2.5 & 0 \\
0 & -0.025
\end{bmatrix}\xv + \yv^\top \begin{bmatrix}
-0.5 & 0 \\
0 & -0.05
\end{bmatrix}\yv + \xv^\top  \begin{bmatrix}
0 & 1 \\
1 & 0
\end{bmatrix}\yv
\tr
&-& 0.01 (y_1^4 + y_2^4) + 0.3 x_1^4 + 0.2 x_2^4 - x_1^3 y_2.
\en
It has a {\slmm} at $(\zero, \zero)$ which is not a saddle point.
At this point, $\yyv$ is ill-conditioned.
\Cref{fig:qv} shows that Newton-type methods can cope with the ill-conditioning well. It can be seen from \Cref{fig:qv} that 2TS-GDA diverges, showing its instability and difficulty to tune. TGDA/FR converge at a similar rate as expected from Theorem~\ref{thm:tgda_fr}, both slower than GDN. Specifically, CN achieves the {\slmm} in very few steps, confirming the superlinear convergence shown in Theorem~\ref{thm:CN}. We also compare GDA-$50$ with GDN. It can be seen from \Cref{fig:qv} that GDA-$50$ does not suffice to approximate GDN with $50$ GA steps since this problem is ill-conditioned. For fair comparison we choose $\a_\La = 0.08$ and $\a_\Fa = 0.5$ for all algorithms. It is not surprising that TGDA and FR have similar performance since they are ``transpose'' of each other (see Section~\ref{sec:tgda_fr_main} and Section~\ref{sec:tgda_fr}).

\subsection{Learning the mean of a single Gaussian}\label{app:mean}

We consider a simple special case of GAN training:
\begin{align}
\label{eq:gan_single_gaussian}
    \min_{\bm\eta \in \R^2} \max_{{\bm \omega}\in \R^2} ~ \ell(\etav, \omegav) := \E_{\xv \sim \Nc({\bm \zero}, \mathbf\Sigma)} \log \sigma({\bm\omega}^\top \xv) + \E_{\zv \sim \Nc(\zero, \mathbf\Sigma)} \log\left(1 - \sigma({\bm \omega}^\top (\zv + {\bm \eta}))\right),
\end{align}
where the discriminator is a linear classifier:
\begin{align}
    D(\xv) = \sigma\left({\bm \omega}^\top \xv\right),
\end{align}
and the generator is a translation:
\begin{align}
    G(\zv) = \zv + \bm\eta.
\end{align}
The problem is concave-concave. 
It is easy to check that $(\bm\eta^*, {\bm \omega}^*) = (\mathbf 0, \mathbf 0)$ is a global minimax point.
We have the gradients:
\begin{align}
    \partial_{\bm\eta} \ell(\bm\eta, \bm \omega) & = - \E_{\zv \sim \Nc(\mathbf0, \mathbf\Sigma)} \sigma(\bm\omega^\top (\zv + {\bm \eta})) \bm\omega
\\
    \partial_{\bm\omega} \ell(\bm\eta, \bm \omega) & = \E_{\xv \sim \Nc(\mathbf0, \mathbf\Sigma)} (1 - \sigma(\bm\omega^\top \xv)) \xv - \E_{\zv \sim \Nc(\mathbf0, \mathbf\Sigma)} \sigma(\bm\omega^\top (\zv + {\bm \eta})) (\zv + \etav )
\end{align}
and the partial Hessians:
\begin{align}
    \partial_{\bm\eta\bm\eta} \ell(\bm\eta, \bm \omega) & = - \E_{\zv \sim \Nc(\mathbf0, \mathbf\Sigma)} \sigma(\bm\omega^\top (\zv + {\bm \eta})) (1 - \sigma(\bm\omega^\top (\zv + {\bm \eta}))) \bm\omega \bm\omega^\top
\\
    \partial_{\bm\omega\bm\omega} \ell(\bm\eta, \bm \omega) & = - \E_{\xv \sim \Nc(\mathbf0, \mathbf\Sigma)} \sigma(\bm\omega^\top \xv) (1 - \sigma(\bm\omega^\top \xv)) \xv \xv^\top, \\
    & ~ ~ ~ ~ - \E_{\zv \sim \Nc(\mathbf0, \mathbf\Sigma)} \sigma(\bm\omega^\top (\zv + {\bm \eta})) (1 - \sigma(\bm\omega^\top (\zv + {\bm \eta}))) (\zv + \bm\eta) (\zv + \bm\eta)^\top,
\\
    \partial_{\bm\eta\bm\omega}\ell(\bm\eta, \bm \omega) & = - \E_{\zv \sim \Nc(\mathbf0, \mathbf\Sigma)} (\sigma(\bm\omega^\top (\zv + {\bm \eta})) \Iv + \sigma'(\bm\omega^\top (\zv + {\bm \eta})) (\zv + \etav)\omegav^\top).
\end{align}
At $(\bm\eta^*, {\bm \omega}^*)$, we have
\begin{align}
    \n_{\bm\eta\bm\eta} \ell (\bm\eta^*, {\bm \omega}^*)& = \zero, \, \n_{{\bm \omega}{\bm \omega}}\ell (\bm\eta^*, {\bm \omega}^*) = - \frac12 \mathbf\Sigma, \, \n_{\bm\eta\bm \omega} \ell (\bm\eta^*, {\bm \omega}^*)=  - \frac12 \Iv,
\end{align}
and thus this point is a \slmm. 
In particular, the covariance of data distribution determines the condition of $\bm\omega$.
We compare convergence speed in two cases: a well-conditioned covariance
\begin{align}
    \mathbf\Sigma = \Iv
\end{align}
and an ill-conditioned covariance
\begin{align}
\mathbf\Sigma =
\begin{bmatrix}
1 & 0 \\
0 & 0.05
\end{bmatrix}.
\end{align}
We set $\alpha_\La = 0.05$, $\alpha_\Fa = 0.5$ for all algorithms.
For GDA-$k$ we set $\alpha = 0.05$.
We run conjugate gradient for up to $8$ iterations and terminate it whenever the norm of residual is smaller than $10^{-40}$.
The size of training data is $10000$.
We random initialize all algorithms using a zero-mean Gaussian with standard deviation $0.1$. 

\subsection{Learning the covariance of a single Gaussian}\label{app:single_gaussian}

Now let us consider learning the covariance of a Gaussian:
\begin{align}
\label{eq:gan_covariance}
    \min_{\Vv \in \R^{2\times 2}} \max_{\Wv\in \R^{2\times 2}} ~ \ell(\Vv, \Wv) := \E_{\xv \sim \Nc({\bm \zero}, \mathbf\Sigma)} \log \sigma\left(\xv^\top \Wv \xv\right) + \E_{\zv \sim \Nc(\zero, \mathbf I )} \log\left(1 - \sigma\left(\zv^\top \Vv^\top \Wv \Vv\zv\right)\right),
\end{align}

with $\xv\in \R^2$ and $\zv\in \R^2$. The generator is $G(\zv) = \Vv \zv$ and the discriminator is $D(\xv) = \s(\xv^\top \Wv \xv)$. The optimal solution satisfies $\Vv\Vv^\top = \Sigmav = \diag(1, 0.04)$ and $\Wv + \Wv^\top = \zero$.


We set $\alpha_\La = 0.02$, $\alpha_\Fa = 0.2$ for all algorithms.
For GDA-$k$ we set $\alpha = 0.02$.
We run conjugate gradient for up to $16$ iterations and terminate it whenever the norm of residual is smaller than $10^{-30}$.
The size of training data is $10000$.
A $\ell_2$ norm regularization is added on the discriminator and the regularization coefficient is $10^{-5}$.
We random initialize all algorithms using a zero-mean Gaussian with standard deviation $0.01$.

\subsection{Learning mixture of Gaussians}\label{app:gmm}

Both the discriminator and the generator are 3-hidden-layer ReLU networks with 256 neurons in each hidden layer. 
The latent variable $\zv$ is sampled from a 100 dimensional standard Gaussian distribution.
The size of training data is $10000$. We first use GDA ($\a_\La = \a_\Fa = 0.01$) with batch size 256 to find the initialization for other methods.
TGDA, FR and GD-Newton use $\a_\La = 0.01$ and $\a_\Fa = 0.02$. We run conjugate gradient for 20 iterations to solve linear systems and terminate it whenever the norm of residual is smaller than $10^{-40}$. For CN, we choose the damping coefficient $\g = 0.1$ (see \eqref{eq:regularization_cn}) with 20 CG iterations for the discriminator, and 32 CG iterations for the generator. We also add a regularization factor $\l = 0.1$ for the generator as in \eqref{eq:regularization_cn}. 

\subsection{MNIST}

Both the discriminator and the generator are 2-hidden-layer LeakyReLU ($\mathtt{negative\_slope}=0.2$) networks with 512 neurons in each hidden layer, and we add the \texttt{tanh} activation for the last layer of the generator. The latent variable $\zv$ is sampled from a 100 dimensional standard Gaussian distribution. For GDA-$20$ and EG we take $\a= 0.01$; for TGDA/FR/GDN, we take $\a_\La = 0.01$, $\a_\Fa = 0.02$ and use 16 CG iterations; for CN, we use 8 CG iterations on the generator and 16 CG iterations on the discriminator. We terminate CG  whenever the norm of residual is smaller than $10^{-50}$.

\subsection{Comparison with Adaptive gradient methods}\label{app:adaptive}

\begin{figure}[t]\label{fig:adaptive}
\centering
\begin{subfigure}[b]{0.36\textwidth}
    \includegraphics[width=\textwidth]{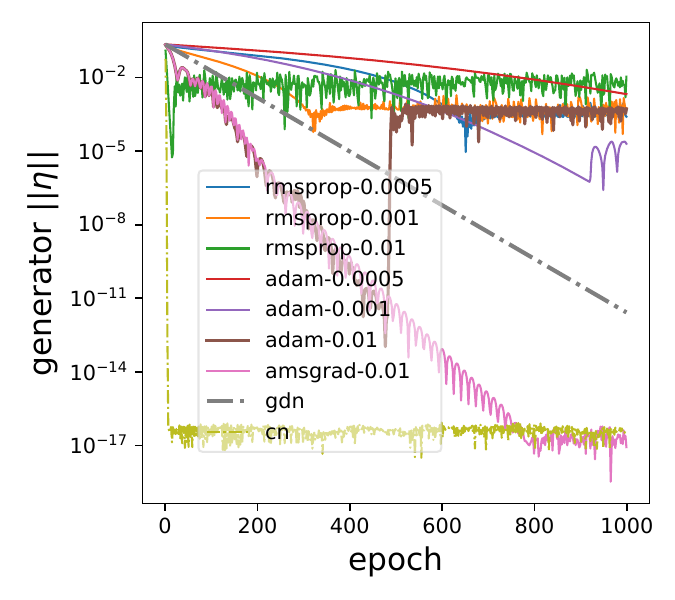}
\caption{well-conditioned}
\end{subfigure}
\begin{subfigure}[b]{0.36\textwidth}
    \includegraphics[width=\textwidth]{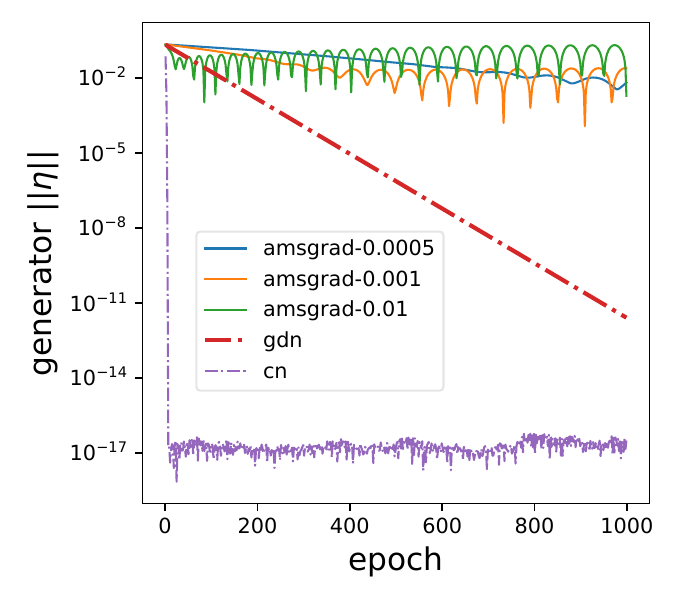}
\caption{ill-conditioned}
\end{subfigure}
\caption{Estimating Gaussian mean using adaptive methods and Newton's method.}
\label{fig:adaptive_gradient_on_single_gaussian}
\end{figure}

We compare our Newton's methods, GDN an CN, with existing adaptive methods such as RMSProp \citep{rmsprop}, Adam \citep{kingma2014adam} and AMSGrad \citep{reddi2018convergence} in \Cref{fig:adaptive_gradient_on_single_gaussian}.
With careful tuning of step sizes, Adam and RMSProp do not converge, rotating around the local minimax point.
In fact, even for minimization problems, RMSProp and Adam may not converge \citep{reddi2018convergence}.
Another adaptive method, AMSGrad, does converge and is even faster than GDN with careful tuning in the well-conditioned setting, but slows down severely in the ill-conditioned case.
It is unclear whether adaptive methods have good preconditioning effects, since using (the square root of) the second moment in adaptive methods might be very different from using second order derivatives.

\subsection{Damping and regularization}\label{app:damp_reg}

It is well-known that Newton-type methods only work in a neighborhood of the optimal solution. Therefore, for convergence to a {\slmm}, we can use gradient descent-ascent to converge to a neighborhood of a local minimax point, and then use Newton-type methods such as GDN or CN. Another modification might be to add damping and regularization. For example, for the Newton step in GDN, we can instead apply:
\begin{align}
    \yv' \leftarrow \yv - \gamma (\yyv - \l \Iv)^{-1} \partial_\yv f(\xv, \yv),
\end{align}
where $\l > 0$ and $0 < \gamma \leq 1$. We call $\gamma$ \emph{the damping coefficient} and $\l$ \emph{the regularization coefficient}. If $\l = 0$ and $\g = 1$, then it is the pure Newton phase. If $\l \to \infty$ while $\g/\l$ stay fixed then the algorithm is simply gradient ascent. We could modify GDN by taking an adaptive scheme of $\gamma$ and $\lambda$ to stabilize this method. In a similar way, the Newton step of $\xv$ in CN could be modified as:
\begin{align}\label{eq:regularization_cn}
    \xv' \leftarrow \xv - \gamma (\TD_{\xv\xv} + \l \Iv)^{-1} \partial_\xv f(\xv, \yv),
\end{align}
We could also choose an adaptive scheme of $\gamma$ and $\lambda$, by choosing two sequences $\{\gamma_n\}$ and $\{\lambda_n\}$ s.t.~$\gamma_n \to 1$ and $\lambda_n \to 0$ as the iteration step goes to infinity. Another way to choose $\g$ is through line search \citep[e.g.][]{boyd2004convex}. 

\subsection{Newton-type methods are sensitive to initialization}

\begin{figure}[t]
\centering
\begin{subfigure}{0.5\textwidth}
    \includegraphics[width=\textwidth]{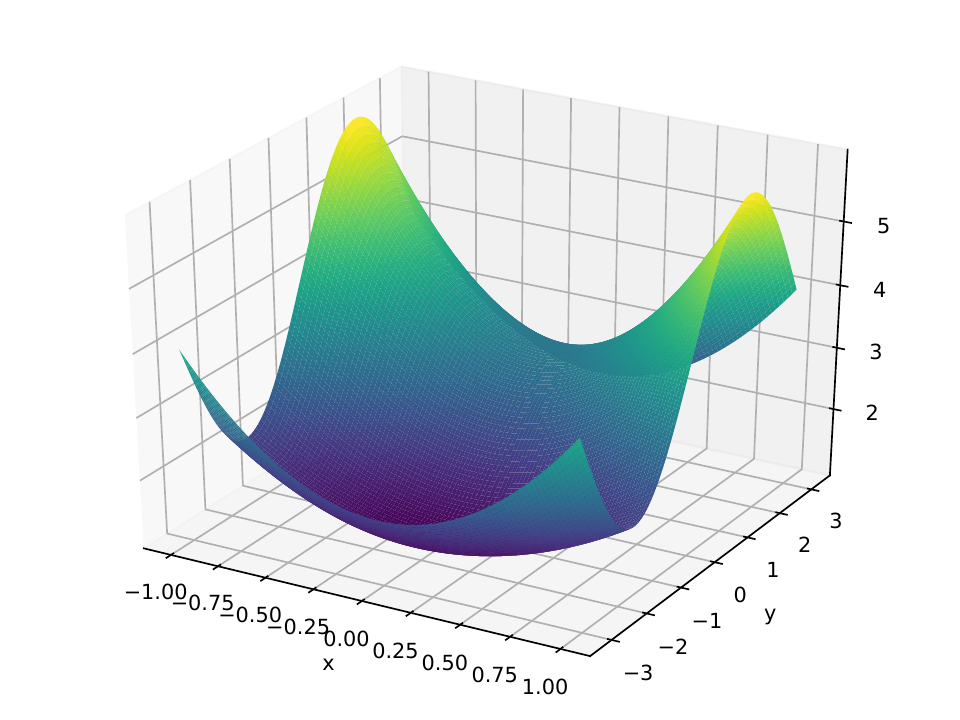}
\end{subfigure}
\begin{subfigure}{0.24\textwidth}
    \includegraphics[width=\textwidth]{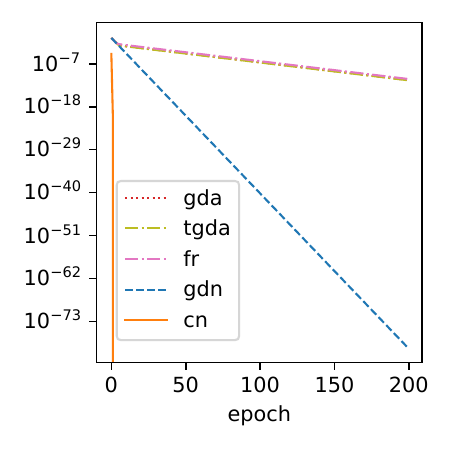}
\caption{near the {\slmm}}
\end{subfigure}
\begin{subfigure}{0.24\textwidth}
    \includegraphics[width=\textwidth]{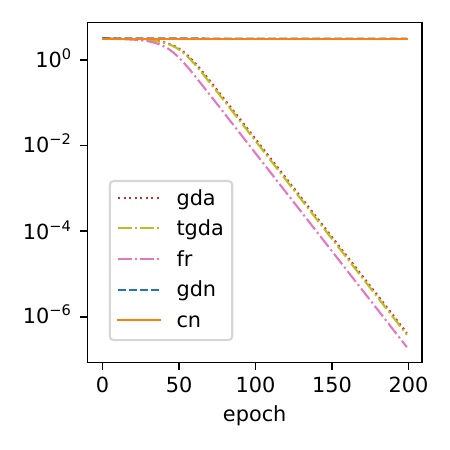}
\caption{near local minimum}
\end{subfigure}
\caption{({\bf left}) Landscape of $f(x, y) = (x^2 + 1)(2 + \sin y)$. $(0, \frac\pi2)$ is a strict local minimax point and $(0, -\frac\pi2)$ is a local minimum. ({\bf right}) The distance to the local minimax point for different algorithms with two initializations: near the local minimax point and near the local minimum. In the second case, Newton-type methods are attracted to the local minimum thus the distances stay constant.
}
\label{fig:minimax_vs_minimum}
\end{figure}

We give a simple example to demonstrate that the stable fixed points of Newton-type methods that we study may not always be {\slmm}s. Consider the following objective:
\begin{align}
    f(x, y) = (x^2 + 1)(2 + \sin y).
\end{align}
WLOG, we can restrict $-\pi \leq y \leq \pi$ due to periodicity. There are two types of stationary points: the local minimax point $(0, \frac\pi2)$ and the local minimum $(0, -\frac\pi2)$.
If initialized close enough to the local minimum, then GDN/CN will converge to the local minimum rather than the local minimax point. In contrast, first-order methods seem more robust to initialization in this case and always converges to the local minimax points.
See \Cref{fig:minimax_vs_minimum} for an illustration. 

%% file: appendices/D-simu-alt.tex
\section{Algorithmic modifications}\label{app:algmod}

In this appendix we study modifications of GDN and CN, including adding momentum, using total gradients, and generalization to general sum games. We perform asymptotic analysis \citep{polyak87book} for simplicity but general non-asymptotic analysis is also possible if we perform similar proofs as in Theorem \ref{thm:GDN_app}. 

\subsection{GDN with momentum}

Notice that in GDN \eqref{eq:gdn}, the minimizer still takes a gradient descent step. Therefore, we can accelerate it using Polyak's momentum \citep{Polyak64}.
Our Theorem~\ref{thm:gdnm} shows that after adding momentum, one can accelerate the convergence rate of GDN in Theorem~\ref{thm:GDN} to $1 - 2/(\sqrt{\kappa}+ 1)$.

We call the following algorithm as GD-Newton with (Polyak's) momentum:
\be
&&\xv_{t+1} = \xv_t - \a \n_\xv f(\xv_t, \yv_t) + \b (\xv_t - \xv_{t-1}),\\
&&\yv_{t+1} = \yv_t - (\yyv^{-1} \cdot \n_\yv) f(\tilde{\xv}_{t + 1}, \yv_t),
\en 
where $\tilde{\xv}_{t + 1} = \xv_t$ if it is simultaneous update and $\tilde{\xv}_{t + 1} = \xv_{t+1}$ if it is alternating update.

\begin{theoremEnd}[end, restate]{thm}\label{thm:gdnm}
At a {\slmm} $(\xv^*,\yv^*)$, denote $\lambda_1$ and $\lambda_n$ as the largest and the smallest eigenvalues of $\dxxv f(\xv^*,\yv^*)$. Given $\a = 4/(\sqrt{\lambda_1} + \sqrt{\lambda_n})^2$ and $\b = ((\sqrt{\kappa} - 1)/(\sqrt{\kappa} + 1))^2$, alternating GD-Newton with momentum can achieve a local convergence rate $1 - 2/(\sqrt{\kappa} + 1)$, with $\kappa := \lambda_1/\lambda_n$. 
\end{theoremEnd}

\begin{proof}
In this proof we use $\xxv f, \xyv f, \yxv f, \yyv f$ to denote second-order derivatives at the local minimax point $(\xv^*, \yv^*)$. For simultaneous GD-Newton with momentum, we use state augmentation $(\xv_t, \yv_t) \to (\xv_t, \yv_t, \xv_{t-1}, \yv_{t-1})$ and compute the corresponding Jacobian as:
\be
\begin{bmatrix}
(\beta + 1)\Iv - \alpha \xxv f & -\alpha \xyv f & -\beta\Iv & \zero \\ 
-(\yyv^{-1}\cdot\yxv) f & \zero & \zero & \zero \\
\Iv & \zero & \zero & \zero \\
\zero & \Iv & \zero & \zero
\end{bmatrix}.
\en
With \citet[Theorem 2.3]{zhang2019convergence} we have that for alternating GDN with momentum, the characteristic polynomial of the Jacobian is:
\be
\mu^2 - (\b + 1) \mu + \a \mu \lambda_i + \b = 0,
\en
with $\lambda_i \in \Sp(\dxxv f(\xv^*,\yv^*))$. This is equivalent to the characteristic polynomial of the process ($\wv_t \in \R^n$):
\be
\wv_{t+1} = \wv_t - \alpha \dxxv f(\xv^*, \yv^*) \cdot \wv_t + \beta (\wv_t - \wv_{t-1}).
\en
From \citet[Theorem 1, p.~65]{polyak87book} we can obtain the fastest convergence rate among all choices of $\alpha, \beta$. 
\end{proof}
\subsection{The effect of total derivative}\label{app:tgd}
In this subsection we discuss the effect of using total derivatives instead of partial derivatives in the update of $\xv$, in Newton-type methods. The effect of using total derivatives is to keep the same convergence rate while we can replace the alternating updates with simultaneous ones. Also, for general sum games it is necessary to use the total derivative in the update of $\xv$ (\Cref{app:gen_sum}).  


\subsubsection{TGD-Newton}\label{app:tgdn}

It is possible to combine TGD with Newton, the continuous version of such method has appeared in \cite{evtushenko1974iterative}.
\be\label{eq:tgdn}
\xv_{t+1} = \xv_t -\a  \TD_\xv f(\xv_t, \yv_t), \, \yv_{t+1} = \yv_t -  [(\yyv^{-1} \cdot \n_\yv) f](\xv_{t}, \yv_t). \en
The Jacobian of TGD-Newton at a \slmm is:
\be\label{eq:jnewton_tgd}
\Jv_{\rm Newton} = \begin{bmatrix}
\Iv - \a \TD_{\xv\xv} f & \zero \\
-(\yyv^{-1}\cdot\yxv) f & \zero
\end{bmatrix}.
\en
The convergence rate is the same as GD-Newton, with simultaneous or alternating updates. 

\subsubsection{Complete Newton}\label{app:tgd_cn}

If we replace the partial derivative with total derivative in \eqref{eq:newton}, we obtain:
\be
\label{eq:newton_tdg}
\xv_{t+1} = \xv_t - [(\TD_{\xv\xv}^{-1} \cdot\TD_\xv) f](\xv_t, \yv_t), \, 
\yv_{t+1} = \yv_t - [(\yyv^{-1} \cdot \n_\yv) f](\xv_{t}, \yv_t).
\en
The Jacobian can be evaluated as:
\be
\begin{bmatrix}
\zero & \zero \\
-(\yyv^{-1} \cdot\yxv) f & \zero
\end{bmatrix},
\en
which implies super-linear convergence. Combining with \Cref{app:tgdn} we can see that if the update of $\yv$ is Newton, the usage of the total derivative with simultaneous updates is equivalent to using the partial derivative with alternating updates.

\subsection{Alternating vs. simultaneous GDN}\label{app:alt_vs_simul}

In this subsection, we compare simultaneous GDN with alternating GDN in cases where $\a$ is small, as often required in experiments due to stochastic noise. In fact, simultaneous GD-Newton can converge even faster than alternating GD-Newton. Let us make clear of the definitions first. In fact, the algorithm we proposed in \eqref{eq:gdn} is using alternating update, whereas its simultaneous version is:
\be
\label{eq:gdn_s}
\xv_{t+1} = \xv_t - \a \n_\xv f(\xv_t, \yv_t), \, \yv_{t+1} = \yv_t - (\n_{\yv\yv}^{-1} \cdot \n_\yv) f(\xv_{t}, \yv_t).
\en

More detailed discussion can be found in \citet{zhang2019convergence}. 

\begin{theoremEnd}[end, restate]{thm}
Suppose at a strict local minimax point $(\xv^*, \yv^*)$, $[\xxv f, (\xyv \cdot \yyv^{-1} \cdot \yxv) f]=\zero$.\footnote{The commutator of two matrices $A, B$ is defined such that $[A, B] := AB - BA$.} Then there exist an orthogonal matrix $\Qv$ s.t.~$\xxv = \Qv \diag\{u_1, \dots, u_n\} \Qv^\top$ and $-(\xyv \cdot\yyv^{-1} \cdot\yxv) f= \Qv \diag\{v_1, \dots, v_n\} \Qv^\top$. If for any $i$, $0 < \a v_i < 1$ and $ \a u_i < 1 - \a v_i - \sqrt{\a v_i}$, then simultaneous GD-Newton always converges faster than alternating GD-Newton near the local minimax point $(\xv^*, \yv^*)$.
\end{theoremEnd}

\begin{proof}
The two characteristic polynomials for simultaneous and alternating methods are separately:
\be
\l^2 - (1 - \a u_i) \l + \a v_i = 0, \, \l(\l -1 + \a u_i + \a v_i) = 0, \, \forall \, i.
\en
We require that $\a v_i + \a u_i < 1$ for all $i$. In order for the simultaneous method to converge faster, suffices to have:
\be
\left|\frac{1}{2}(1 -\a u_i \pm \sqrt{(1 - \a u_i)^2 - 4 \a v_i})\right| < 1 - \a u_i - \a v_i,
\en
if $(1 -\a u_i)^2 \geq 4\a v_i$ and 
\be
\sqrt{\a v_i} < 1 - \a u_i - \a v_i,
\en
if $(1 -\a u_i)^2 < 4\a v_i$. Solving the two cases above gives 
\be
0 < \a v_i < 1, \, \a u_i  < 1 - \a v_i - \sqrt{\a v_i}.
\en
\end{proof}
When $\a$ is small enough, the condition for $\a$ is always satisfied. Specifically, if $\xxv = \zero$, $u_i = 0$ for all $i$, and we have:
\begin{cor}
Suppose at a strict local minimax point  $(\xv^*, \yv^*)$, $\xxv f = \zero$, $\yyv f \cl \zero$ and $0 < \a v_{1} < (3-\sqrt{5})/2$ for $v_{\max} = \l_{1}(-(\xyv \cdot\yyv^{-1}\cdot\yxv) f)$, then simultaneous GD-Newton always converges faster than alternating GD-Newton near $(\xv^*, \yv^*)$.
\end{cor}

\begin{proof}
Solving $1 - \a v_i - \sqrt{\a v_i} > 0$ and $0 < \a v_i < 1$ gives $0 < \a v_i < (3 - \sqrt{5})/{2}$. 
\end{proof}

The condition $\xxv f = \zero$, $\yyv f \cl \zero$ is often satisfied in GANs, see Example~\ref{eg:gan}.

\subsection{General sum games}\label{app:gen_sum}

GD-Newton could naturally be generalized to two-player general sum games \citep{fiez2019convergence}, by which we mean to minimize:
\be
\of(\xv) = \max_{\yv \in \Y^*(\xv)} f(\xv, \yv), \, \mbox{ where }\Y^*(\xv) = \argmin_{\yv\in \Y} g(\xv, \yv).
\en
We use $\n_\yv g(\xv, \yv) = \zero$ as an implicit function that determines the best response $r(\xv)$. It is unique when $\yyv g(\xv, \yv)\cg \zero$. It leads to the definition of a strict local Stackelberg equilibrium $(\xv, \yv) := (\xv, r(\xv))$ (\cite{fiez2019convergence}):
\be
&&\n_\yv g(\xv, \yv) = \zero, \, \TD_\xv f(\xv, r(\xv)) = \zero, \\
&& \yyv g(\xv, \yv)\cg \zero, \, \TD_{\xv\xv} f(\xv, r(\xv)) \cg \zero.
\en
Using implicit function theorem we obtain that:
\be\label{eq:gen_td1}
\TD_\xv f := \n_\xv f - (\xyv \cdot \yyv^{-1})g\cdot \n_\yv f,
\en
and (see also \citet{wang2019solving}),
\be\label{eq:gen_td2}
\TD_{\xv\xv} f &:=& \xxv f -\xyv f \cdot (\xyv \cdot \yyv^{-1})g - \n_\xv [(\xyv \cdot\yyv^{-1})g\cdot \n_\yv f] + \tr &+& \n_\yv [(\xyv \cdot\yyv^{-1})g\cdot \n_\yv f] (\yyv^{-1}\cdot\yxv) g,
\en
where both sides in \eqref{eq:gen_td1} and \eqref{eq:gen_td2} are applied on $(\xv, r(\xv))$. In the zero-sum case, $g= -f$. Hence \eqref{eq:gen_td1} and \eqref{eq:gen_td2} reduce to \eqref{eq:TD1} and \eqref{eq:TD2} respectively. These functions induce GD-Newton for general sum games naturally:
\be
&& \xv_{t+1} = \xv_t - \a_\La \TD_\xv f(\xv_t, \yv_t), \, \yv_{t+1} = \yv_t - [(\yyv^{-1}\cdot\n_\yv)g](\xv_{t+1}, \yv_t).
\en
Let us first compute the Jacobian of the simultaneous version of GDN near a strict local Stackelberg equilibrium, where $\yv_{t+1} = \yv_t - [(\yyv^{-1}\cdot\n_\yv)g](\xv_{t}, \yv_t)$:
\be\label{eq:jnewton}
\Jv_{\rm GDN} = \begin{bmatrix}
\Iv - \a_\La (\xxv f - \n_\xv [(\xyv\cdot \yyv^{-1})g\cdot \n_\yv f] ) & -\a_\La\xyv f +  \a_\La \n_\yv [(\xyv\cdot \yyv^{-1})g\cdot \n_\yv f] \\
-(\yyv^{-1}\cdot\yxv)g  & \zero
\end{bmatrix}.
\en
Computing the spectrum and using \citet[Theorem 2.3]{zhang2019convergence} we have:
\begin{theoremEnd}[end, restate]{thm}\label{thm:gdn_2}
Near a \slmm, GDN can achieve linear convergence $\max_i |1 - \a_\La \l_i|$ with $\l_i \in \Sp(\TD_{\xv\xv}f)$. With $\a_\La = 2/(\l_1 + \l_n)$, GDN can achieve a local convergence rate $1 - 2/(\kappa + 1)$, with $\kappa := \l_1/\l_n$, and $\l_1$ ($\l_n$) the largest (smallest) eigenvalue of $\TD_{\xv\xv}f$. 
\end{theoremEnd}

\begin{proof}
The characteristic equation for simultaneous GDN is:
\be
\det\left(\l ((\l - 1)\Iv \!+\! \a_\La  (\xxv f - \n_\xv [(\xyv\cdot \yyv^{-1})g\cdot \n_\yv f] )) \!-\!  \a_\La (\xyv f \!-\! \n_\yv [(\xyv\cdot \yyv^{-1})g\cdot \n_\yv f]) \cdot ( \yyv^{-1}\cdot\yxv )g \right) = 0.
\en
For alternating updates, using \citet[Theorem 2.3]{zhang2019convergence}, we take $ (\yyv^{-1} \cdot \yxv) g\to \l (\yyv^{-1}\cdot \yxv) g$ in \eqref{eq:jnewton}, and the characteristic equation becomes:
\be
\det(\l ((\l - 1)\Iv + \a_\La \dxxv f)) = 0,
\en
where $\dxxv f$ is defined in \eqref{eq:gen_td2}. which reduces to $\l = 0$ or
\be
\det((\l - 1)\Iv + \a_\La \dxxv f) = 0.
\en

So, for any strict local minimax point, alternating GD-Newton converges for small enough $\a_\La$, and the convergence rate of $\|\zv_t - \zv^*\|$ is:
\be\label{eq:sp_ra_gdn}
\rho(\Jv_{\rm GDN}) = \max_i |1 - \a_\La \l_i|,
\en

From \eqref{eq:sp_ra_gdn} it suffices to solve the following minimization problem:
\be
\min_{\a_\La > 0} \max_i |1 - \a_\La \l_i|.
\en
$\max_i |1 - \a_\La \l_i|$ is a piece-wise linear function and it is minimized at the point where $\a_\La \l_1 - 1 = 1 - \a_\La \l_n$. Solving it gives $\a_\La = 2/(\l_1 + \l_n)$ and the local convergence rate $1 - 2/(\kappa_{\La} + 1)$, with $\kappa_{\La} := \l_1/\l_n$.
\end{proof}


%% file: arxiv.bbl
\begin{thebibliography}{}

\bibitem[Arjovsky et~al., 2017]{arjovsky2017wasserstein}
Arjovsky, M., Chintala, S., and Bottou, L. (2017).
\newblock Wasserstein generative adversarial networks.
\newblock In {\em International Conference on Machine Learning}.

\bibitem[Arrow et~al., 1958]{arrow1958studies}
Arrow, K., Hurwicz, L., and Uzawa, H. (1958).
\newblock {\em Studies in linear and non-linear programming}.
\newblock Stanford University Press.

\bibitem[Balduzzi et~al., 2018]{balduzzi2018mechanics}
Balduzzi, D., Racaniere, S., Martens, J., Foerster, J., Tuyls, K., and Graepel,
  T. (2018).
\newblock The mechanics of n-player differentiable games.
\newblock In {\em International Conference on Machine Learning}, pages
  354--363.

\bibitem[Bertsekas, 1997]{bertsekas1997nonlinear}
Bertsekas, D.~P. (1997).
\newblock Nonlinear programming.
\newblock {\em Journal of the Operational Research Society}, 48(3):334--334.

\bibitem[Borkar, 2008]{Borkar08}
Borkar, V.~S. (2008).
\newblock {\em
  \href{https://link.springer.com/book/10.1007/978-93-86279-38-5}{Stochastic
  Approximation: A Dynamical Systems Viewpoint}}.
\newblock Springer.

\bibitem[Boyd and Vandenberghe, 2004]{boyd2004convex}
Boyd, S. and Vandenberghe, L. (2004).
\newblock {\em Convex optimization}.
\newblock Cambridge university press.

\bibitem[Dai et~al., 2018]{dai2018sbeed}
Dai, B., Shaw, A., Li, L., Xiao, L., He, N., Liu, Z., Chen, J., and Song, L.
  (2018).
\newblock Sbeed: Convergent reinforcement learning with nonlinear function
  approximation.
\newblock In {\em International Conference on Machine Learning}, pages
  1125--1134.

\bibitem[Du et~al., 2017]{du2017stochastic}
Du, S.~S., Chen, J., Li, L., Xiao, L., and Zhou, D. (2017).
\newblock Stochastic variance reduction methods for policy evaluation.
\newblock In {\em International Conference on Machine Learning}, pages
  1049--1058.

\bibitem[Evtushenko, 1974a]{evtushenko1974some}
Evtushenko, Y. (1974a).
\newblock
  \href{https://www.sciencedirect.com/science/article/abs/pii/0041555374901074}{Some
  local properties of minimax problems}.
\newblock {\em USSR Computational Mathematics and Mathematical Physics},
  14(3):129 -- 138.

\bibitem[Evtushenko, 1974b]{evtushenko1974iterative}
Evtushenko, Y.~G. (1974b).
\newblock
  \href{https://www.sciencedirect.com/science/article/abs/pii/0041555374901955}{Iterative
  methods for solving minimax problems}.
\newblock {\em USSR Computational Mathematics and Mathematical Physics},
  14(5):52--63.

\bibitem[Farnia and Ozdaglar, 2020]{farnia2020gans}
Farnia, F. and Ozdaglar, A. (2020).
\newblock Do {GAN}s always have {N}ash equilibria?
\newblock In {\em International Conference on Machine Learning}.

\bibitem[Fiez et~al., 2020]{fiez2019convergence}
Fiez, T., Chasnov, B., and Ratliff, L.~J. (2020).
\newblock
  \href{https://proceedings.icml.cc/static/paper_files/icml/2020/3821-Paper.pdf}{Implicit
  Learning Dynamics in {S}tackelberg Games: Equilibria Characterization,
  Convergence Analysis, and Empirical Study}.
\newblock In {\em International Conference on Machine Learning}.

\bibitem[Foerster et~al., 2018]{foerster2018learning}
Foerster, J., Chen, R.~Y., Al-Shedivat, M., Whiteson, S., Abbeel, P., and
  Mordatch, I. (2018).
\newblock Learning with opponent-learning awareness.
\newblock In {\em Proceedings of the 17th International Conference on
  Autonomous Agents and MultiAgent Systems}, pages 122--130. International
  Foundation for Autonomous Agents and Multiagent Systems.

\bibitem[Ganin et~al., 2016]{ganin2016domain}
Ganin, Y., Ustinova, E., Ajakan, H., Germain, P., Larochelle, H., Laviolette,
  F., Marchand, M., and Lempitsky, V. (2016).
\newblock Domain-adversarial training of neural networks.
\newblock {\em The Journal of Machine Learning Research}, 17(1):2096--2030.

\bibitem[Goodfellow et~al., 2014]{goodfellow2014generative}
Goodfellow, I., Pouget-Abadie, J., Mirza, M., Xu, B., Warde-Farley, D., Ozair,
  S., Courville, A., and Bengio, Y. (2014).
\newblock Generative adversarial nets.
\newblock In {\em Advances in neural information processing systems}, pages
  2672--2680.

\bibitem[Heusel et~al., 2017]{heusel2017gans}
Heusel, M., Ramsauer, H., Unterthiner, T., Nessler, B., and Hochreiter, S.
  (2017).
\newblock {GAN}s trained by a two time-scale update rule converge to a local
  {N}ash equilibrium.
\newblock In {\em Advances in neural information processing systems}, pages
  6626--6637.

\bibitem[Hinton et~al., 2012]{rmsprop}
Hinton, G., Srivastava, N., and Swersky, K. (2012).
\newblock Rmsprop: Divide the gradient by a running average of its recent
  magnitude.
\newblock {\em Neural networks for machine learning, Coursera lecture 6e}.

\bibitem[Hinton and Salakhutdinov, 2006]{hinton2006reducing}
Hinton, G.~E. and Salakhutdinov, R.~R. (2006).
\newblock Reducing the dimensionality of data with neural networks.
\newblock {\em science}, 313(5786):504--507.

\bibitem[Hsieh et~al., 2019]{hsieh2019convergence}
Hsieh, Y.-G., Iutzeler, F., Malick, J., and Mertikopoulos, P. (2019).
\newblock On the convergence of single-call stochastic extra-gradient methods.
\newblock In {\em Advances in Neural Information Processing Systems}, pages
  6936--6946.

\bibitem[Jin et~al., 2020]{jin2019minmax}
Jin, C., Netrapalli, P., and Jordan, M.~I. (2020).
\newblock What is local optimality in nonconvex-nonconcave minimax
  optimization?
\newblock In {\em arxiv: 1902.00618v2 (published at International Conference on
  Machine Learning)}.

\bibitem[Kingma and Ba, 2015]{kingma2014adam}
Kingma, D.~P. and Ba, J. (2015).
\newblock Adam: A method for stochastic optimization.
\newblock In {\em International Conference on Learning Representations}.

\bibitem[Korpelevich, 1976]{korpelevich1976extragradient}
Korpelevich, G. (1976).
\newblock The extragradient method for finding saddle points and other
  problems.
\newblock {\em Matecon}, 12:747--756.

\bibitem[Madry et~al., 2018]{madry2017towards}
Madry, A., Makelov, A., Schmidt, L., Tsipras, D., and Vladu, A. (2018).
\newblock Towards deep learning models resistant to adversarial attacks.
\newblock In {\em International Conference on Learning Representations}.

\bibitem[Martens, 2010]{Martens10}
Martens, J. (2010).
\newblock Deep learning via {H}essian-free optimization.
\newblock In {\em Proceedings of the 27th International Conference on Machine
  Learning}, pages 735--–742.

\bibitem[Mertikopoulos et~al., 2019]{mertikopoulos2018optimistic}
Mertikopoulos, P., Lecouat, B., Zenati, H., Foo, C.-S., Chandrasekhar, V., and
  Piliouras, G. (2019).
\newblock Optimistic mirror descent in saddle-point problems: Going the
  extra(-gradient) mile.
\newblock In {\em International Conference on Learning Representations}.

\bibitem[Mescheder et~al., 2017]{mescheder2017numerics}
Mescheder, L., Nowozin, S., and Geiger, A. (2017).
\newblock The numerics of {GAN}s.
\newblock In {\em Advances in Neural Information Processing Systems}, pages
  1825--1835.

\bibitem[Metz et~al., 2017]{metz2016unrolled}
Metz, L., Poole, B., Pfau, D., and Sohl-Dickstein, J. (2017).
\newblock Unrolled generative adversarial networks.
\newblock In {\em International Conference on Learning Representations}.

\bibitem[Meyer, 2000]{meyer2000matrix}
Meyer, C.~D. (2000).
\newblock {\em Matrix analysis and applied linear algebra}, volume~71.
\newblock Siam.

\bibitem[Miyato et~al., 2018]{miyato2018spectral}
Miyato, T., Kataoka, T., Koyama, M., and Yoshida, Y. (2018).
\newblock Spectral normalization for generative adversarial networks.
\newblock In {\em International Conference on Learning Representations}.

\bibitem[Mohri et~al., 2019]{MohriSS19}
Mohri, M., Sivek, G., and Suresh, A.~T. (2019).
\newblock \href{http://proceedings.mlr.press/v97/mohri19a.html}{Agnostic
  Federated Learning}.
\newblock In {\em International Conference on Machine Learning}, pages
  4615--4625.

\bibitem[Morgenstern and von Neumann, 1953]{morgenstern1953theory}
Morgenstern, O. and von Neumann, J. (1953).
\newblock {\em Theory of games and economic behavior}.
\newblock Princeton university press.

\bibitem[Nagarajan and Kolter, 2017]{nagarajan2017gradient}
Nagarajan, V. and Kolter, J.~Z. (2017).
\newblock Gradient descent {GAN} optimization is locally stable.
\newblock In {\em Advances in Neural Information Processing Systems}, pages
  5585--5595.

\bibitem[Nash, 1950]{nash1950equilibrium}
Nash, J.~F. (1950).
\newblock Equilibrium points in n-person games.
\newblock {\em Proceedings of the national academy of sciences}, 36(1):48--49.

\bibitem[Nemirovski, 2004]{nemirovski2004prox}
Nemirovski, A. (2004).
\newblock Prox-method with rate of convergence o (1/t) for variational
  inequalities with lipschitz continuous monotone operators and smooth
  convex-concave saddle point problems.
\newblock {\em SIAM Journal on Optimization}, 15(1):229--251.

\bibitem[Nesterov, 2003]{nesterov2003introductory}
Nesterov, Y. (2003).
\newblock {\em Introductory lectures on convex optimization: A basic course},
  volume~87.
\newblock Springer Science \& Business Media.

\bibitem[Pearlmutter, 1994]{pearlmutter1994fast}
Pearlmutter, B.~A. (1994).
\newblock Fast exact multiplication by the hessian.
\newblock {\em Neural computation}, 6(1):147--160.

\bibitem[Polyak, 1987]{polyak87book}
Polyak, B. (1987).
\newblock {\em Introduction to Optimization}.
\newblock Optimization Software Inc.

\bibitem[Polyak, 1964]{Polyak64}
Polyak, B.~T. (1964).
\newblock \href{https://doi.org/10.1016/0041-5553(64)90137-5}{Some methods of
  speeding up the convergence of iteration methods}.
\newblock {\em USSR Computational Mathematics and Mathematical Physics},
  4(5):1--17.

\bibitem[Popov, 1980]{popov1980modification}
Popov, L.~D. (1980).
\newblock A modification of the {Arrow--Hurwicz} method for search of saddle
  points.
\newblock {\em Mathematical Notes}, 28(5):845--848.

\bibitem[Reddi et~al., 2018]{reddi2018convergence}
Reddi, S.~J., Kale, S., and Kumar, S. (2018).
\newblock On the convergence of {A}dam and beyond.
\newblock In {\em International Conference on Learning Representations}.

\bibitem[Royer et~al., 2020]{royer2020newton}
Royer, C.~W., O’Neill, M., and Wright, S.~J. (2020).
\newblock A {N}ewton-{CG} algorithm with complexity guarantees for smooth
  unconstrained optimization.
\newblock {\em Mathematical Programming}, 180(1):451--488.

\bibitem[Sagun et~al., 2016]{sagun2016singularity}
Sagun, L., Bottou, L., and LeCun, Y. (2016).
\newblock Singularity of the hessian in deep learning.
\newblock {\em arXiv preprint arXiv:1611.07476}.

\bibitem[Sinha et~al., 2018]{sinha2018certifiable}
Sinha, A., Namkoong, H., and Duchi, J. (2018).
\newblock Certifiable distributional robustness with principled adversarial
  training.
\newblock In {\em International Conference on Learning Representations}.

\bibitem[Song et~al., 2019]{song2019learning}
Song, J., Kalluri, P., Grover, A., Zhao, S., and Ermon, S. (2019).
\newblock Learning controllable fair representations.
\newblock In {\em The 22nd International Conference on Artificial Intelligence
  and Statistics}, pages 2164--2173.

\bibitem[Wang et~al., 2020]{wang2019solving}
Wang, Y., Zhang, G., and Ba, J. (2020).
\newblock \href{https://openreview.net/pdf?id=Hkx7_1rKwS}{On Solving Minimax
  Optimization Locally: A Follow-the-Ridge Approach}.
\newblock In {\em International Conference on Learning Representations}.

\bibitem[Werbos, 1988]{werbos1988backpropagation}
Werbos, P. (1988).
\newblock Backpropagation: Past and future.
\newblock In {\em Proceedings of the Second International Conference on Neural
  Network}, volume~1, pages 343--353. IEEE.

\bibitem[Zhang and Yu, 2020]{zhang2019convergence}
Zhang, G. and Yu, Y. (2020).
\newblock Convergence of gradient methods on bilinear zero-sum games.
\newblock In {\em International Conference on Learning Representations}.

\bibitem[Zhang et~al., 2022]{zhang2020optimality}
Zhang, G., Yu, Y., and Poupart, P. (2022).
\newblock Optimality and stability in non-convex smooth games.
\newblock {\em Journal of Machine Learning Research}.

\end{thebibliography}
